\documentclass{article}
\usepackage[final, nonatbib]{neurips}
\usepackage[utf8]{inputenc}
\usepackage{algpseudocode}
\usepackage{bm}
\usepackage{newtxtext}
\usepackage{tabularx}
\usepackage[backend=biber,style=nature]{biblatex}
\usepackage{textcomp}
\usepackage{amssymb}
\usepackage{pifont}
\usepackage{placeins}
\usepackage{amsmath,amssymb,amsthm, amsfonts}
\usepackage[table]{xcolor}
\usepackage{dsfont}
\usepackage{tcolorbox}
\usepackage{tikz}
\usetikzlibrary{decorations.pathreplacing, positioning, shapes.geometric, arrows.meta, fit}
\usepackage{subcaption}
\usepackage{graphicx, xcolor}
\usepackage{geometry}
\usepackage{booktabs, array}
\usepackage{footnote}
\usepackage{tablefootnote}
\usepackage{fancybox}
\usepackage{adjustbox}
\usepackage[normalem]{ulem}     % for todos
\usepackage{parskip}        % for paragraphing
\usepackage{titlesec}        % for spacing in paragraphs
\usepackage{tabularx}
\usepackage{pifont}
\usepackage{xspace}
\usepackage{wrapfig}
\usepackage{soul}
\usepackage{booktabs}% http://ctan.org/pkg/booktabs
\usepackage{authblk}
\usepackage{setspace}
\usepackage{listings}
\usepackage{hyperref}
\usepackage{cleveref}

\definecolor{mediumred}{RGB}{150,0,0}
\definecolor{green}{RGB}{0,130,0}

\definecolor{gooddarkblue}{rgb}{0,0.1,0.4}
\definecolor{gooddarkred}{rgb}{0.6,0,0.1}
\newcommand{\tightunderline}[1]{%
  \setul{+0.3ex}{.1ex}% Adjust these values: {vertical displacement}{line thickness}
  \ul{#1}%
}

\renewcommand{\url}[1]{\href{#1}{\tightunderline{#1}}}

\usepackage[toc,page,header]{appendix}
\usepackage{minitoc}

\renewcommand \thepart{}
\renewcommand \partname{}

\usepackage{amsmath,amsfonts,bm}

\def\eqref#1{equation~\ref{#1}}
\def\1{\bm{1}}

\def\vh{{\bm{h}}}

\def\vs{{\bm{s}}}
\def\vt{{\bm{t}}}

\def\vx{{\bm{x}}}

\def\mC{{\bm{C}}}

\def\mI{{\bm{I}}}

\def\mM{{\bm{M}}}

\def\mX{{\bm{X}}}

\DeclareMathAlphabet{\mathsfit}{\encodingdefault}{\sfdefault}{m}{sl}
\SetMathAlphabet{\mathsfit}{bold}{\encodingdefault}{\sfdefault}{bx}{n}

\newcommand{\E}{\mathbb{E}}

\newcommand{\R}{\mathbb{R}}

\newcommand{\KL}{D_{\mathrm{KL}}}

\providecommand{\keywords}[1]{
  \small	
  \textbf{Key words:} #1}

\definecolor{mydarkblue}{rgb}{0,0.1,0.4}
\definecolor{mydarkred}{rgb}{0.6,0,0.1}

\hypersetup{% for better colors 
    colorlinks=true,
    citecolor=mydarkblue,
    linkcolor=mydarkred,
    filecolor=mydarkblue,  
    urlcolor=mydarkblue,  
}

\title{Reliable Generation of Privacy-preserving\\ Synthetic Electronic Health Record \\Time Series via Diffusion Models}

\author{\textbf{Muhang Tian}$^1$\hspace{1em}\textbf{Bernie Chen}$^{\dagger,1}$\hspace{1em}\textbf{Allan Guo}$^{\dagger,1}$\hspace{1em}\textbf{Shiyi Jiang}$^1$\hspace{1em}\textbf{Anru R. Zhang}$^{*,1}$}

\date{}

\newcommand{\firstedit}{\textcolor{black}}
\newcommand{\secondedit}{\textcolor{black}}

\def\backbone{{\textnormal{$\vs_{\theta}$}}}
\def\aatrain{{\textnormal{$AA_{\text{train}}$}}}
\def\aatest{{\textnormal{$AA_{\text{test}}$}}}
\newtheorem{definition}{Definition}
\newcommand{\ours}{\textsc{TimeDiff}\xspace}

\begin{document}

\maketitle

\def\thefootnote{}\footnotetext{
$^\dagger$Equal contribution. $^*$Corresponding author. $^1$Duke University. \texttt{\{muhang.tian, bernie.chen, allan.guo, shiyi.jiang, anru.zhang\}@duke.edu}
}
\def\thefootnote{\arabic{footnote}}
\begin{abstract}
%Electronic Health Records (EHRs) contain a large collection of patient-level data including laboratory tests, medications, and diagnoses. EHR provides valuable resources for medical data analysis such as mortality prediction and disease phenotyping. However, due to concerns about patient privacy, access to EHR is usually limited, which impedes the development of downstream analysis. Researchers have been developing various methods that produce privacy-preserved EHR data. One promising solution is to create synthetic data that resembles real EHR data utilizing generative models. Unfortunately, existing works target binary or categorical data generation and the generated data usually has limited diversity and utility. In this study, we propose to create diverse and realistic synthetic EHR time series data using Denoising Diffusion Probabilistic Models (DDPM). We conduct experiments on six datasets and compare the proposed model with seven baselines. Results suggest that our proposed method outperforms all the baselines in terms of data utility and requires less training effort. Our proposed method can provide insights into the downstream medical data analysis.

\noindent\textbf{Objective:} Electronic Health Records (EHRs) are rich sources of patient-level data, offering valuable resources for medical data analysis. However, privacy concerns often restrict access to EHRs, hindering downstream analysis. \firstedit{Current EHR de-identification methods are flawed and can lead to potential privacy leakage. Additionally, existing publicly available EHR databases are limited, preventing the advancement of medical research using EHR. This study aims to overcome these challenges by generating realistic and privacy-preserving synthetic electronic health records (EHRs) time series efficiently.}
%including laboratory tests, medications, and diagnoses
%This study aims to overcome these challenges by generating synthetic electronic health records (EHRs) time series reliably and efficiently.

\noindent\textbf{Materials and Methods:} We introduce a new method for generating diverse and realistic synthetic EHR time series data using Denoising Diffusion Probabilistic Models (DDPM). We conducted experiments on six databases: Medical Information Mart for Intensive Care III and IV (MIMIC-III/IV), the eICU Collaborative Research Database (eICU), %and high time resolution ICU dataset (HiRID), 
and non-EHR datasets on Stocks and Energy. We compared our proposed method with nine existing methods.

\noindent\textbf{Results:} Our results demonstrate that our approach significantly outperforms all existing methods in terms of data \firstedit{fidelity} while requiring less training effort. \firstedit{Additionally, data generated by our method yields a lower discriminative accuracy compared to other baseline methods, indicating the proposed method can generate data with less privacy risk.}

%\noindent\textbf{Discussion}:

\noindent\textbf{Conclusion:} The proposed diffusion-model-based method can reliably and efficiently generate synthetic EHR time series, which facilitates the downstream medical data analysis. Our numerical results show the superiority of the proposed method over all other existing methods. 
\end{abstract}


\section{Introduction \label{sec:intro}}
The Electronic Health Record (EHR) is a digital version of the patient's medical history maintained by healthcare providers.
It includes information such as demographic attributes, vital signals, and lab measurements that are sensitive and important for clinical research.
Researchers have been utilizing statistical and machine learning (ML) methods to analyze EHR for a variety of downstream tasks such as disease diagnosis, in-hospital mortality prediction, and disease phenotyping \cite{Shickel18DeepEHR, Goldstein2017OpportunitiesAC}.
However, due to privacy concerns, EHR data is strictly regulated, and thus the availability of EHR data \secondedit{for research and education} is often limited, creating barriers to the development of computational models in the field of healthcare.
Widely used EHR de-identification methods to preserve patient information privacy are criticized for having high risks of re-identification of the individuals \cite{Benitez2010EvaluatingRR}.

Instead of applying \secondedit{traditional de-identification} methods that can adversely affect EHR data utility \cite{janmey2018re}, EHR synthetic data generation is one promising solution to protect patient privacy.
Realistic synthetic data preserves crucial clinical information in real data while preventing patient information leakage \cite{Yan2022AMB,Yoon2023EHRSafeGH}.
Synthetic data also has the added benefit of providing a larger sample size for downstream analysis than de-identifying real samples \cite{gonzales2023synthetic}.
As a result, more research initiatives have begun to consider synthetic data sharing, such as the National COVID Cohort Collaborative supported by the U.S. National Institutes of Health and the Clinical Practice Research Datalink sponsored by the U.K. National Institute for Health and Care Research \cite{Haendel2020TheNC, Herrett2015DataRP}. 
With the advancement in machine learning techniques, applying generative models to synthesize high-fidelity EHR data is popular research of interest \cite{Yan2022AMB}.
Recent advances in generative models have shown significant success in generating realistic high-dimensional data like images, audio, and texts \cite{Gui2020ARO, Yi2018GenerativeAN}, suggesting the potential for these models to handle EHR data with complex statistical characteristics.

Some representative work utilizing generative models for EHR data synthesis includes medGAN \cite{Choi2017GeneratingMD}, medBGAN \cite{Baowaly2018SynthesizingEH}, and EHR-Safe \cite{Yoon2023EHRSafeGH}.\footnote{\label{note}We could not obtain code implementation for this work even after reaching out to the authors. Therefore, we are unable to compare \ours with this work's proposed methods.}
However, most approaches to EHR data synthesis are GAN-based, and GANs are known for their difficulties in model training and deployments due to training instability and mode collapse \cite{Saxena21GAN}.
Recently, diffusion probabilistic models have shown superb ability over GANs in generating high-fidelity image data \cite{ho2020denoising, Nichol2021ImprovedDD, rombach2022high}.
A few studies thus propose to generate synthetic EHR data via diffusion models given their remarkable data generation performance \cite{he2023meddiff, yuan2023ehrdiff}.
However, most EHR data synthesis methods, either GAN-based or diffusion-based, focus on binary or categorical variables such as the International Classification of Diseases (ICD) codes.
Additionally, there is limited prior work on generating EHR data with temporal information, and most state-of-the-art time series generative models are GAN-based.
\citeauthor{Kuo2023SyntheticHL} \cite{Kuo2023SyntheticHL} studied the diffusion models for EHR time series generation with focus only on continuous-valued time series.\footref{note}
It resorts to Gaussian diffusion for generating discrete sequences, treating them similarly to real-valued sequences but with further post-processing of the model output.
These observations motivate us to bridge the gap by introducing a novel direct diffusion-based method to generate realistic EHR time series data with mixed variable types.

Specifically, we make the following contributions in this paper:
\begin{itemize}
    \item We propose \ours, a new diffusion probabilistic model that uses a bidirectional recurrent neural network (BRNN) architecture for realistic privacy-preserving EHR time series generation.
    
    \item To our best knowledge, \ours is the first work introducing a mixed diffusion approach that combines multinomial and Gaussian diffusion for EHR time series generation. \ours can simultaneously generate both \firstedit{continuous} and discrete-valued time series.
    
    \item We demonstrate that \ours outperforms state-of-the-art methods for time series data generation by a big margin in terms of data fidelity and privacy. Additionally, our model requires less training effort than GAN-based methods.
\end{itemize}

\section{Background and Significance \label{sec:related}}
\subsection{Time series generation}
Prior sequential generation methods using GANs rely primarily on binary adversarial feedback \cite{mogren2016crnngan, esteban2017realvalued}, and supervised sequence models mainly focus on tasks such as prediction \cite{dai2015semisupervised}, forecasting \cite{lyu2018improving}, and classification \cite{srivastava2016unsupervised}.
TimeGAN \cite{yoon2019timegan} was one of the first methods to preserve temporal dynamics in time series synthesis.
The architecture comprises an embedding layer, recovery mechanism, generator, and discriminator, trained using both supervised and unsupervised losses.
GT-GAN \cite{jeon2022gtgan} considers the generation of both regular and irregular time series data using a neural controlled differential equation (NCDE) encoder \cite{kidger2020neural} and GRU-ODE decoder \cite{de2019gru}. 
This framework, combined with a continuous time flow processes (CTFPs) generator \cite{deng2021modeling} and a GRU-ODE discriminator, outperformed existing methods in general-purpose time series generation.
Recently, \secondedit{\citeauthor{pmlr-v202-bilos23a}} \cite{pmlr-v202-bilos23a} proposed to generate time series data for forecasting and imputation using discrete or continuous stochastic process diffusion (DSPD/CSPD).
Their proposed method views time series as discrete realizations of an underlying continuous function.
Both DSPD and CSPD use either the Gaussian or Ornstein-Uhlenbec process to model noise and apply it to the entire time series.
The learned distribution over continuous functions is then used to generate synthetic time series samples.

\subsection{Diffusion models}
Diffusion models \cite{sohl2015deep} have been proposed and achieved excellent performance in the field of computer vision and natural language processing.
\secondedit{\citeauthor{ho2020denoising}} \cite{ho2020denoising} proposed denoising diffusion probabilistic models (DDPM) that generate high-quality images by recovering from white latent noise.
\secondedit{\citeauthor{Gu2021VectorQD}} \cite{Gu2021VectorQD} proposed a vector-quantized diffusion model on text-to-image synthesis with significant improvement over GANs regarding scene complexity and diversity of the generated images.
\secondedit{\citeauthor{Dhariwal2021DiffusionMB}} \cite{Dhariwal2021DiffusionMB} suggested that the diffusion models with optimized architecture outperform GANs on image synthesis tasks.
\secondedit{\citeauthor{Saharia2022PhotorealisticTD}} \cite{Saharia2022PhotorealisticTD} proposed a diffusion model, Imagen, incorporated with a language model for text-to-image synthesis with state-of-the-art results. 
\secondedit{\citeauthor{Kotelnikov2022TabDDPMMT}} \cite{Kotelnikov2022TabDDPMMT} introduced TabDDPM, an extension of DDPM for heterogeneous tabular data generation, outperforming GAN-based models.
\secondedit{\citeauthor{Das2023ChiroDiffMC}} \cite{Das2023ChiroDiffMC} proposed ChiroDiff, a diffusion model that considers temporal information and generates chirographic data.
Besides advancements in practical applications, some recent developments in theory for diffusion models demonstrate the effectiveness of this model class. Theoretical foundations explaining the empirical success of diffusion or score-based generative models have been established \cite{song2019generative, song2020improved, chen2022sampling}.

\subsection{EHR data generation}
There exists a considerable amount of prior work on generating EHR data.
\secondedit{\citeauthor{Choi2017GeneratingMD}} \cite{Choi2017GeneratingMD} proposed medGAN that generates EHR discrete variables.
Built upon medGAN, \secondedit{\citeauthor{Baowaly2018SynthesizingEH}} \cite{Baowaly2018SynthesizingEH} suggested two models, medBGAN and medWGAN, that synthesize EHR binary or discrete variables on International Classification of Diseases (ICD) codes.
\secondedit{\citeauthor{Yan2020GeneratingEH}} \cite{Yan2020GeneratingEH} developed a GAN that can generate high-utility EHR with both discrete and continuous data. 
\secondedit{\citeauthor{Biswal2020EVAGL}} \cite{Biswal2020EVAGL} proposed the EHR Variational Autoencoder that synthesizes sequences of EHR discrete variables (i.e., diagnosis, medications, and procedures). 
\secondedit{\citeauthor{he2023meddiff}} \cite{he2023meddiff} developed MedDiff, a diffusion model that generates user-conditioned EHR discrete variables. 
\secondedit{\citeauthor{yuan2023ehrdiff}} \cite{yuan2023ehrdiff} created EHRDiff by utilizing the diffusion model to generate a collection of ICD diagnosis codes.
\secondedit{\citeauthor{naseer2023scoehr}} \cite{naseer2023scoehr} used continuous-time diffusion models to generate synthetic EHR tabular data.
\secondedit{\citeauthor{ceritli2023synthesizing}} \cite{ceritli2023synthesizing} applied TabDDPM to synthesize tabular healthcare data.

However, most existing work focuses on discrete or tabular data generation.
There is limited literature on EHR time series data generation, and this area of research has not yet received much attention \cite{Koo2023ACS}.
Back in 2017, RCGAN \cite{esteban2017realvalued} was created for generating multivariate medical time series data by employing RNNs as the generator and discriminator.
Until recently, \secondedit{\citeauthor{Yoon2023EHRSafeGH}} \cite{Yoon2023EHRSafeGH} proposed EHR-Safe that consists of a GAN and an encoder-decoder module.
EHR-Safe can generate realistic time series and static variables in EHR with mixed data types.
\secondedit{\citeauthor{li2023generating}} \cite{li2023generating} developed EHR-M-GAN that generates mixed-type time series in EHR using separate encoders for each data type. \secondedit{\citeauthor{Theodorou2023SynthesizeHL}} \cite{Theodorou2023SynthesizeHL} suggested generating longitudinal continuous EHR variables using an autoregressive language model.
Moreover, \secondedit{\citeauthor{Kuo2023SyntheticHL}} \cite{Kuo2023SyntheticHL} suggested utilizing diffusion models to synthesize discrete and continuous EHR time series.
However, their approach mainly relies on Gaussian diffusion and adopts a U-Net architecture \cite{ronneberger2015u}.
The generation of discrete time series is achieved by taking argmax of softmax over real-valued one-hot representations.
By contrast, our proposed method considers multinomial diffusion for discrete time series generation, allowing the generation of discrete variables directly.
\secondedit{\citeauthor{he2024flexible}}\firstedit{\cite{he2024flexible}, a concurrent work to ours, introduces FLEXGEN-EHR for synthesizing heterogeneous longitudinal EHR data through a latent diffusion method.
It also addresses missing modalities by formulating an optimal transport problem to create meaningful latent embedding pairs.
In comparison, our work introduces a direct diffusion model to generate heterogeneous EHR data, effectively handling potential missingness directly within the generation process.}

\section{Materials and Methods} \label{sec:method}
\subsection{Datasets}
We use four publicly available EHR datasets to evaluate \ours: Medical Information Mart for Intensive Care III and IV (MIMIC-III/IV) \cite{johnson2016mimic, Johnson2023MIMICIVAF} and the eICU Collaborative Research Database (eICU) \cite{Pollard2018TheEC}. %, and high time resolution ICU dataset (HiRID) \cite{hyland2020early}. 
Additionally, to evaluate \ours with state-of-the-art methods for time series generation on non-EHR datasets, we include Stocks and Energy datasets from studies that proposed TimeGAN \cite{yoon2019timegan} and GT-GAN \cite{jeon2022gtgan}.

\subsection{Metrics}
We evaluate our methods and make comparisons on a series of metrics, both qualitative and quantitative, characterizing the authenticity of the synthesized data, the \firstedit{performance} for downstream analysis -- in-hospital mortality prediction, and the preservation of privacy:
\subsubsection{Authenticity}
\begin{itemize}
    \item t-SNE visualization: We flatten the feature dimension and use t-SNE dimension reduction visualization \cite{van2008visualizing} on synthetic, real training, and real testing samples.
    This qualitative metric provides visual guidance on the similarity of the synthetic and real samples in two-dimensional space. Details are described in \ref{appendix:detail:evaluate:discripred}.

    \item \firstedit{UMAP visualization: We follow the same procedure as using t-SNE for visualization of distribution similarity between synthetic, real training, and real testing samples.
    UMAP preserves a better global structure compared to t-SNE \cite{McInnes2018UMAPUM}, and thus we provide it as a complementary metric.}
    
    \item Discriminative and Predictive Scores: A GRU-based discriminator is trained to distinguish between the synthetic and real samples.
    For the predictive score, a GRU-based predictor is trained using synthetic samples and evaluated on real samples for next-step vector prediction based on mean absolute error over each sequence.
    Details of the score computations are described in \ref{appendix:detail:evaluate:discripred}.
\end{itemize}

\subsubsection{\firstedit{Performance} for Downstream Task (in-hospital mortality prediction)}
\begin{itemize}
    \item Train on Synthetic, Test on Real (TSTR): We train ML models using synthetic data and evaluate them on real test data based on the area under the receiver operating characteristic curve (AUC) for in-hospital mortality prediction.
    We compare the TSTR score to the Train on Real, Test on Real (TRTR) score, which is the AUC obtained from the model trained on real training data and evaluated on real test data.
    \item Train on Synthetic and Real, Test on Real (TSRTR): Similar to the TSTR, we train ML models and evaluate them on real test data using AUC.
    We use 2,000 real training data in combination with different proportions of synthetic samples to train ML models.
    This metric evaluates the impact of synthetic data for training on ML model performance.
    \firstedit{Note that we use 2,000 real training samples to simulate the real-world scenario where \secondedit{clinical researchers} struggle to obtain limited real EHR data.
    In this case, we evaluate the viability of using \ours to generate realistic samples as a data augmentation technique.}
\end{itemize}

\subsubsection{Privacy}
\begin{itemize}
    \item Nearest Neighbor Adversarial Accuracy Risk (NNAA): This score measures the degree to which a generative model overfits the real training data \cite{yale2020generation}. \firstedit{NNAA is an important metric for evaluating the privacy of synthetic data as it quantifies the risk of re-identification by measuring how easily an adversary can distinguish between real and synthetic data points. Thus, this metric effectively indicates the potency of anonymization techniques in protecting sensitive information within the synthetic dataset.}
    \item Membership Inference Risk (MIR): An F1 score is computed based on whether an adversary can correctly identify the membership of a synthetic data sample \cite{liu2019socinf}. \firstedit{MIR provides a precise measurement of the security of synthetic datasets, particularly in assessing the likelihood that individual data points can be traced back to the original dataset, thereby evaluating the robustness of data anonymization techniques.}
\end{itemize}

For all the experiments, we split each dataset into training and testing sets and used the training set to develop generative models.
The synthetic samples obtained from trained generative models are then used for evaluation.
We repeat each experiment over 10 times and report the mean and standard deviation of each quantitative metric.
Further details for our experiments and evaluation metrics are discussed in \Cref{appendix:detail}.

\subsection{Baselines}
We compare \ours with \firstedit{nine} methods: \firstedit{HALO} \cite{Theodorou2023SynthesizeHL}, EHR-M-GAN \cite{li2023generating}, GT-GAN \cite{jeon2022gtgan}, TimeGAN \cite{yoon2019timegan}, RCGAN \cite{esteban2017realvalued}, C-RNN-GAN \cite{mogren2016crnngan}, RNNs trained with teacher forcing (T-Forcing) \cite{graves2013generating, sutskever2011generating} and professor forcing (P-Forcing) \cite{lamb2016professor}, and discrete or continuous stochastic process diffusion (DSPD/CSPD) with Gaussian (GP) or Ornstein-Uhlenbeck (OU) processes \cite{pmlr-v202-bilos23a}.\footnote{We also hoped to include comparison with EHR-Safe \cite{Yoon2023EHRSafeGH}. However, despite attempts, we were unable to obtain the code implementation.} In addition we compare with standard GRU and LSTM approach, with results in \Cref{result:pred discri scores}.

\subsection{Diffusion process on EHR time series \label{sec:method:diffusion}}
We first introduce our notations for the generation of both {\firstedit continuous-valued} and {\firstedit discrete-valued} time series in our framework, as both are present in EHR.
Specifically, let $\mathcal{D}$ denote our EHR time series dataset.
Each patient in $\mathcal{D}$ has continuous-valued and {\firstedit discrete-valued} multivariate time series $\mX \in \R^{P_{r} \times L}$ and $\mC \in \mathbb{Z}^{P_{d} \times L}$, respectively.
$L$ is the number of time steps, and $P_{r}$ and $P_{d}$ are the number of variables for continuous and discrete data types.

To generate both {\firstedit continuous-valued} and {\firstedit discrete-valued} time series, we consider a ``mixed sequence diffusion'' approach by adding Gaussian and multinomial noises.
For continuous-valued time series, we perform Gaussian diffusion by adding independent Gaussian noise similar to DDPM.
The forward process is thus defined as:
\begin{equation}
    q\big(\mX^{(1:T)} | \mX^{(0)}\big) = \prod_{t=1}^T \prod_{l=1}^L q\big(\mX_{\cdot, l}^{(t)} | \mX_{\cdot, l}^{(t-1)}\big),
\end{equation}
where $q(\mX_{\cdot, l}^{(t)} | \mX_{\cdot, l}^{(t-1)}) = \mathcal{N}(\mX_{\cdot, l}^{(t)}; \sqrt{1 - \beta^{(t)}}\mX_{\cdot, l}^{(t-1)}, \beta^{(t)}\mI )$ and $\mX_{\cdot, l}$ is the $l^{\text{th}}$ observation of the continuous-valued time series. In a similar fashion as \Cref{reverse}, we define the reverse process for continuous-valued features as $p_{\theta}(\mX^{(0:T)}) = p(\mX^{(T)}\big) \prod_{t=1}^T p_{\theta}\big(\mX^{(t-1)} | \mX^{(t)})$, where
\begin{align}
    p_{\theta}\big(\mX^{(t-1)} | \mX^{(t)} \big) &:= \mathcal{N}\big(\mX^{(t-1)}; \bm{\mu}_{\theta}(\mX^{(t)}, t),  \tilde{\beta}^{(t)}\mI \big), \notag \\
    \bm{\mu}_{\theta}(\mX^{(t)}, t) = \frac{1}{\sqrt{\alpha^{(t)}}}\bigg(\mX^{(t)} - &\frac{\beta^{(t)}}{\sqrt{1 - \bar{\alpha}^{(t)}}} \backbone(\mX^{(t)}, t)\bigg), \quad
    \tilde{\beta}^{(t)} = \frac{1 - \bar{\alpha}^{(t-1)}}{1 - \bar{\alpha}^{(t)}}\beta^{(t)}.
\end{align}
To model discrete-valued time series, we use multinomial diffusion \cite{NEURIPS2021_67d96d45}. The forward process is defined as:
\begin{align}
    q\big(\tilde{\mC}^{(1:T)} | \tilde{\mC}^{(0)}\big) &= \prod_{t=1}^T \prod_{p=1}^{P_d}\prod_{l=1}^L q\big(\tilde{\mC}_{p, l}^{(t)} | \tilde{\mC}_{p, l}^{(t-1)}\big), \\
    q\big(\tilde{\mC}_{p, l}^{(t)} | \tilde{\mC}_{p, l}^{(t-1)} \big) := \mathcal{C}\big(&\tilde{\mC}_{p, l}^{(t)} ; (1 - \beta^{(t)})\tilde{\mC}_{p, l}^{(t-1)} + \beta^{(t)} / K \big),
\end{align}
where $\mathcal{C}$ is a categorical distribution, $\tilde{\mC}_{p, l}^{(0)} \in \{0,1\}^K$ is a one-hot encoded representation of $C_{p, l}$\footnote{We perform one-hot encoding on the discrete-valued time series across the feature dimension. For example, if our time series is $\{0,1,2\}$, its one-hot representation becomes $\{[1,0,0]^\top, [0,1,0]^\top, [0,0,1]^\top\}$.}, and the addition and subtraction between scalars and vectors are performed element-wise.
The forward process posterior distribution is defined as follows\secondedit{, where $\odot$ represents the Hadamard product that returns a matrix with each element being the product of the corresponding elements from the original two matrices:}
\begin{align}
    q\big(\tilde{\mC}_{p,l}^{(t-1)} | \tilde{\mC}_{p,l}^{(t)}, \tilde{\mC}_{p,l}^{(0)} \big) &:= \mathcal{C}\left(\tilde{\mC}_{p,l}^{(t-1)} ; \bm{\phi}/\sum_{k=1}^K \phi_{k}\right), \\
    \bm{\phi} = \left(\alpha^{(t)}\tilde{\mC}_{p,l}^{(t)} + (1 - \alpha^{(t)}) / K \right) &\odot \left(\bar{\alpha}^{(t-1)}\tilde{\mC}_{p,l}^{(0)} + (1 - \bar{\alpha}^{(t-1)}) / K \right).
\end{align}
The reverse process $p_{\theta}(\tilde{\mC}_{p,l}^{(t-1)} | \tilde{\mC}_{p,l}^{(t)})$ is parameterized as $q\big(\tilde{\mC}_{p,l}^{(t-1)} | \tilde{\mC}_{p,l}^{(t)}, \backbone(\tilde{\mC}_{p,l}^{(t)}, t) \big)$.
We train our neural network, $\backbone$, using both Gaussian and multinomial diffusion processes:
\begin{align}
    \mathcal{L}_{\mathcal{N}}(\theta) &:= \E_{\mX^{(0)}, \bm{\epsilon}, t} \Bigg[ \Big\| \bm{\epsilon} - \backbone\big( \sqrt{\bar{\alpha}^{(t)}}\mX^{(0)} + \sqrt{1 - \bar{\alpha}^{(t)}}\bm{\epsilon}, t\big) \Big\|^2 \Bigg], \\
    \mathcal{L}_{\mathcal{C}}(\theta) &:= \E_{p,l} \Bigg[\sum_{t=2}^{T}\KL \Big(q\big(\tilde{\mC}_{p,l}^{(t-1)} | \tilde{\mC}_{p,l}^{(t)}, \tilde{\mC}_{p,l}^{(0)} \big) \;\Big\Vert\; p_{\theta}\big(\tilde{\mC}_{p,l}^{(t-1)} | \tilde{\mC}_{p,l}^{(t)}\big) \Big) \Bigg],
\end{align}
where $\mathcal{L}_{\mathcal{N}}$ and $\mathcal{L}_{\mathcal{C}}$ are the losses for continuous-valued and discrete-valued multivariate time series, respectively.
The training of the neural network is performed by minimizing the following loss:
\begin{equation}
    \mathcal{L}_{\text{train}}(\theta) = \lambda \mathcal{L}_{\mathcal{C}}(\theta) + \mathcal{L}_{\mathcal{N}}(\theta),
    \label{method:loss}
\end{equation}
where $\lambda$ is a hyperparameter for creating a balance between the two losses. We investigate the effects of $\lambda$ in \Cref{appendix:add:lambda}.

\subsection{Missing value representation \label{method:feature representation}}
In medical applications, missing data and variable measurement times play a crucial role as they could provide additional information and indicate a patient's health status \cite{zhou2023missing}.
We thus derive a missing indicator mask $\mM \in \{0,1\}^{P_r \times L}$ \footnote{Or alternatively, $\mM \in \{0,1\}^{P_d \times L}$ if the time series is discrete-valued.} for each $\mX \in \mathcal{D}$\footnote{For simplicity in writing, we refer to $\mX$ only, but this procedure can also be applied on $\mC$.}:
\begin{equation}\label{M}
    M_{p, l} = \begin{cases}
    0, & \text{if $X_{p, l}$ is present;} \\
    1, & \text{if $X_{p, l}$ is missing.}
    \end{cases}
\end{equation}
Then $\mM$ encodes the measurement time points of $\mX$.
If $\mX$ contains missing values, we impute them in the initial value of the forward process, i.e., $\mX^{(0)}$, using the corresponding sample mean. \footnote{\firstedit{Using the sample mean for imputation is a straightforward and computationally efficient method. It helps maintain the central tendencies and distributional characteristics of the original data, minimizing the introduction of biases that might occur with more complex methods \cite{little2019statistical,enders2022applied}.}}
Nevertheless, $\mM$ retains the information regarding the positions of missing values.
Our method generates discrete and continuous-valued time series, allowing us to seamlessly represent and generate $\mM$ as a discrete time series.

\subsection{\ours architecture}
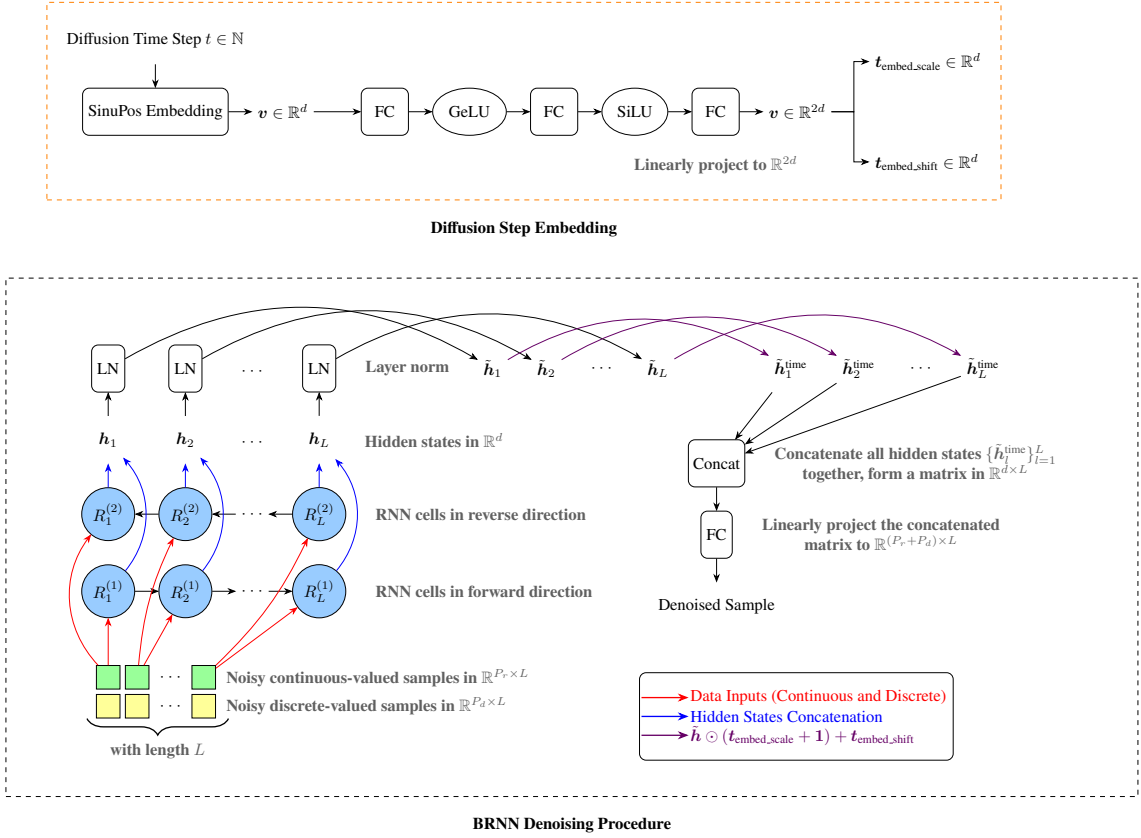
\begin{figure}[h]
\centering
\resizebox{\textwidth}{!}{%
\begin{tikzpicture}[
  every node/.style={draw, minimum height=1cm, align=center},
  >=Stealth
]

% colors
\definecolor{myyellow}{RGB}{255, 255, 153}
\definecolor{mygreen}{RGB}{153, 255, 153}
\definecolor{myrnn}{RGB}{153, 204, 255}
\definecolor{myshift}{RGB}{102, 0, 102}
\definecolor{mygrey}{RGB}{96, 96, 96}

% Nodes
\node (sine) at (0,0) [rectangle, rounded corners, minimum width=2cm] {SinuPos Embedding};
\node (t) [above=.5cm of sine, draw=none] {Diffusion Time Step $t \in \mathbb{N}$};
\node (vector1) [right=.5cm of sine, draw=none] {$\bm{v} \in \mathbb{R}^d$};
\node (fc1) [right=of vector1, rounded corners, rectangle, minimum width=1cm] {FC};
\node (gelu) [right=.5cm of fc1, ellipse] {GeLU};
\node (fc2) [right=.5cm of gelu, rounded corners, rectangle, minimum width=1cm] {FC};
\node (silu) [right=.5cm of fc2, ellipse] {SiLU};
\node (fc3) [right=.5cm of silu, rounded corners, rectangle, minimum width=1cm] {FC};
\node(textfc3) [draw=none, below=.1cm of fc3] {\textcolor{mygrey}{\textbf{Linearly project to} $\mathbb{R}^{2d}$}};
\node (vector2) [right=.5cm of fc3, draw=none] {$\bm{v} \in \mathbb{R}^{2d}$};
\node (scale) [above right=.05cm and .8cm of vector2, draw=none] {$\bm{t}_{\text{embed\_scale}} \in \mathbb{R}^d$};
\node (shift) [below right=.05cm and .8cm of vector2, draw=none] {$\bm{t}_{\text{embed\_shift}} \in \mathbb{R}^{d}$};
% Dotted rectangle around the entire drawing
\node (box) [fit=(t) (sine) (vector1) (fc1) (gelu) (fc2) (silu) (fc3) (vector2) (scale) (shift), draw, dashed, inner sep=.3cm, color=orange]{};
\node (text) [below=.1cm of box, draw=none]{\textbf{Diffusion Step Embedding}};

% Time-conditional BRNN
\node(sample1) at (-1,-12) [rectangle, minimum width=.5cm, minimum height=.5cm, fill=mygreen] {};
\node(sample2) [rectangle, right=.1cm of sample1, minimum width=.5cm, minimum height=.5cm, fill=mygreen] {};
\node(dots1) [draw=none, right=.1cm of sample2] {$\cdot \cdot \cdot$};
\node(sample3) [rectangle, right=.1cm of dots1, minimum width=.5cm, minimum height=.5cm, fill=mygreen] {};
\node(text real) [draw=none, right=.1cm of sample3] {\textcolor{mygrey}{\textbf{Noisy continuous-valued samples in }$\mathbb{R}^{P_r \times L}$}};
\node(sample4) [rectangle, below=.1cm of sample1, minimum width=.5cm, minimum height=.5cm, fill=myyellow] {};
\node(sample5) [rectangle, right=.1cm of sample4, minimum width=.5cm, minimum height=.5cm, fill=myyellow] {};
\node(dots2) [draw=none, right=.1cm of sample5] {$\cdot \cdot \cdot$};
\node(sample6) [rectangle, right=.1cm of dots2, minimum width=.5cm, minimum height=.5cm, fill=myyellow] {};
\node(text real) [draw=none, right=.1cm of sample6] {\textcolor{mygrey}{\textbf{Noisy discrete-valued samples in} $\mathbb{R}^{P_d \times L}$}};
\draw [decorate,decoration={brace=mygrey,amplitude=10pt,mirror,raise=5pt},thick] (-1.45, -12.7) -- (1.55, -12.7) node[midway, below=10pt, draw=none] {\textcolor{mygrey}{\textbf{with length} \(L\)}};
\node(rnn1) [circle, above=1cm of sample1, fill=myrnn] {$R^{(1)}_1$};
\node(rnn2) [circle, right=.5cm of rnn1, fill=myrnn] {$R^{(1)}_2$};
\node(dots3) [draw=none, right=.5cm of rnn2] {$\cdot \cdot \cdot$};
\node(rnn3) [circle, right=.5cm of dots3, fill=myrnn] {$R^{(1)}_L$};
\node(text rnn1) [draw=none, right=.5cm of rnn3] {\textcolor{mygrey}{\textbf{RNN cells in forward direction}}};
\node(rnn4) [circle, above=.5cm of rnn1, fill=myrnn] {$R^{(2)}_1$};
\node(rnn5) [circle, right=.5cm of rnn4, fill=myrnn] {$R^{(2)}_2$};
\node(dots4) [draw=none, right=.5cm of rnn5] {$\cdot \cdot \cdot$};
\node(rnn6) [circle, right=.5cm of dots4, fill=myrnn] {$R^{(2)}_L$};
\node(text rnn2) [draw=none, right=.5cm of rnn6] {\textcolor{mygrey}{\textbf{RNN cells in reverse direction}}};

\node(h1) [draw=none, above=.5cm of rnn4] {$\bm{h}_1$};
\node(h2) [draw=none, above=.5cm of rnn5] {$\bm{h}_2$};
\node(dots5) [draw=none, above=.5cm of dots4]{$\cdot \cdot \cdot$};
\node(h3) [draw=none, above=.5cm of rnn6]{$\bm{h}_L$};
\node(texth) [draw=none, right=.5cm of h3] {
\textcolor{mygrey}{\textbf{Hidden states in} $\mathbb{R}^d$}};

\node(l1) [rectangle, rounded corners, above=.5cm of h1] {LN};
\node(l2) [rectangle, rounded corners, above=.5cm of h2] {LN};
\node(dots6) [draw=none, above=.5cm of dots5] {$\cdot \cdot \cdot$};
\node(l3) [rectangle, rounded corners, above=.5cm of h3] {LN};
\node(textln) [draw=none, right=.5cm of l3] {\textcolor{mygrey}{\textbf{Layer norm}}};

\node(hl1) [draw=none, right=3cm of l3] {$\tilde{\bm{h}}_1$};
\node(hl2) [draw=none, right=.5cm of hl1] {$\tilde{\bm{h}}_2$};
\node(dots7) [draw=none, right=.5cm of hl2]{$\cdot \cdot \cdot$};
\node(hl3) [draw=none, right=.5cm of dots7] {$\tilde{\bm{h}}_L$};

\node(ht1) [draw=none, right=2cm of hl3] {$\tilde{\bm{h}}^{\text{time}}_1$};
\node(ht2) [draw=none, right=.5cm of ht1] {$\tilde{\bm{h}}^{\text{time}}_2$};
\node(dots8) [draw=none, right=.5cm of ht2]{$\cdot \cdot \cdot$};
\node(ht3) [draw=none, right=.5cm of dots8] {$\tilde{\bm{h}}^{\text{time}}_L$};

\node(concat) [rectangle, rounded corners, below left=1cm and .5cm of ht1, inner sep=.1cm] {Concat};
\node(fcfinal) [rectangle, rounded corners, below=.5cm of concat] {FC};
\node(output) [draw=none, below=.5cm of fcfinal]{Denoised Sample};
\node(textconcat) [draw=none, right=.5cm of concat]{\textcolor{mygrey}{\textbf{Concatenate all hidden states} $\{\tilde{\bm{h}}^{\text{time}}_l\}_{l=1}^L$} \\ \textcolor{mygrey}{\textbf{together, form a matrix in} $\mathbb{R}^{d \times L}$}};
\node(textfcfinal) [draw=none, right=.5cm of fcfinal]{\textcolor{mygrey}{\textbf{Linearly project the concatenated}} \\ \textcolor{mygrey}{\textbf{matrix to} $\mathbb{R}^{(P_r + P_d) \times L}$}};

% Legends
\node[draw=black, rounded corners, inner sep=.8pt, below right=.9cm and -3cm of output] (legend) {
    \begin{tikzpicture}[baseline]
        \draw[->, red, thick] (0, 0) -- (1, 0) node[anchor=west, black, draw=none] {\textcolor{red}{Data Inputs (Continuous and Discrete)}};
        \draw[->, blue, thick] (0, -0.4) -- (1, -0.4) node[anchor=west, black, draw=none] {\textcolor{blue}{Hidden States Concatenation}};
        \draw[->, myshift, thick] (0, -0.8) -- (1, -0.8) node[anchor=west, black, draw=none] {\textcolor{myshift}{$\tilde{\bm{h}} \odot (\bm{t}_{\text{embed\_scale}} + \bm{1}) + \bm{t}_{\text{embed\_shift}}$}};
    \end{tikzpicture}
};
\node(box2)[fit=(sample1)(sample2)(dots1)(sample3)(text real)(sample4)(sample5)(dots2)(sample6)(text real)(rnn1)(rnn2)(dots3)(rnn3)(text rnn1)(rnn4)(rnn5)(dots4)(rnn6)(text rnn2)(h1)(h2)(dots5)(h3)(texth)(l1)(l2)(dots6)(l3)(hl1)(hl2)(dots7)(hl3)(ht1)(ht2)(dots8)(ht3)(concat)(fcfinal)(output)(textconcat)(textfcfinal), draw, dashed, minimum height=11cm, minimum width=24cm]{};
\node (text2) [below=.1cm of box2, draw=none]{\textbf{BRNN Denoising Procedure}};

% Arrows
\draw[->] (t) -- (sine);
\draw[->] (sine) -- (vector1);
\draw[->] (vector1) -- (fc1);
\draw[->] (fc1) -- (gelu);
\draw[->] (gelu) -- (fc2);
\draw[->] (fc2) -- (silu);
\draw[->] (silu) -- (fc3);
\draw[->] (fc3) -- (vector2);
\draw[->] (vector2) -- ++(1.2cm,0) |- (scale);
\draw[->] (vector2) -- ++(1.2cm,0) |- (shift);
\draw[->] (rnn1) -- (rnn2);
\draw[->] (rnn2) -- (dots3);
\draw[->] (dots3) -- (rnn3);
\draw[->] (rnn5) -- (rnn4);
\draw[->] (dots4) -- (rnn5);
\draw[->] (rnn6) -- (dots4);
\draw[->, red] (sample1) -- (rnn1);
\draw[->, red, bend left=40] (sample1) to (rnn4);
\draw[->, red] (sample2) to (rnn2);
\draw[->, red, bend left=10] (sample2) to (rnn5);
\draw[->, red] (sample3) to (rnn3);
\draw[->, red, bend right=10] (sample3) to (rnn6);
\draw[->, blue] (rnn4) to (h1);
\draw[->, blue, bend right=40] (rnn1) to (h1);
\draw[->, blue] (rnn5) to (h2);
\draw[->, blue, bend right=40] (rnn2) to (h2);
\draw[->, blue] (rnn6) to (h3);
\draw[->, blue, bend right=40] (rnn3) to (h3);
\draw[->] (h1) to (l1);
\draw[->] (h2) to (l2);
\draw[->] (h3) to (l3);
\draw[->, bend left=30] (l1) to (hl1);
\draw[->, bend left=30] (l2) to (hl2);
\draw[->, bend left=30] (l3) to (hl3);
\draw[->, myshift, bend left=30] (hl1) to (ht1);
\draw[->, myshift, bend left=30] (hl2) to (ht2);
\draw[->, myshift, bend left=30] (hl3) to (ht3);
\draw[->] (ht1) to (concat);
\draw[->] (ht2) to (concat);
\draw[->] (ht3) to (concat);
\draw[->] (concat) to (fcfinal);
\draw[->] (fcfinal) to (output);

\end{tikzpicture}
}
\caption{\firstedit{Visualization of \ours architecture.} \secondedit{FC represents a fully connected layer, SiLU is sigmoid linear unit activation, SinuPos Embedding is a shorthand for sinusoidal positional embedding, and GeLU is Gaussian error linear unit activation.}}
\label{architecture diagram}
\end{figure}
In this section, we describe our architecture for the diffusion model.
A commonly used architecture in DDPM is U-Net \cite{ronneberger2015u}.
However, most U-Net-based models are tailored to image generation tasks, requiring the neural network to process pixel-based data rather than sequential information \cite{song2020score, ho2020denoising, rombach2022high}.
Even its one-dimensional variant, 1D-U-Net, comes with limitations such as restriction on the input sequence length (which must be a multiple of U-Net multipliers) and a tendency to lose temporal dynamics information during down-sampling.
On the other hand, TabDDPM \cite{Kotelnikov2022TabDDPMMT} proposed a mixed diffusion approach for tabular data generation but relied on a multilayer perceptron architecture, making it improper for multivariate time series generation.

To address this challenge of handling EHR time series, we need an architecture capable of encoding sequential information while being flexible to the input sequence length.
The time-conditional bidirectional RNN (BRNN) or neural controlled differential equation (NCDE) \cite{kidger2020neural} can be possible options.
After careful evaluation, we found that BRNN without attention mechanism offers superior computational efficiency and have chosen it as the neural backbone $\backbone$ for all of our experiments.
A more detailed discussion of NCDE is provided in Supplementary Material \ref{appendix:detail:hyperparams:ncde}.

\paragraph{Diffusion Step Embedding:} 
To inform the model about the current diffusion time step $t$, we use sinusoidal positional embedding \cite{vaswani2017attention}.
The embedding vector output from the embedding layer then goes through two fully connected (FC) layers with GeLU activation in between \cite{hendrycks2016gaussian}.
The embedding vector is then fed to a SiLU activation \cite{hendrycks2016gaussian} and another FC layer.
The purpose of this additional FC layer is to adjust the dimensionality of the embedding vector to match the stacked hidden states from BRNN.
Specifically, we set the dimensionality of the output to be two times the size of the hidden dimension from BRNN.
We denote the transformed embedding vector as $\vt_{\text{embed}}$.
This vector is then split into two vectors, each with half of the current size, namely $\vt_{\text{embed\_scale}}$ and $\vt_{\text{embed\_shift}}$.
Both vectors share the same dimensionality as BRNN's hidden states and serve to inform the network about the current diffusion time step.

\paragraph{Time-conditional BRNN:} 
In practice, BRNN can be implemented with either LSTM or GRU units.
To condition BRNN on time, we follow these steps.
We first obtain noisy samples from Gaussian (for continuous-valued data) and multinomial (for discrete-valued data) diffusion.
The two samples are concatenated and fed to our BRNN, which returns a sequence of hidden states $\{\vh_l\}_{l=1}^L$ that stores the temporal dynamics information about the time series.
To stabilize learning and enable proper utilization of $\vt_{\text{embed}}$, we apply layernorm \cite{ba2016layer} on $\{\vh_l\}_{l=1}^L$.
The normalized sequence of hidden states, $\{\tilde{\bm{h}}_l\}_{l=1}^L$, is then scaled and shifted using $\{\tilde{\bm{h}}_l \odot (\vt_{\text{embed\_scale}} + \bm{1}) + \vt_{\text{embed\_shift}}\}_{l=1}^L$.
These scaled hidden states contain information about the current diffusion step $t$, which is then passed through an FC layer to produce the final output.
The output contains predictions for both multinomial and Gaussian diffusions, which are extracted correspondingly and used to calculate $\mathcal{L}_{\text{train}}$ in \Cref{method:loss}.
\firstedit{A visual demonstration of our architecture is shown in \Cref{architecture diagram}, where the use of BRNN allows the denoising of noisy time series samples of arbitrary length $L$, and the diffusion step embedding is utilized to inform the model about the stage of the reverse diffusion process.}

\section{Results \label{sec:results}}
\subsection{Authenticity of the generated EHR time series} \label{results:authenticity}

\begin{table}[hbt!]
\centering
% \vspace{-8pt}
\caption{Predictive and discriminative scores of \ours and the baselines.}
\begin{adjustbox}{width=\textwidth}
\begin{tabular}{@{}clccccc@{}}
    \toprule
    Metric & Method & Stocks & Energy & MIMIC-III & MIMIC-IV  & eICU \\
    \midrule
    & \textbf{\ours} & \textbf{.048$\pm$.028} & \textbf{.088$\pm$.018} & \textbf{.028$\pm$.023} & \textbf{.030$\pm$.022} &  \textbf{.015$\pm$.007} \\
    & EHR-M-GAN & .483$\pm$.027 & .497$\pm$.006 & .499$\pm$.002 & .499$\pm$.001 & .488$\pm$.022\\
    & DSPD-GP & .081$\pm$.034 & .416$\pm$.016 & .491$\pm$.002 & .478$\pm$.020 & .327$\pm$.020 \\
    & DSPD-OU & .098$\pm$.030 & .290$\pm$.010 & .456$\pm$.014 & .444$\pm$.037 & .367$\pm$.018 \\
    & CSPD-GP & .313$\pm$.061 & .392$\pm$.007 & .498$\pm$.001 & .488$\pm$.010 & .489$\pm$.010 \\
    & CSPD-OU & .283$\pm$.039 & .384$\pm$.012 & .494$\pm$.002 & .479$\pm$.005  & .479$\pm$.017 \\
    % \addlinespace
    Discriminative & GT-GAN & .077$\pm$.031 & .221$\pm$.068 & .488$\pm$.026 & .472$\pm$.014 & .448$\pm$.043 \\
    Score & TimeGAN  & .102$\pm$.021 & .236$\pm$.012 & .473$\pm$.019 & .452$\pm$.027 & .434$\pm$.061\\
    ($\downarrow$) & RCGAN & .196$\pm$.027 & .336$\pm$.017 & .498$\pm$.001 & .490$\pm$.003 & .490$\pm$.023 \\
    & C-RNN-GAN & .399$\pm$.028 & .499$\pm$.001 & .500$\pm$.000 & .499$\pm$.000 & .493$\pm$.010 \\
    & T-Forcing & .226$\pm$.035 & .483$\pm$.004 & .499$\pm$.001 & .497$\pm$.002 & .479$\pm$.011\\
    & P-Forcing & .257$\pm$.026 & .412$\pm$.006 & .494$\pm$.006 & .498$\pm$.002 & .367$\pm$.047\\
    & \firstedit{HALO} & \firstedit{.491$\pm$.006} & \firstedit{.500$\pm$.000} & \firstedit{.497$\pm$.003} & \firstedit{.494$\pm$.004} &  \firstedit{.370$\pm$.074} \\
    & \textit{Real Data} & \textit{.019$\pm$.016} & \textit{.016$\pm$.006} & \textit{.012$\pm$.006} & \textit{.014$\pm$.011} &  \textit{.004$\pm$.003} \\
    \midrule
    & \textbf{\ours} & \textbf{.037$\pm$.000}  & \textbf{.251$\pm$.000} & \textbf{.469$\pm$.003} & \textbf{.432$\pm$.002} &  \textbf{.309$\pm$.019} \\
    & EHR-M-GAN & .120$\pm$.047 & .254$\pm$.001 & .861$\pm$.072 & .880$\pm$.079 & .913$\pm$.179\\
    & DSPD-GP & .038$\pm$.000 & .260$\pm$.001 & .509$\pm$.014 & .586$\pm$.026 & .320$\pm$.018 \\
    & DSPD-OU & .039$\pm$.000 & .252$\pm$.000 & .497$\pm$.006 & .474$\pm$.023 & .317$\pm$.023 \\
    & CSPD-GP & .041$\pm$.000 & .257$\pm$.001 & 1.083$\pm$.002 & .496$\pm$.034 & .624$\pm$.066 \\
    & CSPD-OU & .044$\pm$.000 & .253$\pm$.000 & .566$\pm$.006 & .516$\pm$.051 & .382$\pm$.026 \\
    % \addlinespace
    Predictive & GT-GAN & .040$\pm$.000 & .312$\pm$.002 & .584$\pm$.010 & .517$\pm$.016 & .487$\pm$.033 \\
    Score & TimeGAN & .038$\pm$.001 & .273$\pm$.004 & .727$\pm$.010 & .548$\pm$.022 & .367$\pm$.025 \\
    ($\downarrow$) & RCGAN & .040$\pm$.001 & .292$\pm$.005 & .837$\pm$.040 & .700$\pm$.014 & .890$\pm$.017 \\
    & C-RNN-GAN & .038$\pm$.000 & .483$\pm$.005 & .933$\pm$.046 & .811$\pm$.048 & .769$\pm$.045 \\
    & T-Forcing & .038$\pm$.001 & .315$\pm$.005 & .840$\pm$.013 & .641$\pm$.017 & 
    .547$\pm$.069\\
    & P-Forcing & .043$\pm$.001 & .303$\pm$.006 & .683$\pm$.031 & .557$\pm$.030 & .345$\pm$.021 \\
    & \firstedit{HALO} & \firstedit{.042$\pm$.006} & \firstedit{.299$\pm$.053} & \firstedit{.816$\pm$.020} & \firstedit{.767$\pm$.012}  & \firstedit{.378$\pm$.038} \\
    & \textit{Real Data} & \textit{.036$\pm$.001} & \textit{.250$\pm$.003}  & \textit{.467$\pm$.005} & \textit{.433$\pm$.001} & \textit{.304$\pm$.017} \\
    \bottomrule
\end{tabular}
\end{adjustbox}
% \vspace{-12pt}
\label{result:pred discri scores}
\end{table}

\begin{figure}[hbt!]
  \centering
  \includegraphics[width=\linewidth]{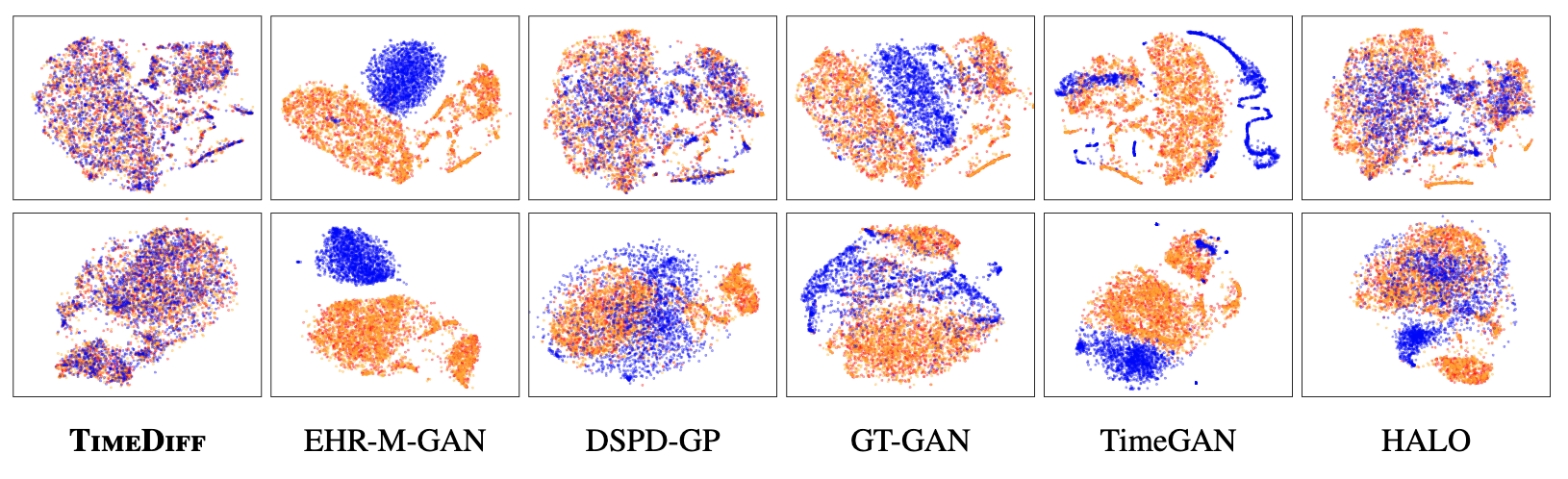}
  % \includegraphics[width=\linewidth]{pics/main_paper_t-sne_unmarked.png}
  % \vspace{-14pt}
  \caption{\firstedit{t-SNE visualization of the eICU ($1^{\text{st}}$ row) and the MIMIC-IV ($2^{\text{rd}}$ row) datasets. Synthetic samples in \textcolor{blue}{\textbf{blue}}, real training samples in \textcolor{red}{\textbf{red}}, and real testing samples in \textcolor{orange}{\textbf{orange}}.} \secondedit{We observe that there is a significant overlap between synthetic samples from \ours and real testing samples, suggesting \ours produces realistic synthetic EHR data. DSPD-GP and HALO also yield noticeable overlap.}}
  \label{result:tsne}
  % \vspace{-12pt}
\end{figure}

\begin{figure}[htb!]
    \includegraphics[width=\textwidth]{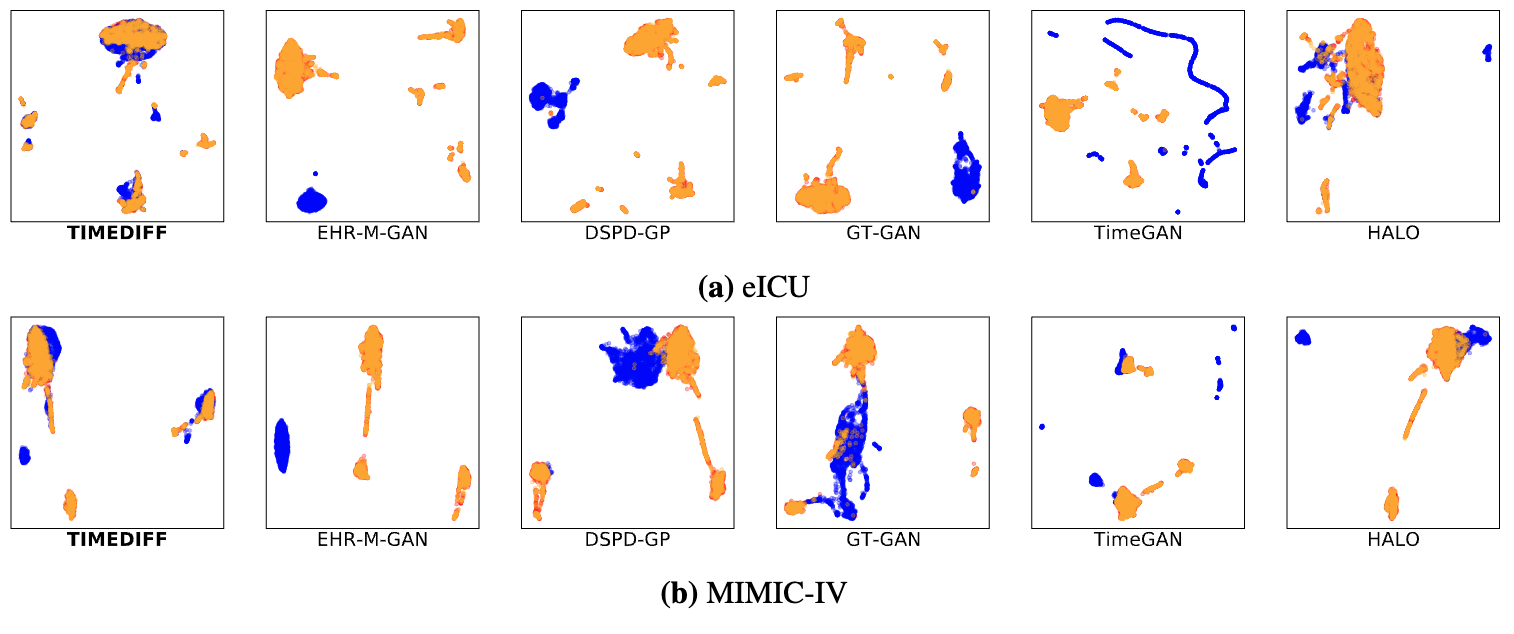}
    \caption{\firstedit{UMAP visualization of the eICU and the MIMIC-IV datasets. Synthetic samples in \textcolor{blue}{\textbf{blue}}, real training samples in \textcolor{red}{\textbf{red}}, and real testing samples in \textcolor{orange}{\textbf{orange}}.} \secondedit{We observe a similar result as the t-SNE visualizations, where there is an overlap between synthetic and real testing samples for \ours.
    The overlap for other models is less significant.}}
\label{result:umap}
\end{figure}

We evaluate the authenticity of the generated \secondedit{synthetic} EHR time series both qualitatively and quantitatively.
We provide a visualization of the distributions of the synthetic and real data using t-SNE \firstedit{, following \cite{Yoon2023EHRSafeGH}}, shown in \Cref{result:tsne}.
\firstedit{Additionally, we present another visualization metric using UMAP in \Cref{result:umap}.
Both visualization methods indicate the synthesized data generated from \ours overlaps with real training and test data, suggesting \ours can generate more realistic data compared to other baselines.}
Visualizations of the raw synthetic and real data per feature are presented in \Cref{appendix:add:visual}.
\firstedit{Note that t-SNE and UMAP are only for qualitative evaluation and are not precise. We next present quantitative metrics for precise evaluation.}
By comparing the predictive and discriminative scores in \Cref{result:pred discri scores}, we observe that \ours yields significantly lower scores than all the baseline methods across six datasets.
For instance, \ours yields a 95.4\% lower mean discriminative score compared to DSPD-GP and obtains a 1.6\% higher mean predictive score than real testing data on the eICU dataset.
For non-EHR datasets, \ours achieves a 37.7\% lower and a 60.2\% lower mean discriminative scores on the Stocks and Energy datasets than GT-GAN while having similar mean predictive scores as using real testing data.

\subsection{Data \firstedit{performance} on in-hospital mortality prediction} \label{results:mortality}

\begin{figure}[hbt!]
    \centering
    %\vspace{-15pt}
    \includegraphics[width=\linewidth]{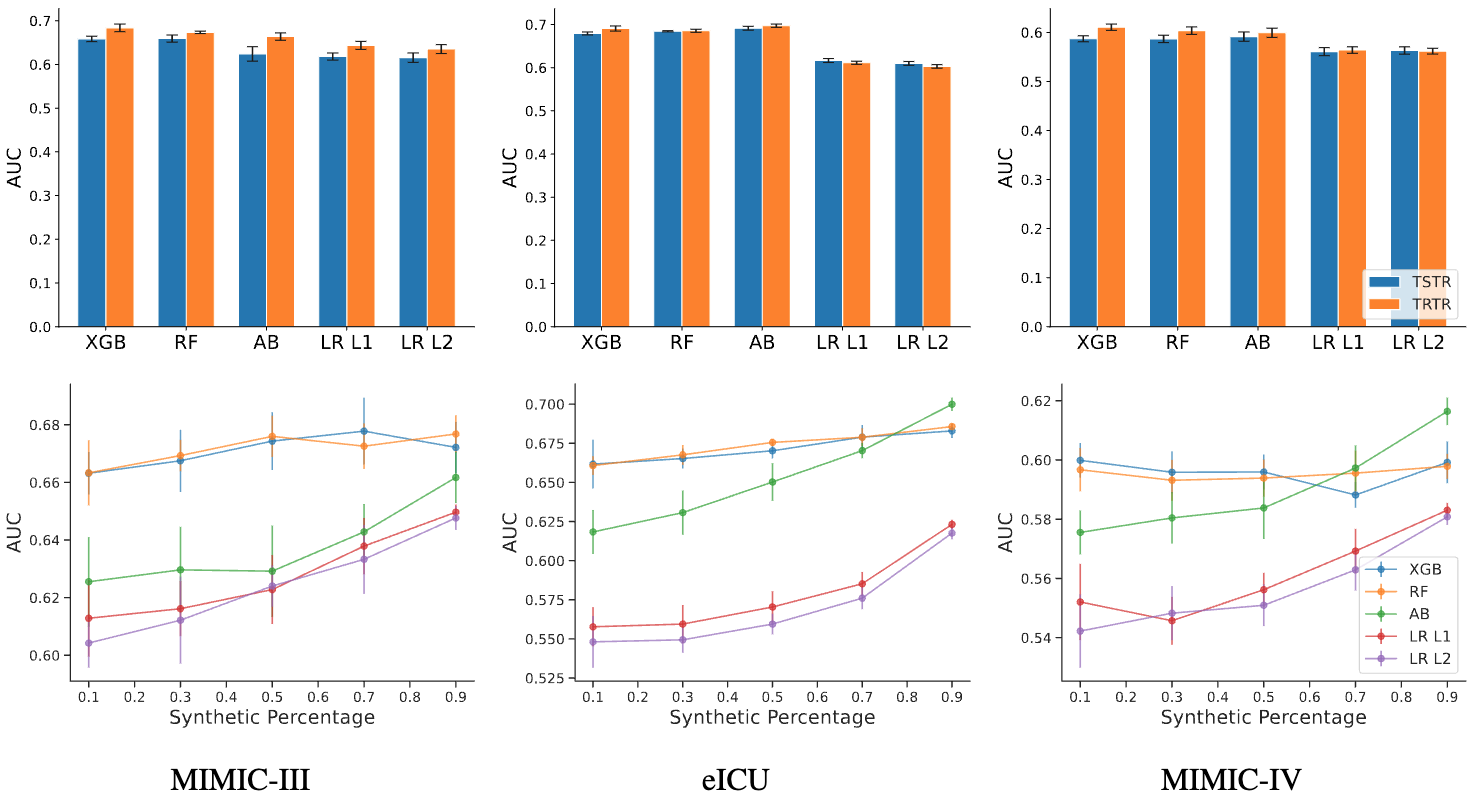}
    %\vspace{-18pt}
    \caption{TSTR scores compared to TRTR scores (Top); TSRTR scores (Bottom).} 
    \label{result:tstr}
    %\vspace{-25pt}
\end{figure}

We evaluate the data \firstedit{performance} of the generated \secondedit{synthetic} EHR time series on one common downstream task: in-hospital mortality prediction \cite{sadeghi2018early, sheikhalishahi2019benchmarking}.
We use six ML algorithms: XGBoost (XGB) \cite{chen2016xgboost}, Random Forest (RF) \cite{breiman2001random}, AdaBoost (AB) \cite{freund1997decision}, and $\ell_1$ and $\ell_2$ regularized Logistic Regression (LR L1/L2) \cite{friedman2010regularization}.
Additionally, to simulate the practical scenario where synthetic samples are used for data augmentation, we compute the TSRTR score for each ML model.
The prediction models are trained using synthetic samples from \ours and assessed on real test data.

From \Cref{result:tstr}, we observe that the TSTR scores obtained from models trained using synthetic EHR time series are close to the TRTR scores yielded from models trained using real data. We also notice a non-decreasing trend in the TSRTR scores as the percentage of synthetic EHR data increases for ML model training.
\firstedit{Additional TSTR and TSRTR evaluations for all baseline generative models can be found in Supplementary Material \ref{tstr_tsrtr:baselines}}.

\firstedit{Note that in addition to the six ML classifiers mentioned above, we utilize GRU and LSTM for prediction due to their ability in handling sequential data. We present the TSTR and TRTR scores obtained from RNNs in Supplementary Material \ref{appendix:add:tstr:rnn}, \Cref{appendix:add:tstr:rnn_table}. We observe that they achieve lower scores compared to the conventional classifiers.}

\subsection{Data privacy of \secondedit{sythetic EHR time series}}

\begin{table}[hbt!]
    \caption{Privacy scores of the synthesized EHR time series yielded from \ours and the baseline methods.}
    %\vspace{-10pt}
    \centering
    \begin{tabular}{@{}llccc@{}}
    \toprule
    Metric & Method & MIMIC-III & MIMIC-IV & eICU \\
    \midrule
    $\aatest$ ($\sim$0.5) & \textbf{\ours} & \textbf{.574$\pm$.002} & \textbf{.517$\pm$.002}  &  \textbf{.537$\pm$.001} \\
    & EHR-M-GAN & .998$\pm$.000 & 1.000$\pm$.000 & .977$\pm$.000 \\
    & DSPD-GP & .974$\pm$.001 & .621$\pm$.002 & .888$\pm$.000 \\
    & DSPD-OU & .927$\pm$.000 & .804$\pm$.003 & .971$\pm$.000\\
    & CSPD-GP & .944$\pm$.001 & .623$\pm$.002 & .851$\pm$.001\\
    & CSPD-OU & .967$\pm$.001 & .875$\pm$.002 & .982$\pm$.000\\
    & GT-GAN & .995$\pm$.000 & .910$\pm$.001 & .981$\pm$.000 \\
    & TimeGAN & .997$\pm$.000 & .974$\pm$.001 & 1.000$\pm$.000\\
    & RCGAN & .983$\pm$.001 & .999$\pm$.000 & 1.000$\pm$.000\\
    & \firstedit{HALO} & \firstedit{.698$\pm$.002} & \firstedit{.709$\pm$.002} &  \firstedit{.653$\pm$.001} \\
    & \textit{Real Data} & \textit{.552$\pm$.002} & \textit{.497$\pm$.002} &  \textit{.501$\pm$.002} \\
    \midrule
    $\aatrain$ ($\sim$0.5) & \textbf{\ours} & \textbf{.573$\pm$.002} & \textbf{.515$\pm$.002} &  \textbf{.531$\pm$.002} \\
    & EHR-M-GAN & .999$\pm$.000 & 1.000$\pm$.000 & .965$\pm$.002\\
    & DSPD-GP & .968$\pm$.002 & .620$\pm$.003 & .888$\pm$.001 \\
    & DSPD-OU & .928$\pm$.001 & .788$\pm$.003 & .971$\pm$.000\\
    & CSPD-GP & .940$\pm$.002 & .629$\pm$.005 & .852$\pm$.001\\
    & CSPD-OU & .966$\pm$.001 & .880$\pm$.003 & .983$\pm$.000\\
    & GT-GAN & .995$\pm$.001 & .907$\pm$.002 & .981$\pm$.000 \\
    & TimeGAN & .997$\pm$.000 & .969$\pm$.003 & 1.000$\pm$.000 \\
    & RCGAN & .984$\pm$.001 & .999$\pm$.000 & 1.000$\pm$.000\\
    & \firstedit{HALO} & \firstedit{.696$\pm$.001} & \firstedit{.717$\pm$.002} &  \firstedit{.653$\pm$.002} \\
    & \textit{Real Data} & \textit{.286$\pm$.003} & \textit{.268$\pm$.004} &  \textit{.266$\pm$.002} \\
    \midrule
    NNAA ($\downarrow$) & \textbf{\ours} & .002$\pm$.002 & .002$\pm$.002 & .006$\pm$.002 \\
    & EHR-M-GAN & .000$\pm$.000 & .000$\pm$.000 & .012$\pm$.003\\
    & DSPD-GP & .005$\pm$.003 & .003$\pm$.003 & .001$\pm$.001\\
    & DSPD-OU & .001$\pm$.001 & .016$\pm$.004 & .000$\pm$.000\\
    & CSPD-GP & .004$\pm$.002 & .007$\pm$.005 & .001$\pm$.001\\
    & CSPD-OU & .001$\pm$.001 & .005$\pm$.003 & .001$\pm$.001\\
    & GT-GAN & .001$\pm$.000 & .004$\pm$.002 & .000$\pm$.000 \\
    & TimeGAN & .000$\pm$.000 & .005$\pm$.003 & .000$\pm$.000 \\
    & RCGAN & .001$\pm$.000 & .000$\pm$.000 & .000$\pm$.000\\
    & \firstedit{HALO} & \firstedit{.002$\pm$.002} & \firstedit{.008$\pm$.002} &  \firstedit{.002$\pm$.001} \\
    & \textit{Real Data} & \textit{.267$\pm$.004} & \textit{.229$\pm$.003} &  \textit{.235$\pm$.003} \\
    \midrule
    MIR ($\downarrow$) & \textbf{\ours} & .191$\pm$.008 & .232$\pm$.048 & .227$\pm$.021 \\
    & EHR-M-GAN & .025$\pm$.007 & .435$\pm$.031 & .049$\pm$.006 \\
    & DSPD-GP & .032$\pm$.021 & .050$\pm$.009 & 
    .000$\pm$.000 \\
    & DSPD-OU & .060$\pm$.032 & .007$\pm$.006 & .000$\pm$.000 \\
    & CSPD-GP & .060$\pm$.028 & .034$\pm$.017 & .000$\pm$.000 \\
    & CSPD-OU & .066$\pm$.046 & .016$\pm$.020 & .000$\pm$.000 \\
    & GT-GAN & .005$\pm$.002 & .046$\pm$.013 & .000$\pm$.000 \\
    & TimeGAN & .010$\pm$.002 & .173$\pm$.020 & .000$\pm$.000 \\
    & RCGAN & .013$\pm$.002 & .277$\pm$.049 & .000$\pm$.000 \\
    & \firstedit{HALO} & \firstedit{.189$\pm$.007} & \firstedit{.019$\pm$.012} &  \firstedit{.036$\pm$.040} \\
    & \textit{Real Data} & \textit{.948$\pm$.000} & \textit{.929$\pm$.005} &  \textit{.927$\pm$.001}\\
    \bottomrule
\end{tabular}
    \label{result:privacy}
\end{table}

We also assess the risks of the generated \secondedit{synthetic} EHR time series being attacked by malicious entities using the NNAA and the MIR scores.
\secondedit{These metrics allow us to evaluate whether our approach can produce privacy-preserving synthetic EHR samples.}
As presented in \Cref{result:privacy}, we observe that \ours yields the $\aatest$ and $\aatrain$ scores around 0.5 across all four EHR datasets. \ours also obtains low NNAA and MIR scores compared to baseline methods.
Note that the full results are presented in Supplementary Material \ref{appendix:add:tstr}.

\subsection{Model runtime comparison} \label{results:runtime}

%\begin{wraptable}{r}{0.5\textwidth}
\begin{table}[hbt!]
    %\vspace{+2pt}
    \caption{Runtime comparisons between the \ours and baseline methods (hours).}
    %\vspace{-10pt}
    \centering
    \small
    \begin{adjustbox}{width=0.5\textwidth}
\begin{tabular}{@{}lllll@{}}
    \toprule
    Dataset & \textbf{\ours} & EHR-M-GAN & TimeGAN & GT-GAN \\
    \midrule
    MIMIC-III & \textbf{2.7} & 18.9 & 10.8 & 21.8 \\
    MIMIC-IV & \textbf{2.7} & 28.8 & 29.5 & 47.3 \\
    %HiRID & \textbf{2.5} & 29.7 & 46.2 & 58.3 \\
    eICU & \textbf{8.7} & 87.1 & 110 & 59.1 \\
    \bottomrule
\end{tabular}
\end{adjustbox}
    \label{result:time and size}
    %\vspace{-10pt}
\end{table}
%\end{wraptable}

We compare the number of hours to train \ours with EHR-M-GAN, TimeGAN, and GT-GAN presented in \Cref{result:time and size}.
As shown in \Cref{result:time and size}, \ours requires less training time compared to GAN-based approaches.

\firstedit{Note that in our experiments, to demonstrate that our proposed approach outperforms GANs in terms of training time, we primarily compared the training time of our proposed diffusion model with GANs. Most of the GANs primarily involve pre-training the embedding layer and subsequently training with adversarial feedback. This staged procedure made GANs more computationally heavy to train than diffusion models (which only requires optimizing one loss function for one neural network in our proposed approach).}

\subsection{Ablation study}

\begin{table}[hbt!]
    %\vspace{-10pt}
    \caption{Ablation study on generating missing indicators using multinomial diffusion.}
    %\vspace{-10pt}
    \centering
    \begin{tabular}{@{}llccc@{}}
    \toprule
    Metric & Method & MIMIC-III & MIMIC-IV & eICU \\
    \midrule
    & with Gaussian and rounding & .355$\pm$.020 & .121$\pm$.025 & .030$\pm$.018 \\
    Discriminative Score ($\downarrow$) & with Gaussian and softmax & .088$\pm$.023 & .155 $\pm$.032 & .042$\pm$.045\\
    & with multinomial & \textbf{.028$\pm$.023} & \textbf{.030$\pm$.022} & \textbf{.015$\pm$.007} \\
    \midrule
    & with Gaussian and rounding & .486$\pm$.005 & .433$\pm$.003 & .312$\pm$.031 \\
    Predictive Score ($\downarrow$) & with Gaussian and softmax & .472$\pm$.004 & .434$\pm$.002 & .320$\pm$.035 \\
    & with multinomial & \textbf{.469$\pm$.003} & \textbf{.432$\pm$.002} & \textbf{.309$\pm$.019} \\
    \bottomrule
\end{tabular}
    \label{result:ablation}
\end{table}

We further investigate the effect of utilizing multinomial diffusion in \ours on missing indicators for EHR discrete sequence generation.
We compare it with \ours using Gaussian diffusion, with the following two methods applied to the resulting output as transformations to discrete sequences:
(1) direct rounding; (2) applying argmax to the softmax output of real-valued, one-hot encoded representations.\footnote{The synthetic one-hot encoding is not discrete since we use Gaussian diffusion. This method is also adopted by \cite{Kuo2023SyntheticHL} for generating discrete time series with diffusion models.}

We present the discriminative and predictive scores obtained using the aforementioned methods on the MIMIC-III/IV and the eICU datasets in \Cref{result:ablation}.
%HiRID is excluded from this ablation study since it does not contain missing values.
We notice that \ours using multinomial diffusion obtains lower discriminative and predictive scores across all three datasets.

\section{Discussion \label{sec:discuss}}
% Authenticity
The synthetic samples generated by \ours exhibit remarkable overlap with real training and testing data (see \Cref{result:tsne}), indicating that the generated samples preserve similar data distribution to real data.
Note that we obtain the same observation across all datasets, with the rest of the visualizations presented in Supplementary Material \ref{appendix:add:tsne}.

The discriminative and predictive scores in \Cref{result:pred discri scores} suggest that the synthetic time series generated by \ours has the closest data distribution to real data compared to samples generated by other baseline methods.
We notice that the DSPD/CSPD baseline method from a recent work does not yield good performance on EHR datasets.
This observation can be attributed to its temporal modeling, which treats time series as discrete realizations of an underlying continuous process.
This continuity assumption may not hold for EHR time series data, which are highly discontinuous.

Additionally, \Cref{result:ablation} from the ablation study indicates that the synthesized EHR time series is more realistic when using multinomial diffusion in \ours.
\firstedit{Evaluation using TSTR and TSRTR metrics is also performed to compare \ours with the ``with Gaussian and softmax'' alternative, where we observe that \ours outperforms the alternative.
The result is presented in Supplementary Material \ref{tstr_tsrtr:baselines}, \Cref{gauss softmax tstr} and \Cref{gauss softmax tsrtr}.}

%Utility
We observe from \Cref{result:tstr} that models trained using synthetic time series yield similar AUC scores compared to those trained on the real data, indicating that the synthetic EHR time series obtained from \ours maintains high data utility for performing downstream tasks.
Additionally, we notice that most of the ML models yield increasing AUC scores with the increase in the number of synthetic samples added to model training.
This observation is consistent with our previous findings, indicating the high utility of our synthetic data.

%Privacy
The close to 0.5 scores of $\aatest$ and $\aatrain$ computed from \ours shown in \Cref{result:privacy} suggest that \ours generates high-fidelity synthetic time series and does not overfit its training data.
By contrast, although most of the baseline methods have low NNAA and MIR scores, they all have higher $\aatest$ and $\aatrain$ scores, which implies that there may be overfitting on the training data for baseline methods.

\firstedit{Lastly, to assess the effects of EHR time series generation, most features selected in our study are frequent measurements such as vital signs.
This design choice enables us to evaluate the ability of \ours to generate sequential measurements without interference from measurement frequencies.
Thus, our study does not focus on infrequent time series measurements or static features.
Nevertheless, we acknowledge that this is a limitation in our study and have conducted additional experiments on the ability of \ours to generate static and infrequent measurements.
The results can be found in Supplementary Material \ref{appendix:infrequent static}.}

\FloatBarrier
% \vspace{-10pt}
\section{Conclusion \label{sec:conclusion}}
%\vspace{-2pt}
We propose \ours for \secondedit{synthetic} EHR time series generation by using mixed sequence diffusion and demonstrate its superior performance compared with all state-of-the-art time series generation methods in terms of data utility.
We also demonstrate that \ours can facilitate downstream analysis in healthcare while protect patient privacy.
\secondedit{Thus, we believe \ours could be a useful tool to support medical data analysis by producing realistic, synthetic, and privacy-preserving EHR data to tackle data scarcity issues in healthcare.}
However, it is important to acknowledge the limitations of our study. 
While our results suggest that \ours offers some degree of patient privacy protection, it should not be seen as a replacement for official audits, which may still be necessary prior to data sharing. 
It is also interesting to investigate \ours within established privacy frameworks, e.g., differential privacy. Additionally, to provide better interpretability and explainability of \ours, subgroup analysis and theoretical analysis are to be developed.
\secondedit{While we utilized sample mean imputation for computational efficiency, more advanced missing value imputation techniques could be considered to further evaluate \ours's behavior.}
Lastly, it would also be meaningful to investigate the modeling of highly sparse and irregular temporal data, such as lab tests and medications.
We leave the above potential improvements of \ours for future work.

\section*{Competing Interests}
All authors declare no financial or non-financial competing interests.

\section*{Data Availability\label{sec:data availability}}
The EHR datasets utilized during the current study are available in the following repositories: \href{https://mimic.mit.edu}{the MIMIC repository} and \href{https://eicu-crd.mit.edu}{the eICU Collaborative Research Database}. %and \href{https://hirid.intensivecare.ai/}{high time resolution ICU dataset (HiRID)}. 
% Both Stocks and Energy datasets can be downloaded from \href{https://github.com/jsyoon0823/TimeGAN/tree/master/data}{TimeGAN's official repository}.

We also consider the following non-EHR time-series dataset for comparisons.
Stocks: the dataset is available online and can be accessed from the \href{https://finance.yahoo.com/quote/GOOG/history?p=GOOG&guccounter=1}{historical Google stock price on Yahoo}; Energy: this dataset can be obtained from \href{https://archive.ics.uci.edu/dataset/374/appliances+energy+prediction}{UCI machine learning repository.}

\section*{Code Availability\label{sec:code availability}}
Our code is available at \href{https://github.com/MuhangTian/TimeDiff}{https://github.com/MuhangTian/TimeDiff}.

\section*{Funding Statement\label{sec:funding statement}}
This work was supported by CS+ program in 2023 at the Department of Computer Science, Duke University. S. J. and A. R. Z were also partially supported by NSF Grant CAREER-2203741 and NIH Grants R01HL169347 and R01HL168940.

\section*{Contributorship Statement\label{sec:contributorship statement}}
Muhang Tian, Bernie Chen, and Allan Guo were primarily responsible for the design and execution of the experiments and for writing the manuscript.
Shiyi Jiang and Anru Zhang were responsible for overseeing the project and writing the manuscript.

\section*{Competing Interests Statement\label{sec:competing interests statement}}
The authors have no competing interests to declare.

%\clearpage

\printbibliography

\newpage
\renewcommand \thepart{}
\renewcommand \partname{}
%\begin{comment}
\hypersetup{
    colorlinks=false,
    linkcolor=black,  % set to black or any desired color
    citecolor=black,  % set to black or any desired color
    filecolor=black,  % set to black or any desired color
    urlcolor=black    % set to black or any desired color
}
\setcounter{parttocdepth}{3} % 3 includes subsubsections
%\end{comment}
\doparttoc % Tell to minitoc to generate a toc for the parts
\faketableofcontents % Run a fake tableofcontents command for the partocs
\appendix
\addcontentsline{toc}{section}{Appendix} % Add the appendix text to the document TOC
\part{Supplementary Text to ``Reliable Generation of \secondedit{Privacy-preserving Synthetic} EHR Time Series via Diffusion Models"} % Start the appendix part
\parttoc % Insert the appendix TOC
\newpage
%\begin{comment}
\hypersetup{% for better colors
    colorlinks=true,
    citecolor=mydarkblue,
    linkcolor=mydarkred,
    filecolor=mydarkblue,  
    urlcolor=mydarkblue,  
}
%\end{comment}

\section{Experiment Details \label{appendix:detail}}
\newcommand{\AppendixTSNEeICU}{
\begin{subfigure}{0.25\linewidth}
     \centering
     \includegraphics[width=\linewidth]{figs/t-sne/eicu_layernorm_regular.pdf}
     \caption*{\textbf{\ours}}
 \end{subfigure}%
 \begin{subfigure}{0.25\linewidth}
     \centering
     \includegraphics[width=\linewidth]{figs/t-sne/eicu_ehrmgan_regular.pdf}
     \caption*{EHR-M-GAN}
 \end{subfigure}%
  \begin{subfigure}{0.25\linewidth}
     \centering
     \includegraphics[width=\linewidth]{figs/t-sne/eicu_dspd-gp_regular.pdf}
     \caption*{DSPD-GP}
\end{subfigure}%
 \begin{subfigure}{0.25\linewidth}
     \centering
     \includegraphics[width=\linewidth]{figs/t-sne/eicu_gtgan_regular.pdf}
     \caption*{GT-GAN}
\end{subfigure}\\
\begin{subfigure}{0.25\linewidth}
    \centering
    \includegraphics[width=\linewidth]{figs/t-sne/eicu_timegan_regular.pdf}
    \caption*{TimeGAN}
\end{subfigure}%
\begin{subfigure}{0.25\linewidth}
    \centering
    \includegraphics[width=\linewidth]{figs/t-sne/eicu_rcgan_regular.pdf}
    \caption*{RCGAN}
\end{subfigure}%
\begin{subfigure}{0.25\linewidth}
    \centering
    \includegraphics[width=\linewidth]{figs/t-sne/eicu_dspd-ou_regular.pdf}
    \caption*{DSPD-OU}
\end{subfigure}%
\begin{subfigure}{0.25\linewidth}
    \centering
    \includegraphics[width=\linewidth]{figs/t-sne/eicu_dspd-cont-gp_regular.pdf}
    \caption*{CSPD-GP}
\end{subfigure}\\
\begin{subfigure}{0.2\linewidth}
    \centering
    \includegraphics[width=\linewidth]{figs/t-sne/eicu_dspd-cont-ou_regular.pdf}
    \caption*{CSPD-OU}
\end{subfigure}%
\begin{subfigure}{0.2\linewidth}
    \centering
    \includegraphics[width=\linewidth]{figs/t-sne/eicu_crnngan_regular.pdf}
    \caption*{C-RNN-GAN}
\end{subfigure}%
\begin{subfigure}{0.2\linewidth}
    \centering
    \includegraphics[width=\linewidth]{figs/t-sne/eicu_t-forcing_regular.pdf}
    \caption*{T-Forcing}
\end{subfigure}%
\begin{subfigure}{0.2\linewidth}
    \centering
    \includegraphics[width=\linewidth]{figs/t-sne/eicu_p-forcing_regular.pdf}
    \caption*{P-Forcing}
\end{subfigure}%
\begin{subfigure}{0.2\linewidth}
    \centering
    \includegraphics[width=\linewidth]{figs/pdfs/eicu_halo_regular_tsne.pdf}
    \caption*{HALO}
\end{subfigure}%
}

\newcommand{\AppendixTSNEMIMICIV}{
\begin{subfigure}{0.25\linewidth}
     \centering
     \includegraphics[width=\linewidth]{figs/t-sne/mimiciv_layernorm_regular.pdf}
     \caption*{\textbf{\ours}}
\end{subfigure}%
\begin{subfigure}{0.25\linewidth}
     \centering
     \includegraphics[width=\linewidth]{figs/t-sne/mimiciv_ehrmgan_regular.pdf}
     \caption*{EHR-M-GAN}
\end{subfigure}%
\begin{subfigure}{0.25\linewidth}
     \centering
     \includegraphics[width=\linewidth]{figs/t-sne/mimiciv_dspd-gp_regular.pdf}
     \caption*{DSPD-GP}
\end{subfigure}%
 \begin{subfigure}{0.25\linewidth}
     \centering
     \includegraphics[width=\linewidth]{figs/t-sne/mimiciv_gtgan_regular.pdf}
     \caption*{GT-GAN}
\end{subfigure}\\
\begin{subfigure}{0.25\linewidth}
    \centering
    \includegraphics[width=\linewidth]{figs/t-sne/mimiciv_timegan_regular.pdf}
    \caption*{TimeGAN}
\end{subfigure}%
\begin{subfigure}{0.25\linewidth}
    \centering
    \includegraphics[width=\linewidth]{figs/t-sne/mimiciv_rcgan_regular.pdf}
    \caption*{RCGAN}
\end{subfigure}%
\begin{subfigure}{0.25\linewidth}
    \centering
    \includegraphics[width=\linewidth]{figs/t-sne/mimiciv_dspd-ou_regular.pdf}
    \caption*{DSPD-OU}
\end{subfigure}%
\begin{subfigure}{0.25\linewidth}
    \centering
    \includegraphics[width=\linewidth]{figs/t-sne/mimiciv_dspd-cont-gp_regular.pdf}
    \caption*{CSPD-GP}
\end{subfigure}\\
\begin{subfigure}{0.2\linewidth}
    \centering
    \includegraphics[width=\linewidth]{figs/t-sne/mimiciv_dspd-cont-ou_regular.pdf}
    \caption*{CSPD-OU}
\end{subfigure}%
\begin{subfigure}{0.2\linewidth}
    \centering
    \includegraphics[width=\linewidth]{figs/t-sne/mimiciv_crnngan_regular.pdf}
    \caption*{C-RNN-GAN}
\end{subfigure}%
\begin{subfigure}{0.2\linewidth}
    \centering
    \includegraphics[width=\linewidth]{figs/t-sne/mimiciv_tforcing_regular.pdf}
    \caption*{T-Forcing}
\end{subfigure}%
\begin{subfigure}{0.2\linewidth}
    \centering
    \includegraphics[width=\linewidth]{figs/t-sne/mimiciv_pforcing_regular.pdf}
    \caption*{P-Forcing}
\end{subfigure}%
\begin{subfigure}{0.2\linewidth}
    \centering
    \includegraphics[width=\linewidth]{figs/pdfs/mimiciv_halo_regular_tsne.pdf}
    \caption*{HALO}
\end{subfigure}%
}

\newcommand{\AppendixTSNEMIMICIII}{
\begin{subfigure}{0.25\linewidth}
     \centering
     \includegraphics[width=\linewidth]{figs/t-sne/mimiciii_layernorm_regular.pdf}
     \caption*{\textbf{\ours}}
\end{subfigure}%
\begin{subfigure}{0.25\linewidth}
     \centering
     \includegraphics[width=\linewidth]{figs/t-sne/mimiciii_ehrmgan_regular.pdf}
     \caption*{EHR-M-GAN}
\end{subfigure}%
  \begin{subfigure}{0.25\linewidth}
     \centering
     \includegraphics[width=\linewidth]{figs/t-sne/mimiciii_dspd-gp_regular.pdf}
     \caption*{DSPD-GP}
\end{subfigure}%
 \begin{subfigure}{0.25\linewidth}
     \centering
     \includegraphics[width=\linewidth]{figs/t-sne/mimiciii_gtgan_regular.pdf}
     \caption*{GT-GAN}
\end{subfigure}\\
\begin{subfigure}{0.25\linewidth}
    \centering
    \includegraphics[width=\linewidth]{figs/t-sne/mimiciii_timegan_regular.pdf}
    \caption*{TimeGAN}
\end{subfigure}%
\begin{subfigure}{0.25\linewidth}
    \centering
    \includegraphics[width=\linewidth]{figs/t-sne/mimiciii_rcgan_regular.pdf}
    \caption*{RCGAN}
\end{subfigure}%
\begin{subfigure}{0.25\linewidth}
    \centering
    \includegraphics[width=\linewidth]{figs/t-sne/mimiciii_dspd-ou_regular.pdf}
    \caption*{DSPD-OU}
\end{subfigure}%
\begin{subfigure}{0.25\linewidth}
    \centering
    \includegraphics[width=\linewidth]{figs/t-sne/mimiciii_dspd-cont-gp_regular.pdf}
    \caption*{CSPD-GP}
\end{subfigure}\\
\begin{subfigure}{0.2\linewidth}
    \centering
    \includegraphics[width=\linewidth]{figs/t-sne/mimiciii_dspd-cont-ou_regular.pdf}
    \caption*{CSPD-OU}
\end{subfigure}%
\begin{subfigure}{0.2\linewidth}
    \centering
    \includegraphics[width=\linewidth]{figs/t-sne/mimiciii_crnngan_regular.pdf}
    \caption*{C-RNN-GAN}
\end{subfigure}%
\begin{subfigure}{0.2\linewidth}
    \centering
    \includegraphics[width=\linewidth]{figs/t-sne/mimiciii_tforcing_regular.pdf}
    \caption*{T-Forcing}
\end{subfigure}%
\begin{subfigure}{0.2\linewidth}
    \centering
    \includegraphics[width=\linewidth]{figs/t-sne/mimiciii_pforcing_regular.pdf}
    \caption*{P-Forcing}
\end{subfigure}%
\begin{subfigure}{0.2\linewidth}
    \centering
    \includegraphics[width=\linewidth]{figs/pdfs/mimiciii_halo_regular_tsne.pdf}
    \caption*{HALO}
\end{subfigure}%
}

\newcommand{\AppendixTimeModelSize}{
\begin{adjustbox}{width=\linewidth}
\begin{tabular}{@{}lllllllll@{}}
    \toprule
    Dataset & \textbf{\ours} & EHR-M-GAN & TimeGAN & GT-GAN & DSPD-GP & DSPD-OU & CSPD-GP & CSPD-OU \\
    \midrule
    MIMIC-III & 2.7 & 18.9 & 10.8 & 21.8 & 2.5 & 2.5 & 2.5 & 2.5  \\
    MIMIC-IV & 2.7 & 28.8 & 29.5 & 47.3 & 2.6 & 2.6 & 2.6 & 2.6 \\
    %HiRID & 2.5 & 29.7 & 46.2 & 58.3 & 2.8 & 2.8 & 2.8 & 2.8 \\
    eICU & 8.7 & 87.1 & 110 & 59.1 & 7.0 & 7.0 & 7.0 & 7.0 \\
    \bottomrule
\end{tabular}
\end{adjustbox}
}

\newcommand{\AppendixTSTRAUROC}{
\begin{subfigure}{0.33\linewidth}
     \centering
     \includegraphics[width=\linewidth]{figs/tstr/eicu_tstrtrtr.pdf}
     \caption*{eICU}
 \end{subfigure}%
 %\begin{subfigure}{0.5\linewidth}
 %    \centering
     %\includegraphics[width=\linewidth]{figs/tstr/hirid_tstrtrtr.pdf}
     %\caption*{HiRID}
 %\end{subfigure}
 \begin{subfigure}{0.33\linewidth}
     \centering
     \includegraphics[width=\linewidth]{figs/tstr/mimiciii_tstrtrtr.pdf}
     \caption*{MIMIC-III}
 \end{subfigure}%
 \begin{subfigure}{0.33\linewidth}
     \centering
     \includegraphics[width=\linewidth]{figs/tstr/mimiciv_tstrtrtr.pdf}
     \caption*{MIMIC-IV}
 \end{subfigure}
}

\newcommand{\AppendixTSTRAUROCSummary}{
\begin{subfigure}{0.33\linewidth}
     \centering
     \includegraphics[width=\linewidth]{figs/tstr/eicu_tstrtrtr_summary.pdf}
     \caption*{eICU}
 \end{subfigure}%
 %\begin{subfigure}{0.5\linewidth}
 %    \centering
     %\includegraphics[width=\linewidth]{figs/tstr/hirid_tstrtrtr_summary.pdf}
     %\caption*{HiRID}
 %\end{subfigure}
 \begin{subfigure}{0.33\linewidth}
     \centering
     \includegraphics[width=\linewidth]{figs/tstr/mimiciii_tstrtrtr_summary.pdf}
     \caption*{MIMIC-III}
 \end{subfigure}%
 \begin{subfigure}{0.33\linewidth}
     \centering
     \includegraphics[width=\linewidth]{figs/tstr/mimiciv_tstrtrtr_summary.pdf}
     \caption*{MIMIC-IV}
 \end{subfigure}
}

\newcommand{\AppendixTSRTRAUROC}{
\begin{subfigure}{0.33\linewidth}
     \centering
     \includegraphics[width=\linewidth]{figs/tsrtr/eicu_tsrtr.pdf}
     \caption*{eICU}
 \end{subfigure}%
 %\begin{subfigure}{0.5\linewidth}
 %    \centering
     %\includegraphics[width=\linewidth]{figs/tsrtr/hirid_tsrtr.pdf}
     %\caption*{HiRID}
 %\end{subfigure}\\
 \begin{subfigure}{0.33\linewidth}
     \centering
     \includegraphics[width=\linewidth]{figs/tsrtr/mimiciii_tsrtr.pdf}
     \caption*{MIMIC-III}
 \end{subfigure}%
 \begin{subfigure}{0.33\linewidth}
     \centering
     \includegraphics[width=\linewidth]{figs/tsrtr/mimiciv_tsrtr.pdf}
     \caption*{MIMIC-IV}
 \end{subfigure}
}

\newcommand{\AppendixTSRTRAUROCSummary}{
\begin{subfigure}{0.33\linewidth}
     \centering
     \includegraphics[width=\linewidth]{figs/tsrtr/eicu_tsrtr_summary.pdf}
     \caption*{eICU}
 \end{subfigure}%
 %\begin{subfigure}{0.5\linewidth}
 %    \centering
     %\includegraphics[width=\linewidth]{figs/tsrtr/hirid_tsrtr_summary.pdf}
     %\caption*{HiRID}
 %\end{subfigure}\\
 \begin{subfigure}{0.33\linewidth}
     \centering
     \includegraphics[width=\linewidth]{figs/tsrtr/mimiciii_tsrtr_summary.pdf}
     \caption*{MIMIC-III}
 \end{subfigure}%
 \begin{subfigure}{0.33\linewidth}
     \centering
     \includegraphics[width=\linewidth]{figs/tsrtr/mimiciv_tsrtr_summary.pdf}
     \caption*{MIMIC-IV}
 \end{subfigure}
}

\newcommand{\AppendixAblationMissing}{
% \begin{adjustbox}{width=\textwidth}
\begin{tabular}{@{}clccc@{}}
    \toprule
    Metric & Method & MIMIC-III & MIMIC-IV & eICU \\
    \midrule
    & with Gaussian and rounding & .355$\pm$.020 & .121$\pm$.025 & .030$\pm$.018 \\
    Discriminative & with Gaussian and softmax & .088$\pm$.023 & .155 $\pm$.032 & \\
    Score ($\downarrow$) & with multinomial & \textbf{.028$\pm$.023} & \textbf{.030$\pm$.022} & \textbf{.015$\pm$.007} \\
    & \textit{Real Data} & \textit{.012$\pm$.006} & \textit{.014$\pm$.011} & \textit{.004$\pm$.003} \\
    \midrule
    & with Gaussian and rounding & .486$\pm$.005 & .433$\pm$.003 & .312$\pm$.031 \\
    Predictive & with Gaussian and softmax & .472$\pm$.004 & & \\
    Score ($\downarrow$) & with multinomial & \textbf{.469$\pm$.003} & \textbf{.432$\pm$.002} & \textbf{.309$\pm$.019} \\
    & \textit{Real Data} & \textit{.467$\pm$.005} & \textit{.433$\pm$.001} & \textit{.304$\pm$.017} \\
    \bottomrule
\end{tabular}
% \end{adjustbox}
}

\newcommand{\AppendixAblationLambda}{
\begin{tabular}{@{}clccc@{}}
    \toprule
    Metric & Method & MIMIC-III & MIMIC-IV & eICU \\
    \midrule
    & $\lambda = 0.001$ & .106$\pm$.047 & .054$\pm$.023 & .018$\pm$.010 \\
    & $\lambda = 0.01$ & .028$\pm$.023 & .030$\pm$.022 & .015$\pm$.007 \\
    Discriminative & $\lambda = 0.1$ & .045$\pm$.046 & .036$\pm$.026 & .027$\pm$.011 \\
    Score & $\lambda = 1$ & .108$\pm$.041 & .125$\pm$.068 & .068$\pm$.016 \\
     ($\downarrow$) & $\lambda = 10$ & .430$\pm$.037 & .441$\pm$.090 & .299$\pm$.048 \\
    \addlinespace
    & \textit{Real Data} & \textit{.012$\pm$.006} & \textit{.014$\pm$.011} & \textit{.004$\pm$.003} \\
    \midrule
    & $\lambda = 0.001$ & .472$\pm$.002 & .433$\pm$.002 & .305$\pm$.017\\
    & $\lambda = 0.01$ & .469$\pm$.003 & .432$\pm$.002 & 309$\pm$.019 \\
    Predictive & $\lambda = 0.1$ & .469$\pm$.002 & .434$\pm$.002 & .319$\pm$.036 \\
    Score & $\lambda = 1$ & .472$\pm$.003 & .435$\pm$.002 & .317$\pm$.036 \\
    ($\downarrow$) & $\lambda = 10$ & .496$\pm$.002 & .488$\pm$.008 & .314$\pm$.018 \\
    \addlinespace
    & \textit{Real Data} & \textit{.467$\pm$.005} & \textit{.433$\pm$.001} & \textit{.304$\pm$.017} \\
    \bottomrule
\end{tabular}
}

\newcommand{\AppendixAblationTSNE}{
\begin{subfigure}{0.2\linewidth}
     \centering
     \includegraphics[width=\linewidth]{figs/t-sne/mimiciii_layernorm_ablation_0.001_regular.pdf}
\end{subfigure}%
\begin{subfigure}{0.2\linewidth}
     \centering
     \includegraphics[width=\linewidth]{figs/t-sne/mimiciii_layernorm_regular.pdf}
\end{subfigure}%
\begin{subfigure}{0.2\linewidth}
     \centering
     \includegraphics[width=\linewidth]{figs/t-sne/mimiciii_layernorm_ablation_0.1_regular.pdf}
\end{subfigure}%
\begin{subfigure}{0.2\linewidth}
     \centering
     \includegraphics[width=\linewidth]{figs/t-sne/mimiciii_layernorm_ablation_1_regular.pdf}
\end{subfigure}%
\begin{subfigure}{0.2\linewidth}
     \centering
     \includegraphics[width=\linewidth]{figs/t-sne/mimiciii_layernorm_ablation_10_regular.pdf}
\end{subfigure}\\
\begin{subfigure}{0.2\linewidth}
     \centering
     \includegraphics[width=\linewidth]{figs/t-sne/mimiciv_layernorm_ablation_0.001_regular.pdf}
\end{subfigure}%
\begin{subfigure}{0.2\linewidth}
     \centering
     \includegraphics[width=\linewidth]{figs/t-sne/mimiciv_layernorm_regular.pdf}
\end{subfigure}%
\begin{subfigure}{0.2\linewidth}
     \centering
     \includegraphics[width=\linewidth]{figs/t-sne/mimiciv_layernorm_ablation_0.1_regular.pdf}
\end{subfigure}%
\begin{subfigure}{0.2\linewidth}
     \centering
     \includegraphics[width=\linewidth]{figs/t-sne/mimiciv_layernorm_ablation_1_regular.pdf}
\end{subfigure}%
\begin{subfigure}{0.2\linewidth}
     \centering
     \includegraphics[width=\linewidth]{figs/t-sne/mimiciv_layernorm_ablation_10_regular.pdf}
\end{subfigure}\\
\begin{subfigure}{0.2\linewidth}
     \centering
     \includegraphics[width=\linewidth]{figs/t-sne/eicu_layernorm_ablation_0.001_regular.pdf}
     \caption*{$\lambda = 0.001$}
\end{subfigure}%
\begin{subfigure}{0.2\linewidth}
     \centering
     \includegraphics[width=\linewidth]{figs/t-sne/eicu_layernorm_regular.pdf}
     \caption*{$\lambda = 0.01$}
\end{subfigure}%
\begin{subfigure}{0.2\linewidth}
     \centering
     \includegraphics[width=\linewidth]{figs/t-sne/eicu_layernorm_ablation_0.1_regular.pdf}
     \caption*{$\lambda = 0.1$}
\end{subfigure}%
\begin{subfigure}{0.2\linewidth}
     \centering
     \includegraphics[width=\linewidth]{figs/t-sne/eicu_layernorm_ablation_1_regular.pdf}
     \caption*{$\lambda = 1$}
\end{subfigure}%
\begin{subfigure}{0.2\linewidth}
     \centering
     \includegraphics[width=\linewidth]{figs/t-sne/eicu_layernorm_ablation_10_regular.pdf}
     \caption*{$\lambda = 10$}
\end{subfigure}\\
}

\newcommand{\AppendixTSNENONEHR}{
\begin{subfigure}{0.143\linewidth}
     \centering
     \includegraphics[width=\linewidth]{figs/t-sne/stock_layernorm_regular.pdf}
\end{subfigure}%
\begin{subfigure}{0.143\linewidth}
     \centering
     \includegraphics[width=\linewidth]{figs/t-sne/stock_ehrmgan_regular.pdf}
\end{subfigure}%
\begin{subfigure}{0.143\linewidth}
     \centering
     \includegraphics[width=\linewidth]{figs/t-sne/stock_dspd-gp_regular.pdf}
\end{subfigure}%
\begin{subfigure}{0.143\linewidth}
     \centering
     \includegraphics[width=\linewidth]{figs/t-sne/stock_dspd-ou_regular.pdf}
\end{subfigure}%
\begin{subfigure}{0.143\linewidth}
     \centering
     \includegraphics[width=\linewidth]{figs/t-sne/stock_dspd-cont-gp_regular.pdf}
\end{subfigure}%
\begin{subfigure}{0.143\linewidth}
     \centering
     \includegraphics[width=\linewidth]{figs/t-sne/stock_dspd-cont-ou_regular.pdf}
\end{subfigure}%
\begin{subfigure}{0.143\linewidth}
     \centering
     \includegraphics[width=\linewidth]{figs/pdfs/stocks_halo_regular_tsne.pdf}
\end{subfigure}\\
\begin{subfigure}{0.143\linewidth}
     \centering
     \includegraphics[width=\linewidth]{figs/t-sne/energy_layernorm_regular.pdf}
     \caption*{\textbf{\ours}}
\end{subfigure}%
\begin{subfigure}{0.143\linewidth}
     \centering
     \includegraphics[width=\linewidth]{figs/t-sne/energy_ehrmgan_regular.pdf}
     \caption*{EHR-M-GAN}
\end{subfigure}%
\begin{subfigure}{0.143\linewidth}
     \centering
     \includegraphics[width=\linewidth]{figs/t-sne/energy_dspd-gp_regular.pdf}
     \caption*{DSPD-GP}
\end{subfigure}%
\begin{subfigure}{0.143\linewidth}
     \centering
     \includegraphics[width=\linewidth]{figs/t-sne/energy_dspd-ou_regular.pdf}
     \caption*{DSPD-OU}
\end{subfigure}%
\begin{subfigure}{0.143\linewidth}
     \centering
     \includegraphics[width=\linewidth]{figs/t-sne/energy_dspd-cont-gp_regular.pdf}
     \caption*{CSPD-GP}
\end{subfigure}%
\begin{subfigure}{0.143\linewidth}
     \centering
     \includegraphics[width=\linewidth]{figs/t-sne/energy_dspd-cont-ou_regular.pdf}
     \caption*{CSPD-OU}
\end{subfigure}%
\begin{subfigure}{0.143\linewidth}
     \centering
     \includegraphics[width=\linewidth]{figs/pdfs/energy_halo_regular_tsne.pdf}
     \caption*{HALO}
\end{subfigure}\\
}

\newcommand{\AppendixPrivacyTableone}{
\begin{tabular}{@{}llccc@{}}
    \toprule
    Metric & Method & MIMIC-III & MIMIC-IV & eICU \\
    \midrule
    $\aatest$ ($\sim$0.5) & \textbf{\ours} & \textbf{.574$\pm$.002} & \textbf{.517$\pm$.002}  &  \textbf{.537$\pm$.001} \\
    & EHR-M-GAN & .998$\pm$.000 & 1.000$\pm$.000 & .977$\pm$.000 \\
    & DSPD-GP & .974$\pm$.001 & .621$\pm$.002 & .888$\pm$.000 \\
    & DSPD-OU & .927$\pm$.000 & .804$\pm$.003 & .971$\pm$.000\\
    & CSPD-GP & .944$\pm$.001 & .623$\pm$.002 & .851$\pm$.001\\
    & CSPD-OU & .967$\pm$.001 & .875$\pm$.002 & .982$\pm$.000\\
    & GT-GAN & .995$\pm$.000 & .910$\pm$.001 & .981$\pm$.000 \\
    & TimeGAN & .997$\pm$.000 & .974$\pm$.001 & 1.000$\pm$.000\\
    & RCGAN & .983$\pm$.001 & .999$\pm$.000 & 1.000$\pm$.000\\
    & C-RNN-GAN & 1.000$\pm$.000 & .993$\pm$.000 & 1.000$\pm$.000\\
    & T-Forcing & 1.000$\pm$.000 & .928$\pm$.001 & .999$\pm$.000\\
    & P-Forcing & 1.000$\pm$.000 & .977$\pm$.001 & 1.000$\pm$.000\\
    & \firstedit{HALO} & \firstedit{.698$\pm$.002} & \firstedit{.709$\pm$.002}  & \firstedit{.653$\pm$.001} \\
    & \textit{Real Data} & \textit{.552$\pm$.002} & \textit{.497$\pm$.002} & \textit{.501$\pm$.002} \\
    \midrule
    $\aatrain$ ($\sim$0.5) & \textbf{\ours} & \textbf{.573$\pm$.002} & \textbf{.515$\pm$.002} &  \textbf{.531$\pm$.002} \\
    & EHR-M-GAN & .999$\pm$.000 & 1.000$\pm$.000 & .965$\pm$.002\\
    & DSPD-GP & .968$\pm$.002 & .620$\pm$.003 & .888$\pm$.001 \\
    & DSPD-OU & .928$\pm$.001 & .788$\pm$.003 & .971$\pm$.000\\
    & CSPD-GP & .940$\pm$.002 & .629$\pm$.005 & .852$\pm$.001\\
    & CSPD-OU & .966$\pm$.001 & .880$\pm$.003 & .983$\pm$.000\\
    & GT-GAN & .995$\pm$.001 & .907$\pm$.002 & .981$\pm$.000 \\
    & TimeGAN & .997$\pm$.000 & .969$\pm$.003 & 1.000$\pm$.000 \\
    & RCGAN & .984$\pm$.001 & .999$\pm$.000 & 1.000$\pm$.000\\
    & C-RNN-GAN & 1.000$\pm$.000 & .992$\pm$.001 & 1.000$\pm$.000\\
    & T-Forcing & 1.000$\pm$.000 & .927$\pm$.002 & .999$\pm$.000\\
    & P-Forcing & 1.000$\pm$.000 & .976$\pm$.002 & 1.000$\pm$.000\\
    & \firstedit{HALO} & \firstedit{.696$\pm$.001} & \firstedit{.717$\pm$.002} &  \firstedit{.653$\pm$.002} \\
    & \textit{Real Data} & \textit{.286$\pm$.003} & \textit{.268$\pm$.004} & \textit{.266$\pm$.002} \\
    \bottomrule
\end{tabular}
}
\newcommand{\AppendixPrivacyTabletwo}{
\begin{tabular}{@{}llccc@{}}
    \toprule
    Metric & Method & MIMIC-III & MIMIC-IV & eICU \\
    \midrule
    NNAA ($\downarrow$) & \textbf{\ours} & .002$\pm$.002 & .002$\pm$.002 & .006$\pm$.002 \\
    & EHR-M-GAN & .000$\pm$.000 & .000$\pm$.000 & .012$\pm$.003\\
    & DSPD-GP & .005$\pm$.003 & .003$\pm$.003 & .001$\pm$.001\\
    & DSPD-OU & .001$\pm$.001 & .016$\pm$.004 & .000$\pm$.000\\
    & CSPD-GP & .004$\pm$.002 & .007$\pm$.005 & .001$\pm$.001\\
    & CSPD-OU & .001$\pm$.001 & .005$\pm$.003 & .001$\pm$.001\\
    & GT-GAN & .001$\pm$.000 & .004$\pm$.002 & .000$\pm$.000 \\
    & TimeGAN & .000$\pm$.000 & .005$\pm$.003 & .000$\pm$.000 \\
    & RCGAN & .001$\pm$.000 & .000$\pm$.000 & .000$\pm$.000\\
    & C-RNN-GAN & .000$\pm$.000 & .001$\pm$.000 & .000$\pm$.000\\
    & T-Forcing & .000$\pm$.000 & .002$\pm$.001 & .000$\pm$.000\\
    & P-Forcing & .000$\pm$.000 & .002$\pm$.002 & .000$\pm$.000\\
    & \firstedit{HALO} & \firstedit{.002$\pm$.002} & \firstedit{.008$\pm$.002} &  \firstedit{.002$\pm$.001} \\
    & \textit{Real Data} & \textit{.267$\pm$.004} & \textit{.229$\pm$.003} &  \textit{.235$\pm$.003} \\
    \midrule
    MIR ($\downarrow$) & \textbf{\ours} & .191$\pm$.008 & .232$\pm$.048 & .227$\pm$.021 \\
    & EHR-M-GAN & .025$\pm$.007 & .435$\pm$.031 & .049$\pm$.006 \\
    & DSPD-GP & .032$\pm$.021 & .050$\pm$.009 & 
    .000$\pm$.000 \\
    & DSPD-OU & .060$\pm$.032 & .007$\pm$.006 & .000$\pm$.000 \\
    & CSPD-GP & .060$\pm$.028 & .034$\pm$.017 & .000$\pm$.000 \\
    & CSPD-OU & .066$\pm$.046 & .016$\pm$.020 & .000$\pm$.000 \\
    & GT-GAN & .005$\pm$.002 & .046$\pm$.013 & .000$\pm$.000 \\
    & TimeGAN & .010$\pm$.002 & .173$\pm$.020 & .000$\pm$.000 \\
    & RCGAN & .013$\pm$.002 & .277$\pm$.049 & .000$\pm$.000 \\
    & C-RNN-GAN & .015$\pm$.005 & .011$\pm$.006 & .000$\pm$.000 \\
    & T-Forcing & .007$\pm$.003 & .215$\pm$.052 & .000$\pm$.000 \\
    & P-Forcing & .004$\pm$.004 & .131$\pm$.045 & .003$\pm$.001\\
    & \firstedit{HALO} & \firstedit{.189$\pm$.007} & \firstedit{.019$\pm$.012} & \firstedit{.036$\pm$.040} \\
    & \textit{Real Data} & \textit{.948$\pm$.000} & \textit{.929$\pm$.005} & \textit{.927$\pm$.001}\\
    \bottomrule
\end{tabular}
}
\newcommand{\AppendixTSTRRNN}{
\begin{tabular}{@{}llccc@{}}
    \toprule
    Method & Metric & MIMIC-III & MIMIC-IV & eICU \\
    \midrule
    GRU & TSTR & .584$\pm$.016 & .516$\pm$.025 & .544$\pm$.020 \\
    & TRTR & .543$\pm$.018 & .507$\pm$.022 & .476$\pm$.029 \\
    LSTM & TSTR & .581$\pm$.019 & .484$\pm$.010 & .558$\pm$.037 \\
    & TRTR & .587$\pm$.026 & .473$\pm$.025 & .531$\pm$.029\\
    \bottomrule
\end{tabular}
}

\newcommand{\AppendixMIMICIVRevisionAllMetrics}{
    \begin{tabular}{@{}clcc@{}}
    \toprule
    Metric & Method & MIMIC-IV \\
    \midrule
    Discriminative Score ($\downarrow$) & \textbf{\ours} & .050$\pm$.029 \\
    \addlinespace
     & \textit{Real Data} & \textit{.010$\pm$.004}  \\
    \midrule
    Predictive Score ($\downarrow$) & \textbf{\ours} & .317$\pm$.001 \\
    \addlinespace
     & \textit{Real Data} & \textit{.312$\pm$.001}\\
    \midrule
    $\aatrain$ ($\sim$0.5) & \textbf{\ours} & .568$\pm$.005 \\
    \addlinespace
     & \textit{Real Data} & \textit{.268$\pm$.004}\\
    \addlinespace
    \addlinespace
     $\aatest$ ($\sim$0.5) & \textbf{\ours} & .569$\pm$.004 \\
     \addlinespace
     & \textit{Real Data} & \textit{.496$\pm$.002}\\
     \addlinespace
     \addlinespace
     NNAA ($\downarrow$) & \textbf{\ours} & .004$\pm$.003\\ 
     \addlinespace
     & \textit{Real Data} & \textit{.228$\pm$.005}\\
    \addlinespace
    \addlinespace
     MIR ($\downarrow$) & \textbf{\ours} & .097$\pm$.04 \\
     \addlinespace
     & \textit{Real Data} & \textit{.971$\pm$.005}\\
    \bottomrule
    \end{tabular}
}

\subsection{Datasets \label{appendix:detail:datasets}}
In this section, we provide further information on the datasets used in this study and the corresponding data processing procedures.
Unless specified otherwise, all datasets are normalized by min-max scaling for model training, and the minimums and maximums are calculated feature-wise, i.e., we normalize each feature by its corresponding sample minimum and maximum, and this procedure is applied across all the features.
For all EHR datasets, we extract the in-hospital mortality status as our class labels for TSTR and TSRTR evaluations.

\begin{table}[h!bt]
\caption{Dataset statistics.}
%\vspace{-10pt}
\centering
\begin{adjustbox}{width=\textwidth}
\begin{tabular}{@{}lllllll@{}}
    \toprule
    Dataset &\phantom{abc}& Sample Size & Number of Features & Sequence Length & Missing (\%) & Mortality Rate (\%) \\
    \midrule
    Stocks && 3,773 & 6 & 24 & 0 & ---\\
    Energy && 19,711 & 28 & 24 & 0 & ---\\
    MIMIC-III && 26,150 & 15 & 25 & 17.9 & 7.98\\
    MIMIC-IV && 21,593 & 11 & 72 & 7.9 & 23.67 \\
    %HiRID && 6,709 & 8 & 100 & 0 & 16.83\\
    eICU && 62,453 & 9 & 276 & 10.5 & 10.63\\
    \bottomrule
\end{tabular}
\end{adjustbox}
\end{table}

\subsubsection{Stocks \& Energy}
As mentioned earlier in the Results section, we use the Stocks and Energy datasets for a fair comparison between \ours and the existing GAN-based time-series generation methods. %\Cref{sec:results}

\textbf{Stocks:} The Stocks dataset contains daily Google stock data recorded between 2004 and 2019.
It contains features such as volume, high, low, opening, closing, and adjusted closing prices.
Each data point represents the value of those six features on a single day. 

\textbf{Energy:} The Energy dataset is from the UCI Appliances energy prediction data. It contains multivariate continuous-valued time series and has high-dimensional, correlated, and periodic features. 

To prepare both datasets for training and ensure consistency with previous approaches for a fair comparison, we use the same procedure as TimeGAN.
We then apply training and testing splits for both datasets.
For the Stocks dataset, we use 80\% for training and 20\% for testing.
For the Energy dataset, we use 75\% for training and 25\% for testing.

\subsubsection{MIMIC-III}
The \href{https://physionet.org/content/mimiciii/1.4/}{Medical Information Mart for Intensive Care-III} (MIMIC-III) is a single-center database consisting of a large collection of EHR data for patients admitted to critical care units at Beth Israel Deaconess Medical Center between 2001 and 2012.
The dataset contains information such as demographics, lab results, vital measurements, procedures, caregiver notes, and patient outcomes.
It contains data for 38,597 distinct adult patients and 49,785 hospital admissions.

\textbf{Variable Selection:}
In our study, we use the following vital sign measurements from MIMIC-III: heart rate (beats per minute), systolic blood pressure (mm Hg), diastolic blood pressure (mm Hg), mean blood pressure (mm Hg), respiratory rate (breaths per minute), body temperature (Celsius), and oxygen saturation (\%).
To ensure consistency and reproducibility, we adopt the scripts in \href{https://github.com/MIT-LCP/mimic-code/tree/main/mimic-iii}{official MIMIC-III repository} for data pre-processing that selects the aforementioned features based on \textit{itemid} and filters potential outliers \footnote{For reproducibility, the thresholds for the outliers are defined by the official repository.}.
We then extract records of the selected variables within the first 24 hours of a patient's unit stay at one-hour intervals, where the initial measurement is treated as time step 0.
This procedure gives us a multivariate time series of length 25 for each patient.

\textbf{Cohort Selection:}
We select our MIMIC-III study cohort by applying the outlier filter criteria adopted by the official MIMIC-III repository.
The filtering rules can be accessed \href{https://github.com/MIT-LCP/mimic-code/blob/main/mimic-iii/concepts_postgres/firstday/vitals_first_day.sql}{here}.
We select patients based on the unit stay level using \textit{icustay\_id}.
We only include patients who have spent at least 24 hours in their ICU stay.
We use 80\% of the dataset for training and 20\% for testing while ensuring a similar class ratio between the splits.

\subsubsection{MIMIC-IV}
The \href{https://physionet.org/content/mimiciv/2.2/}{Medical Information Mart for Intensive Care-IV} (MIMIC-IV) is a large collection of data for over 40,000 patients at intensive care units at the Beth Israel Deaconess Medical Center. 
It contains retrospectively collected medical data for 299,712 patients, 431,231 admissions, and 73,181 ICU stays.
It improves upon the MIMIC-III dataset, incorporating more up-to-date medical data with an optimized data storage structure.
In our study, we use vital signs for time-series generation.
To simplify the data-cleaning process, we adopt scripts from the \href{https://github.com/MIT-LCP/mimic-code/tree/main}{MIMIC Code Repository}. 

\textbf{Variable Selection:}
We extracted five vital signs for each patient from MIMIC-IV.
The selected variables are heart rate (beats per minute), systolic blood pressure (mm Hg), diastolic blood pressure (mm Hg), respiratory rate (breaths per minute), and oxygen saturation (\%).
We extract all measurements of each feature within the first 72 hours of each patient's ICU admission.
Similar to MIMIC-III, we encode the features using the method described in the Missing value representation section for model training. %\Cref{method:feature representation}

\textbf{Cohort Selection:}
Similar to MIMIC-III, we select our MIMIC-IV study cohort by applying filtering criteria provided by the official MIMIC-IV repository.
The criteria can be accessed \href{https://github.com/MIT-LCP/mimic-code/blob/main/mimic-iv/concepts_postgres/measurement/vitalsign.sql}{here}.
We also select patients at the unit stay level and include those who stayed for at least 72 hours in the ICU.
We use 75\% for training and 25\% for testing, and the class ratio is kept similar across the training and testing data.

\subsubsection{eICU}
The \href{https://eicu-crd.mit.edu/}{eICU Collaborative Research Database} is a multi-center database with 200,859 admissions to intensive care units monitored by the eICU programs across the United States.
It includes various information for the patients, such as vital sign measurements, care plan documentation, severity of illness measures, diagnosis information, and treatment information.
The database contains 139,367 patients admitted to critical care units between 2014 and 2015.
%The Philips eICU program, a telehealth program that transfers patient information to caregivers at the bedside, is responsible for data collection for eICU.

%This table contains consistently interfaced bedside vital signs monitored into eCareManager.
\textbf{Variable Selection:}
We select four vital sign variables from the \textit{vitalPeriodic} table in our study: heart rate (beats per minute), respiratory rate (breaths per minute), oxygen saturation (\%), and mean blood pressure (mm Hg).
The measurements are recorded as one-minute averages and are then stored as five-minute medians.
We extract values between each patient's first hour of the ICU stay and the next 24 hours for the selected variables.
Since the measurements are recorded at 5-minute intervals, we obtain a multivariate time series of length 276 for each patient in our study cohort.

\textbf{Cohort Selection:}
We select patients for our eICU study cohort by filtering the time interval.
Specifically, we include patients who stay for at least 24 hours in their ICU stay, and the time series measurements are extracted.
We did not use filtering criteria for time series in eICU.
This is a design choice that allows us to evaluate \ours when unfiltered time series are used as the input.
We also select patients at the unit stay level.
We use 75\% for training and 25\% for testing while ensuring the class ratio is similar between the two data splits.

\subsection{Baselines \label{appendix:detail:baselines}}
We reference the following source code for implementations of our baselines.
\begin{table}[hbt!]
    \caption{Source code links for all baselines.}
    %\vspace{-10pt}
    \centering
    \begin{tabular}{@{}ll@{}}
        \toprule
        Method & Source Code Link \\
        \midrule
        EHR-M-GAN \cite{li2023generating} & \href{https://github.com/jli0117/ehrMGAN}{LINK} \\
        DSPD/CSPD (GP or OU) \cite{pmlr-v202-bilos23a} & \href{https://github.com/morganstanley/MSML/tree/main/papers/Stochastic_Process_Diffusion}{LINK} \\
        GT-GAN \cite{jeon2022gtgan} & \href{https://github.com/Jinsung-Jeon/GT-GAN}{LINK} \\
        TimeGAN \cite{yoon2019timegan} & \href{https://github.com/jsyoon0823/TimeGAN}{LINK} \\
        RCGAN \cite{esteban2017realvalued} & \href{https://github.com/ratschlab/RGAN}{LINK} \\
        C-RNN-GAN \cite{mogren2016crnngan} & \href{https://github.com/cjbayron/c-rnn-gan.pytorch}{LINK} \\
        T-Forcing \cite{graves2013generating, sutskever2011generating} & \href{https://github.com/mojesty/professor_forcing/tree/master}{LINK} \\
        P-Forcing \cite{lamb2016professor} & \href{https://github.com/mojesty/professor_forcing/tree/master}{LINK} \\
        \bottomrule
    \end{tabular}
    \label{appendix:baselines:source_code_links}
\end{table}

\subsection{Diffusion Probabilistic Models \label{sec:prelim}}
In this section, we explain diffusion models following the work of \cite{sohl2015deep} and \cite{ho2020denoising}.
Diffusion models belong to a class of latent variable models formulated as $p_{\theta}(\vx^{(0)}) = \int p_{\theta}(\vx^{(0:T)}) \; d\vx^{(1:T)}$, where $\vx^{(0)}$ is a sample following the data distribution $q(\vx^{(0)})$ and $\{\vx^{(t)}\}_{t=1}^{T}$ are latent variables with the same dimensionality as $\vx^{(0)}$.

The \textit{forward process} is defined as a Markov chain that gradually adds Gaussian noise to $\vx^{(0)}$ via a sequence of variances $\left\{ \beta^{(t)}\right\}_{t=1}^T$:
\begin{equation}\label{forward}
    \resizebox{0.94\linewidth}{!}{$\displaystyle{q\big(\vx^{(1:T)} | \vx^{(0)} \big) = \prod_{t=1}^{T} q\big( \vx^{(t)} | \vx^{(t-1)} \big), \quad q\big( \vx^{(t)} | \vx^{(t-1)} \big) := \mathcal{N}\big(\vx^{(t)}; \sqrt{1 - \beta^{(t)}}\vx^{(t-1)}, \beta^{(t)}\mI \big).}$}
\end{equation}
The process successively converts data $\vx^{(0)}$ to white latent noise $\vx^{(T)}$.
The noisy sample $\vx^{(t)}$ can be obtained directly from the original sample $\vx^{(0)}$ by sampling from $q(\vx^{(t)} | \vx^{(0)}) = \mathcal{N}\big(\vx^{(t)} ; \sqrt{\bar{\alpha}^{(t)}} \vx^{(0)}, (1-\bar{\alpha}^{(t)})\mI \big)$, where $\alpha^{(t)} = 1 - \beta^{(t)}$ and $\bar{\alpha}^{(t)} = \prod_{i=1}^t \alpha^{(i)}$.

The \textit{reverse process} is the joint distribution $p_{\theta}(\vx^{(0:T)}) = p(\vx^{(T)}) \prod_{t=1}^T p_{\theta}(\vx^{(t-1)} | \vx^{(t)})$, which is a Markov chain that starts from white latent noise and gradually denoises noisy samples to generate synthetic samples:
\begin{align}\label{reverse}
    p_{\theta}\big(\vx^{(t-1)} | \vx^{(t)}\big) := \mathcal{N}\big(\vx^{(t-1)}; \bm{\mu}_{\theta}(\vx^{(t)}, t), \bm{\Sigma}_{\theta}(\vx^{(t)}, t)\big), \quad
    p\big(\vx^{(T)}\big) := \mathcal{N}\big(\vx^{(T)}; \mathbf{0}, \mI \big).
\end{align}

Under a specific parameterization described in \cite{ho2020denoising}, the training objective can be expressed as follows:
\begin{align}
\begin{split}
    \E_{\vx^{(0)}, \bm{\epsilon}} \Bigg[\frac{(\beta^{(t)})^2}{2(\sigma^{(t)})^2 \alpha^{(t)} ( 1 - \bar{\alpha}^{(t)})} \Big\|\bm{\epsilon} - \backbone\big(\sqrt{\bar{\alpha}^{(t)}}\vx^{(0)} + \sqrt{1 - \bar{\alpha}^{(t)}}\bm{\epsilon}, t\big) \Big\|^2 \Bigg] + C,
\end{split}
\end{align}
where $C$ is a constant that is not trainable.
Empirically, a neural network $\backbone$ is trained to approximate $\bm{\epsilon}$.
This $\bm{\epsilon}$-prediction objective resembles denoising score matching, and the sampling procedure resembles Langevin dynamics using $\backbone$ as an estimator of the gradient of the data distribution \cite{song2019generative, song2020improved}.

\subsection{Model Training and Hyperparameter Selection \label{appendix:detail:hyperparams}}
\subsubsection{Neural Controlled Differential Equation \label{appendix:detail:hyperparams:ncde}}
We attempted to use neural controlled differential equation (NCDE) \cite{kidger2020neural} as our architecture for $\backbone$.
We expect the continuous property of the NCDE to yield better results for time-series generation.
NCDE is formally defined as the following:
\begin{definition}
Suppose we have a time-series $\vs = \{(r_1, \vx_{1}),...,(r_n, \vx_{n})\}$ and $D$ is the dimensionality of the series.
Let $Y: [r_1, r_n] \rightarrow \R^{D+1}$ be the natural cubic spline with knots at $r_1,...,r_n$ such that $Y_{t_{i}} = (\vx_{i}, r_i)$.
$\vs$ is often assumed to be a discretization of an underlying process that is approximated by $Y$.
Let $f_{\theta} : \R^h \rightarrow \R^{h \times (D+1)}$ and $\zeta_{\theta}: \R^{D+1} \rightarrow \R^h$ be any neural networks, where $h$ is the size of hidden states. Let $z_{r_1} = \zeta_{\theta} (r_1,\vx_{1})$

The NCDE model is then defined to be the solution to the following CDE:
\begin{equation}
    z_r = z_{r_1} + \int_{r_1}^{r} f_{\theta}( z_s)\mathrm{d}Y_s \quad \text{for $r \in (r_1, r_n]$}
\end{equation}
where the integral is a Riemann–Stieltjes integral.
%Similar to RNN, the hidden state(s) of the model may be taken as the entire trajectory of $z$ or the final output $z_{r_n}$.
\end{definition}

However, we find that this approach suffers from high computational cost since it needs to calculate cubic Hermite spline and solve the CDE for every noisy sample input during training.
It thus has low scalability for generating time-series data with long sequences.
Nevertheless, we believe this direction is worth exploring for future research.

\subsubsection{\ours Training \label{appendix:detail:hyperparams:training}}
The diffusion model is trained using $\mathcal{L}_{\text{train}}$ in \Cref{method:loss}.
We set $\lambda$ to 0.01.
We use cosine scheduling \cite{nichol2021improved} for the variances $\left\{ \beta^{(t)}\right\}_{t=1}^T$.
We apply the exponential moving average to model parameters with a decay rate of 0.995. We use Adam optimizer \cite{KingBa15} with a learning rate of 0.00008, $\beta_1 = 0.9$, and $\beta_2 = 0.99$.
We set the total diffusion step $T$ to be 1000, accumulate the gradient for every 2 steps, use 2 layers for the BRNN, and use a batch size of 32 across all our experiments.

\subsubsection{Baselines \label{appendix:detail:hyperparams:baselines}}
For a fair comparison, we use a 2-layer RNN with a hidden dimension size of four times the number of input features.
We utilize the LSTM as our architecture whenever applicable.
We use a hidden dimension size of 256 for the eICU dataset.

For deterministic models such as the T-Forcing and P-Forcing, we uniformly sample the initial data vector from the real training data.
We subsequently use the initial data vector as an input to the deterministic models to generate the synthetic sequence by unrolling.

For stochastic process diffusion, we set \textit{gp\_sigma} to be 0.1 for Gaussian process (GP) and \textit{ou\_theta} to be 0.5 for Ornstein-Uhlenbeck (OU) process.
For discrete diffusion, we set the total diffusion step at 1000.
We use Adam optimizer with a learning rate of 0.00001 and batch size of 32 across all the experiments.

\subsubsection{Software \label{appendix:detail:hyperparams:software}}
We set the seed to 2023 and used the following software for our experiments (\Cref{software}).
\begin{table}[hbt!]
    \caption{Software packages.}
    \label{software}
    %\vspace{-10pt}
    \centering
    \begin{tabular}{@{}ll@{}}
        \toprule
        Method & Software \\
        \midrule
        \ours & \href{https://pypi.org/project/torch/}{PyTorch 2.0.1} \\
        EHR-M-GAN \cite{li2023generating} & \href{https://pypi.org/project/tensorflow-gpu/1.14.0/}{TensorFlow 1.14.0} \\
        DSPD/CSPD (GP or OU) \cite{pmlr-v202-bilos23a} & \href{https://pypi.org/project/torch/}{PyTorch 2.0.1} \\
        GT-GAN \cite{jeon2022gtgan} & \href{https://pypi.org/project/torch/2.0.0/}{PyTorch 2.0.0} \\
        TimeGAN \cite{yoon2019timegan} & \href{https://pypi.org/project/tensorflow/1.10.0/}{TensorFlow 1.10.0} \\
        RCGAN \cite{esteban2017realvalued} & \href{https://pypi.org/project/tensorflow/1.10.0/}{TensorFlow 1.10.0} \\
        C-RNN-GAN \cite{mogren2016crnngan} & \href{https://pypi.org/project/torch/}{PyTorch 2.0.1} \\
        T-Forcing \cite{graves2013generating, sutskever2011generating} & \href{https://pypi.org/project/torch/1.0.0/}{PyTorch 1.0.0} \\
        P-Forcing \cite{lamb2016professor} & \href{https://pypi.org/project/torch/1.0.0/}{PyTorch 1.0.0} \\
        \bottomrule
    \end{tabular}
    %\label{appendix:baselines:source_code_links}
\end{table}

\subsection{Evaluation Metrics \label{appendix:detail:evaluate}}
\subsubsection{t-SNE \label{appendix:detail:evaluate:tsne}}
We note that t-SNE is qualitative in nature and therefore subject to observer bias, but despite this limitation it can still provide a general visual guide to the usefulness of a method, and is used in related papers \cite{Yoon2023EHRSafeGH}. We perform the hyperparameter search on the number of iterations, learning rate, and perplexity to optimize the performance of t-SNE \cite{wattenberg2016use}. %The result of t-SNE visualizations could be largely affected by its Hyperparameters \cite{wattenberg2016use}. Thus, we tried different combinations of the number of iterations, learning rate, and perplexity.
We use 300 iterations, perplexity 30, and scaled learning rate \cite{belkina2019automated}.
We flatten the input data along the feature dimension, perform standardization, and then apply t-SNE directly to the data without using any summary statistics.
We uniformly randomly select 2000 samples from the synthetic, real training, and real testing data for t-SNE visualizations on the eICU, MIMIC-III, MIMIC-IV, and Energy datasets.
%Due to the large sample size and inconsistent number of samples across training and testing data, we do not visualize all the samples with t-SNE. Instead, for eICU, MIMIC-IV, and MIMIC-III datasets, we uniformly randomly select 2000 samples from synthetic, real training, and real testing data for t-SNE visualization.
For the %HiRID and 
Stocks dataset, we use 1000 and 700 samples, respectively, due to the limited size of real testing data.
%Compared with the t-SNE visualization approach used by TimeGAN and GT-GAN, we do not take the mean over the feature dimension for visualization.
%While this technique can speed up the computing time for t-SNE, we believe it could be biased and has less focus on individual features' values.

\subsubsection{Discriminative and Predictive Scores \label{appendix:detail:evaluate:discripred}}
To ensure consistency with results obtained from TimeGAN and GT-GAN, we adopt the same source code from TimeGAN for calculating discriminative scores.
We train a GRU time-series classification model to distinguish between real and synthetic samples, and $|0.5 - \text{Accuracy}|$ is used as the score.
% Both synthetic and real datasets are split into training and testing sets.
% The training set is used to train a 2-layer 
% The training set are fed into a given 2-layer GRU time-series classification model to distinguish between real and synthetic samples. 
% One distinction between our usage and TimeGAN is that we standardize healthcare datasets instead of normalizing for discriminative scores.
% Since there was no standard method of determining minimum and maximum values for different features for both synthetic and real samples, we chose to standardize the samples.

For predictive scores, we use the implementation from GT-GAN, which computes the mean absolute error based on the next step \textit{vector} prediction (see Appendix D of the GT-GAN paper \cite{jeon2022gtgan}).
For consistency, we compute the predictive scores for the Stocks and Energy datasets by employing the implementation from TimeGAN that calculates the error for the next step \textit{scalar} prediction.
%However, given that the previous results for Stocks and Energy datasets from TimeGAN and GT-GAN use the scalar prediction objective, for the sake of consistency, we adopt the same objective for our results on these datasets as well.
%We believe utilizing the vector prediction objective can incorporate more information about the fidelity of the synthetic multivariate time series.
We apply standardization to the inputs of the discriminator and predictor and use linear activation for the predictor for all EHR datasets.

\subsubsection{In-hospital Mortality Prediction \label{appendix:detail:evaluate:mortality}}
\textbf{Train on Synthetic, Test on Real (TSTR):}
%We selected six commonly used machine learning (ML) algorithms --- XGBoost, Random Forest, AdaBoost, $\ell_1$ and $\ell_2$ regularized Logistic Regression --- and trained them entirely on synthetic EHR time-series and in-hospital mortality outcomes generated by \ours.
We use the default hyperparameters for the six ML models described in the Results section using the scikit-learn software package. %\Cref{sec:results}
The models are trained using two input formats: (1) raw multivariate time-series data flattened along the feature dimension; (2) summary statistics for each feature (the first record since ICU admission, minimum, maximum, record range, mean, standard deviation, mode, and skewness).
After training, the models are evaluated on real testing data in terms of AUC.
%To compare the amount of degradation in performance for ML models trained on synthetic data, we compare the AUC for models trained on real training data and evaluated on real testing data (namely TRTR score).

\textbf{Train on Synthetic and Real, Test on Real (TSRTR):}
To evaluate the effect of the increased proportion of the synthetic samples for training on model performance, we uniformly randomly sample 2,000 real training data from our training set and use this subset to train \ours.
After the training of \ours is complete, we subsequently add different amounts of the synthetic samples to the 2,000 real samples to train ML models for in-hospital mortality prediction.
We set the synthetic percentages to be 0.1, 0.3, 0.5, 0.7, 0.9.
%To evaluate whether a greater proportion of synthetic samples could degrade ML models' performances, we fixed the sample size for real training data at 2000 and trained multiple ML models using different percentages of added synthetic samples.
In other words, the ML models are trained with at most 20,000 samples (18,000 synthetic and 2,000 real).
This evaluation also simulates the scenario where synthetic samples from \ours are used for data augmentation tasks in medical applications.
Similar to computing the TSTR score, we train the ML models using either raw time-series data or summary statistics of each feature as the input.
Results obtained using summary statistics as the input are presented in \Cref{appendix:add:tstr}.

\subsubsection{Nearest Neighbor Adversarial Accuracy Risk (NNAA) \label{appendix:privacy}}
We calculate the NNAA risk score \cite{yale2020generation} by using the implementation from \href{https://github.com/yy6linda/synthetic-ehr-benchmarking/blob/main/privacy_evaluation/Synthetic_risk_model_nnaa.py}{this repository}.
Similar to performing t-SNE, we flatten the data along the feature dimension and apply standardization for preprocessing.
The scaled datasets are then used to calculate the NNAA risk score.

For reference, we describe the components of the NNAA score below.
\begin{definition}
    Let $S_T = \{x_{T}^{(1)},...,x_{T}^{(n)} \}$, $S_E = \{x_{E}^{(1)},...,x_{E}^{(n)} \}$ and $S_S = \{x_{S}^{(1)},...,x_{S}^{(n)} \}$ be data samples with size $n$ from real training, real testing, and synthetic datasets, respectively. The NNAA risk is the difference between two accuracies:
    \begin{align}
        \text{NNAA} &= \aatest - \aatrain, \\
        \aatest = \frac{1}{2} \Bigg(\frac{1}{n} \sum_{i=1}^n \mathbb{I}\big\{d_{\text{ES}}(i) &> d_{\text{EE}}(i)\big\} + \frac{1}{n} \sum_{i=1}^n \mathbb{I}\big\{d_{\text{SE}}(i) > d_{\text{SS}}(i) \big\} \Bigg),\\
        \aatrain = \frac{1}{2} \Bigg(\frac{1}{n} \sum_{i=1}^n \mathbb{I}\big\{d_{\text{TS}}(i) &> d_{\text{TT}}(i)\big\} + \frac{1}{n} \sum_{i=1}^n \mathbb{I}\big\{d_{\text{ST}}(i) > d_{\text{SS}}(i) \big\} \Bigg),
        \label{appendix:nnaa:decomposition}
    \end{align}
    where $\mathbb{I}\{\cdot\}$ is the indicator function, and
    \begin{align}
        d_{\text{TS}}(i) &= \min_{j} \left\|x_{T}^{(i)} - x_{S}^{(j)} \right\|, \quad d_{\text{ST}}(i) = \min_{j} \left\|x_{S}^{(i)} - x_{T}^{(j)} \right\|, \\
        d_{\text{ES}}(i) &= \min_{j} \left\|x_{E}^{(i)} - x_{S}^{(j)} \right\|, \quad d_{\text{SE}}(i) = \min_{j} \left\|x_{S}^{(i)} - x_{E}^{(j)} \right\|, \\
        d_{\text{TT}}(i) = \min_{j, j \neq i} \left\|x_{T}^{(i)} - x_{T}^{(j)} \right\|&, \quad d_{\text{SS}}(i) = \min_{j, j\neq i} \left\|x_{S}^{(i)} - x_{S}^{(j)} \right\|, \quad d_{\text{EE}}(i) = \min_{j, j\neq i} \left\|x_{E}^{(i)} - x_{E}^{(j)} \right\|.
    \end{align}
\end{definition}
In our experiments, there are instances where $\aatrain > \aatest$. To consistently obtain positive values, we use $\text{NNAA} = |\aatest - \aatrain|$ for our evaluations.

%We cite justification for using this metric \cite{yale2020generation}: intuitively, if the generator performs well, then the adversarial classifier cannot distinguish between generated data and real data, train and test adversarial accuracy should both be 0.5, and the privacy loss will be zero. If G does a poor job and underfits, it will serve generated data that does not resemble real data. Thus the adversarial classifier will have no problem classifying real vs. artificial so the train and test adversarial accuracy will both be high ( $>$ 0.5) and similar, and the privacy loss will also be near zero. In this last case, privacy is preserved but the utility of the data may be low. If the generator overfits the training data, the Train AA will be near zero (good training resemblance), but the Test AA will be around 0.5 (poor test resemblance). Thus the privacy loss will be high (near 0.5).

\subsubsection{Membership Inference Risk (MIR)}
Our implementation of the MIR score \cite{liu2019socinf} follows the source code in \href{https://github.com/yy6linda/synthetic-ehr-benchmarking/blob/main/privacy_evaluation/Synthetic_risk_model_mem.py}{this repository}.
To keep a similar scale of the distance across different datasets, we apply normalization on the computed distances so that they are in the [0,1] range.
We use a threshold of 0.08 for the MIMIC-IV and MIMIC-III %, and HiRID 
datasets.
We set the decision threshold to 0.005 for eICU.
All the input data is normalized to the [0,1] range.
\newpage
\section{Additional Experiments \label{appendix:add}}
%\clearpage
\subsection{Real and Synthetic Time Series Visualizations \label{appendix:add:visual}}
\begin{figure}[hbt!]
    \includegraphics[width=\linewidth]{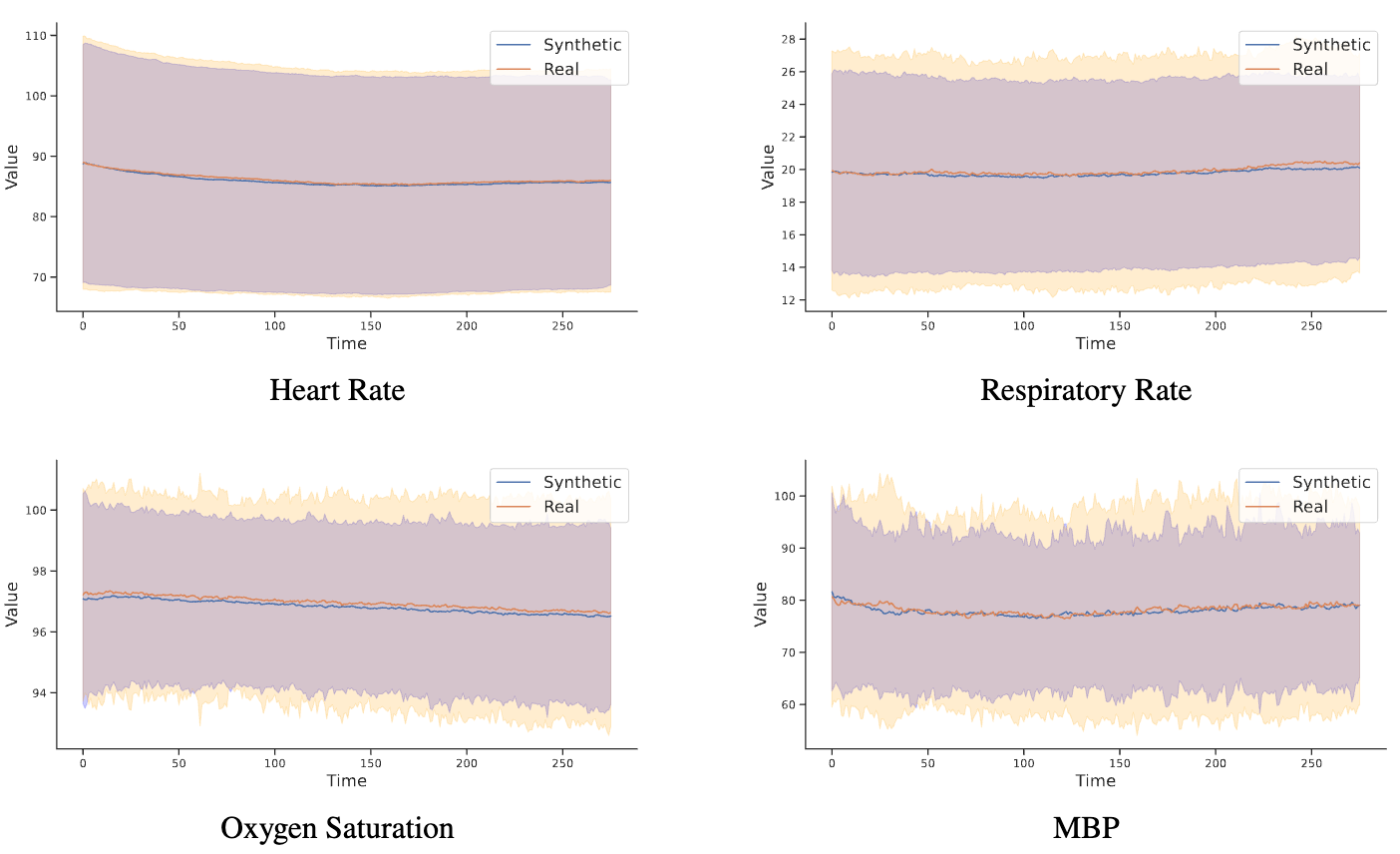}
    \caption{eICU, where the mean is the solid line and $\pm$ one standard deviation is the shaded area.}
\end{figure}

\begin{figure}[hbt!]
    \includegraphics[width=\linewidth]{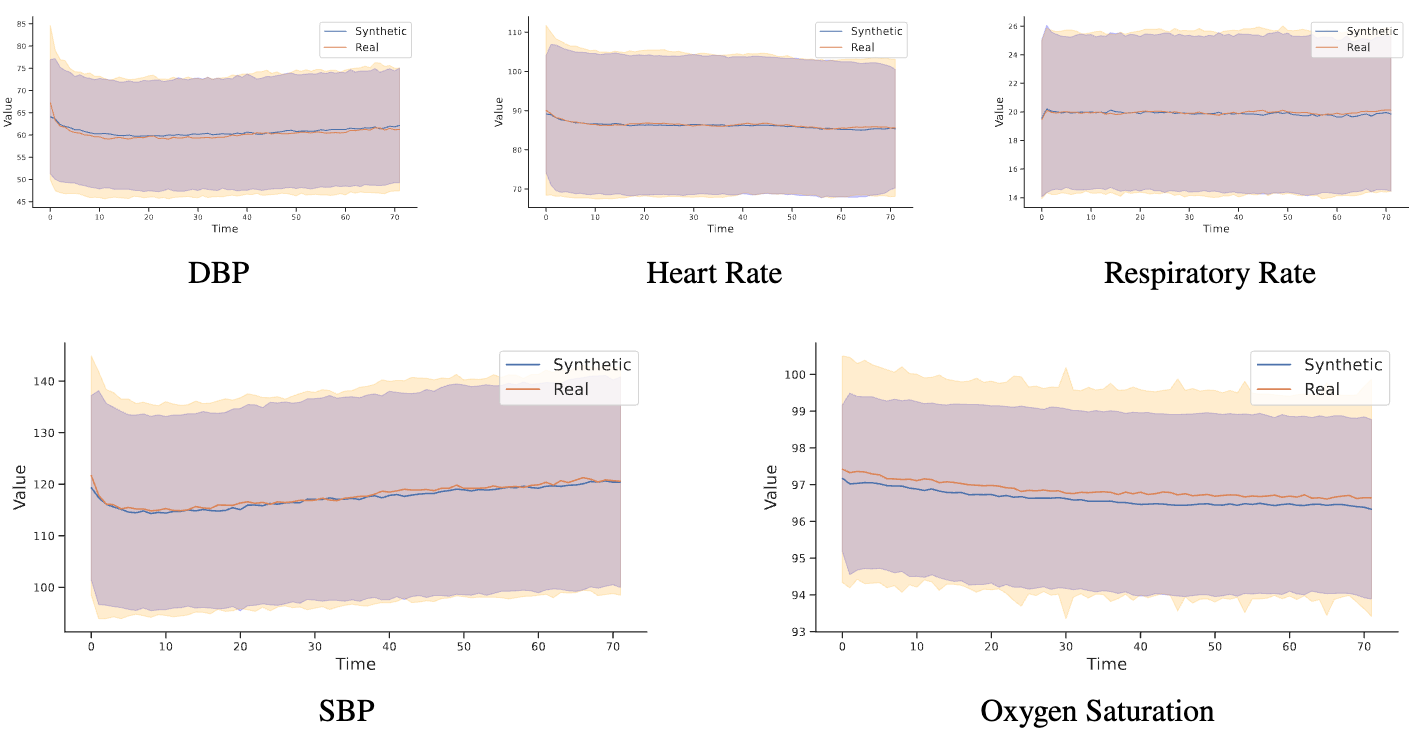}
    \caption{MIMIC-IV, where the mean is the solid line and $\pm$ one standard deviation is the shaded area.}
\end{figure}

\begin{figure}[hbt!]
    \includegraphics[width=\linewidth]{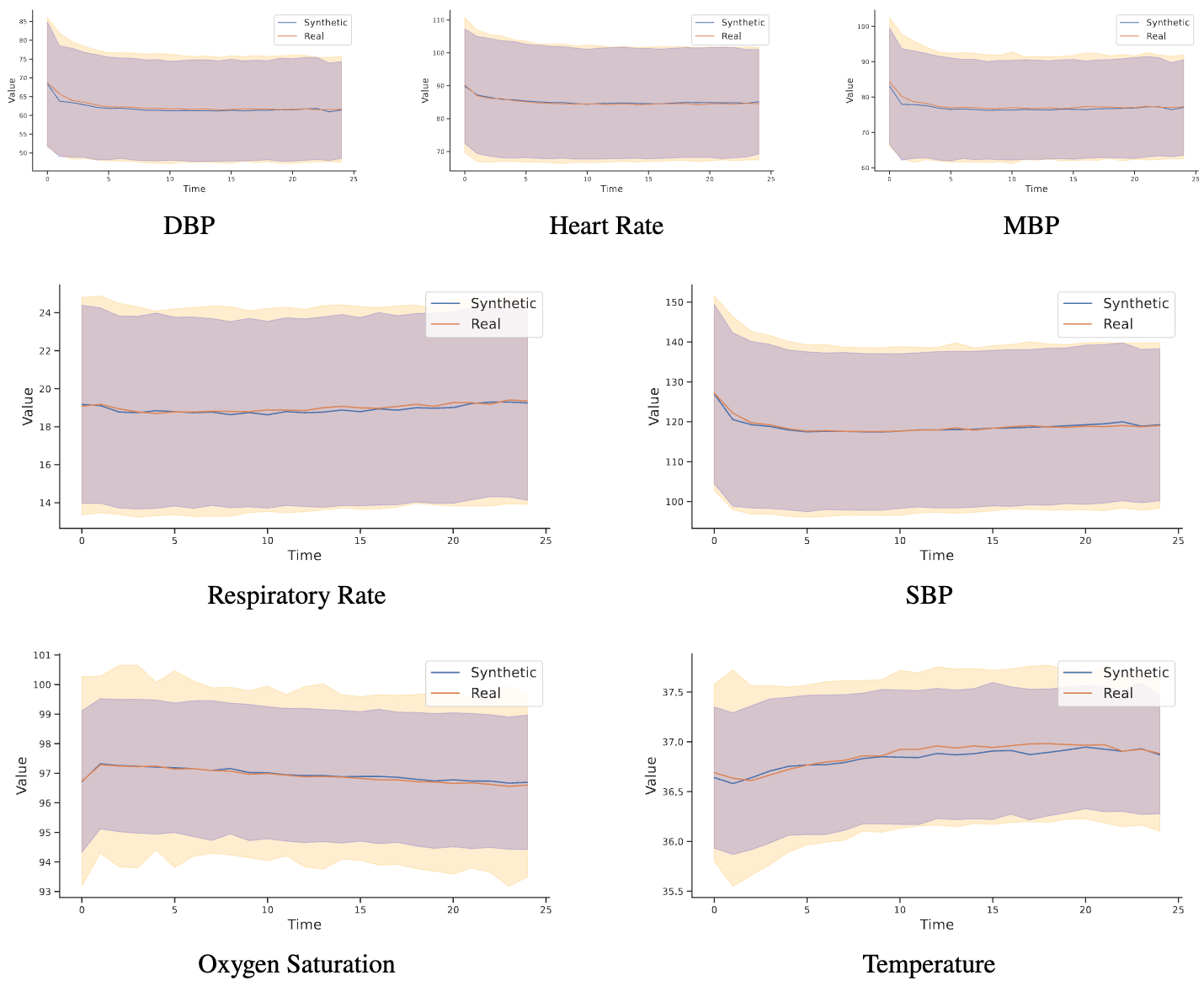}
    \caption{MIMIC-III, where the mean is the solid line and $\pm$ one standard deviation is the shaded area.}
\end{figure}

\begin{comment}
\begin{figure}[hbt!]
\begin{subfigure}{0.33\linewidth}
    \centering
    \includegraphics[width=\linewidth]{figs/time-series/trajectory_hirid_cvp.pdf}
    \caption*{CVP}
\end{subfigure}%
\begin{subfigure}{0.33\linewidth}
    \centering
    \includegraphics[width=\linewidth]{figs/time-series/trajectory_hirid_dbp.pdf}
    \caption*{DBP}
\end{subfigure}
\begin{subfigure}{0.33\linewidth}
    \centering
    \includegraphics[width=\linewidth]{figs/time-series/trajectory_hirid_heartrate.pdf}
    \caption*{Heart Rate}
\end{subfigure}\\
\begin{subfigure}{0.5\linewidth}
    \centering
    \includegraphics[width=\linewidth]{figs/time-series/trajectory_hirid_map.pdf}
    \caption*{MBP}
\end{subfigure}%
\begin{subfigure}{0.5\linewidth}
    \centering
    \includegraphics[width=\linewidth]{figs/time-series/trajectory_hirid_sbp.pdf}
    \caption*{SBP}
\end{subfigure}\\
\begin{subfigure}{0.5\linewidth}
    \centering
    \includegraphics[width=\linewidth]{figs/time-series/trajectory_hirid_spo2.pdf}
    \caption*{Oxygen Saturation}
\end{subfigure}%
\begin{subfigure}{0.5\linewidth}
    \centering
    \includegraphics[width=\linewidth]{figs/time-series/trajectory_hirid_st.pdf}
    \caption*{ST Elevation}
\end{subfigure}%
\caption{HiRID, where the mean is the solid line and $\pm$ one standard deviation is the shaded area.}
\end{figure}
\end{comment}

\begin{figure}[h!]
    \centering
    \includegraphics[width=\linewidth]{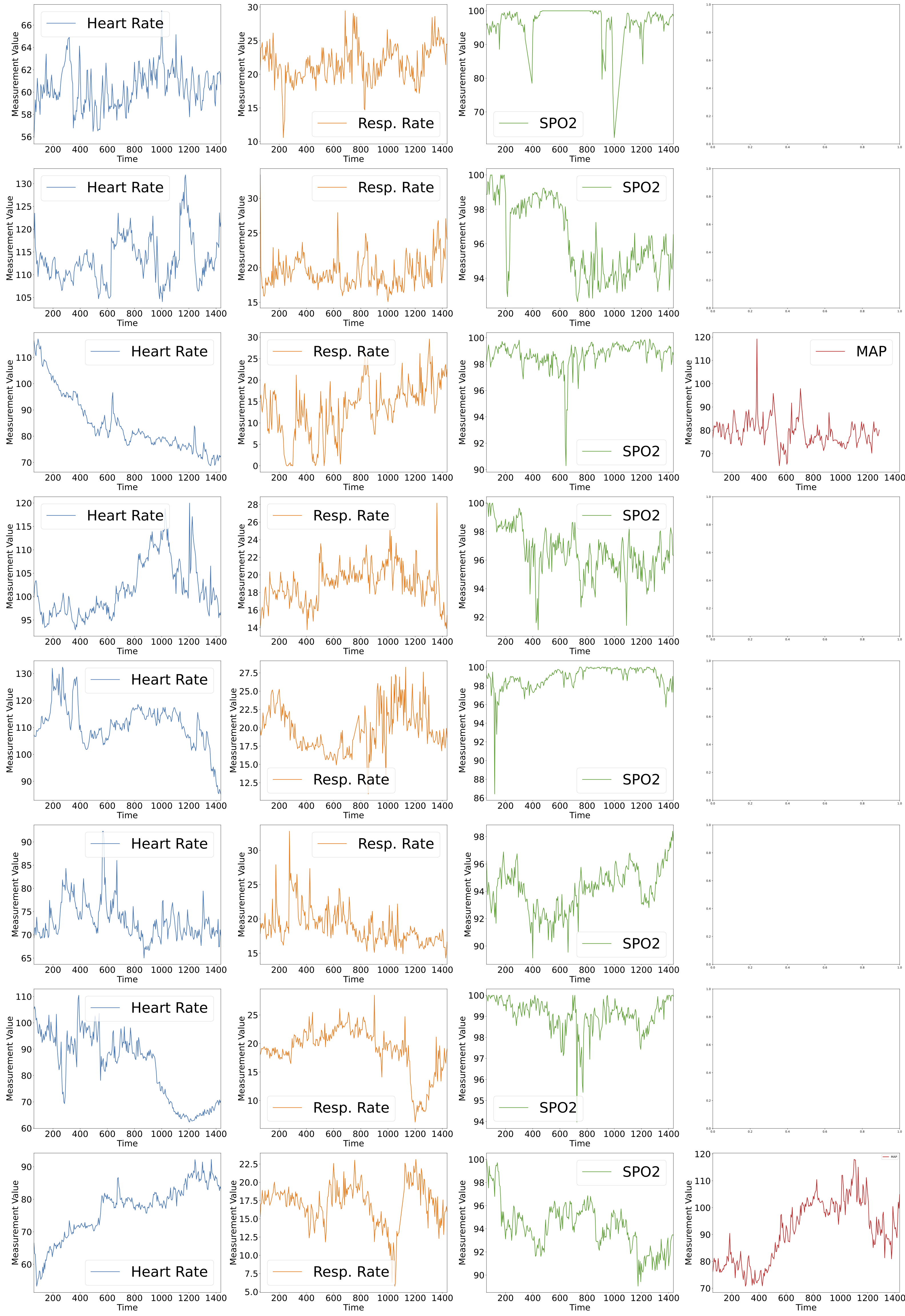}
    \caption*{eICU: synthetic time series produced by \ours.}
\end{figure}
\begin{figure}[h!]
    \centering
    \includegraphics[width=\linewidth]{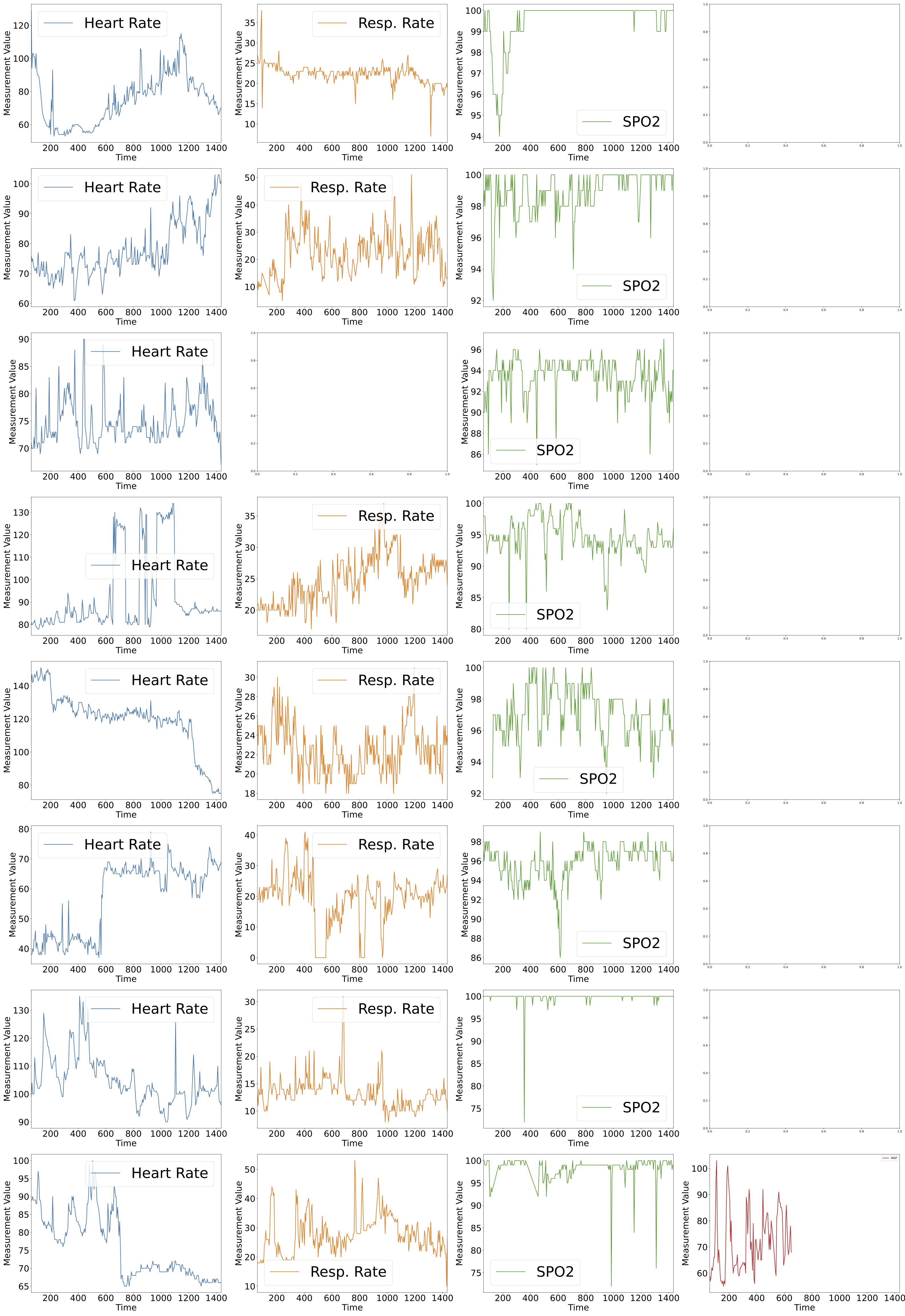}
    \caption*{eICU: time series in real testing data.}
\end{figure}

\begin{figure}[h!]
    \centering
    \includegraphics[width=\linewidth]{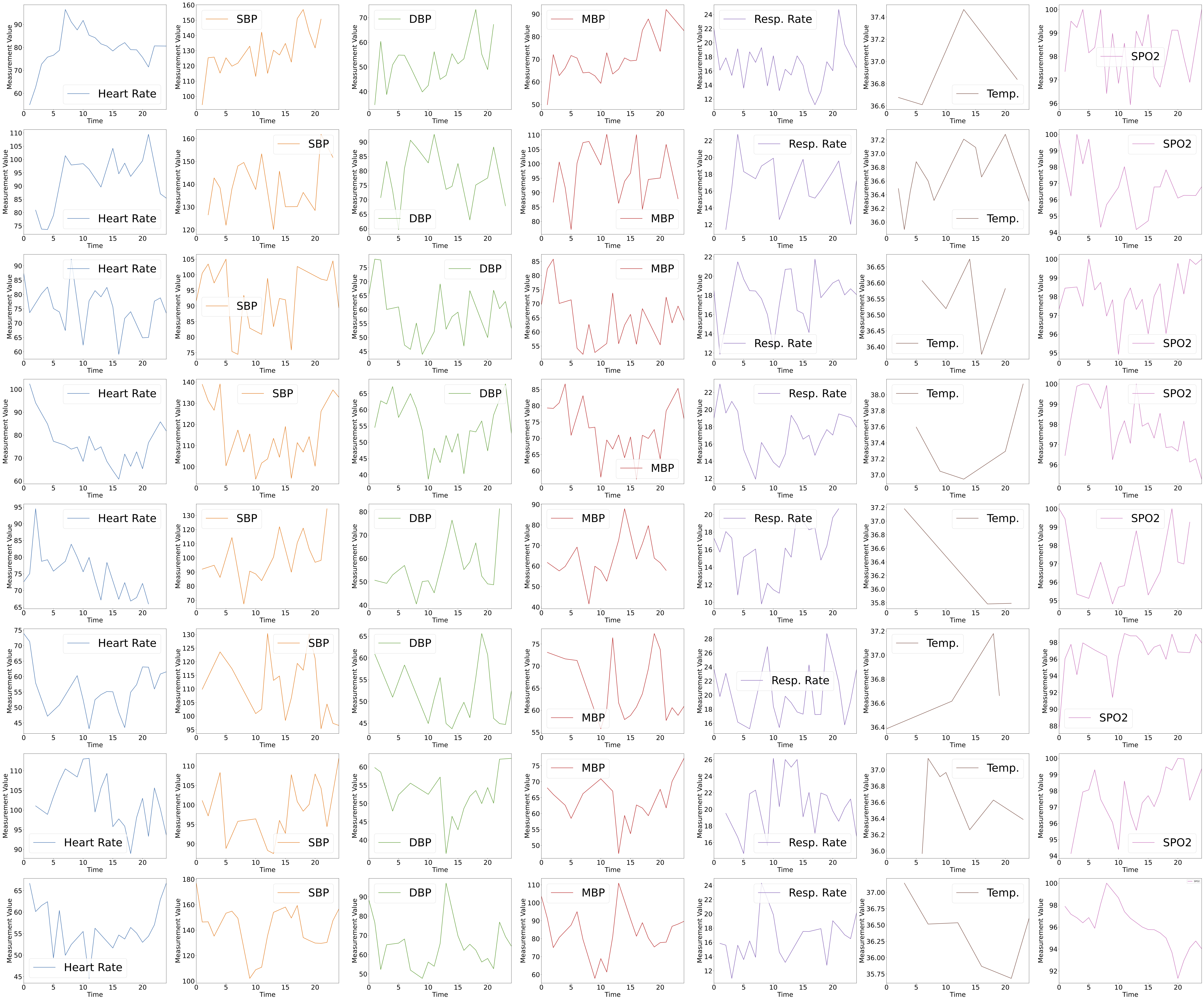}
    \caption*{MIMIC-III: synthetic time series produced by \ours.}
\end{figure}
\begin{figure}[h!]
    \centering
    \includegraphics[width=\linewidth]{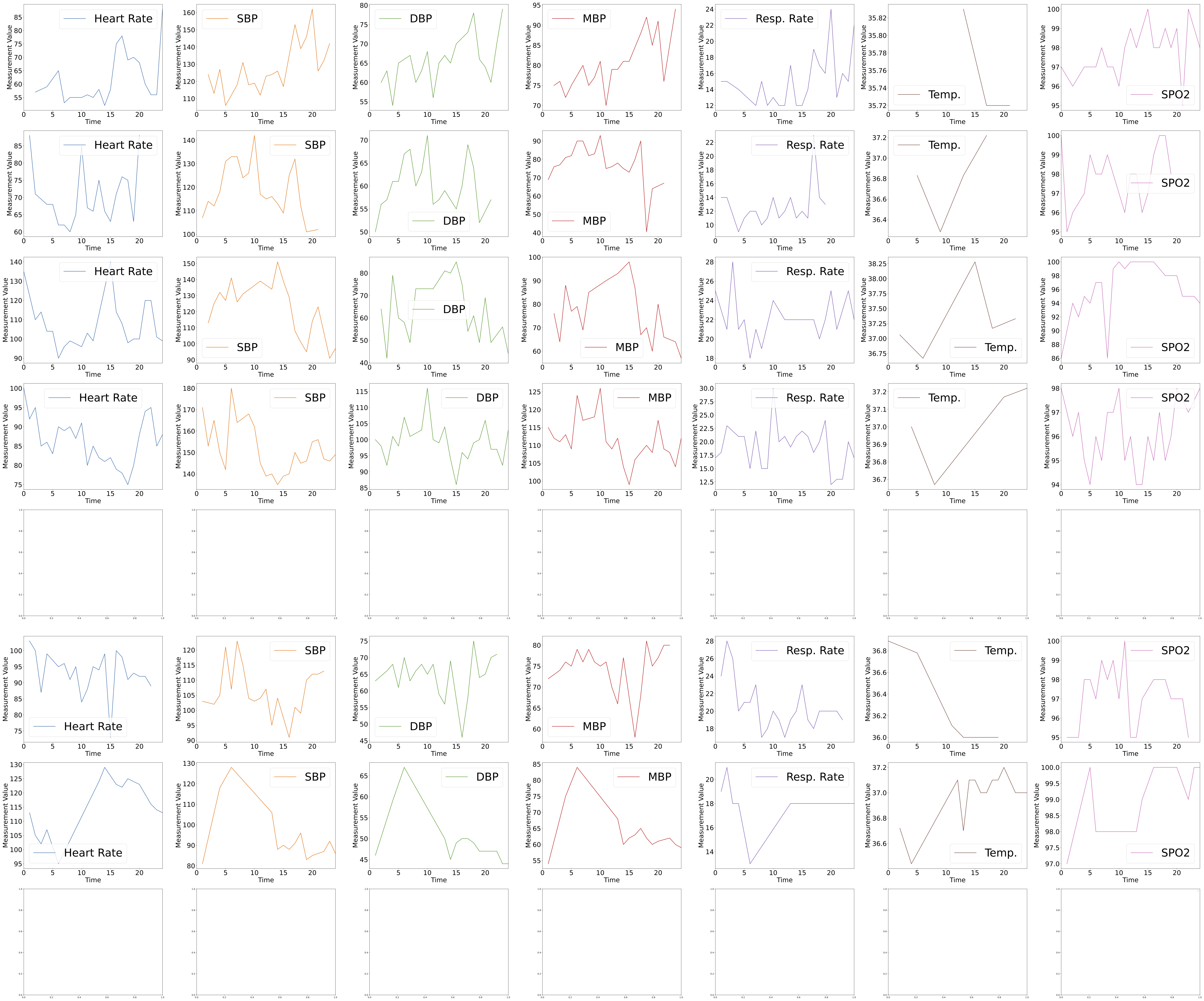}
    \caption*{MIMIC-III: time series in real testing data.}
\end{figure}

\begin{figure}[h!]
    \centering
    \includegraphics[width=\linewidth]{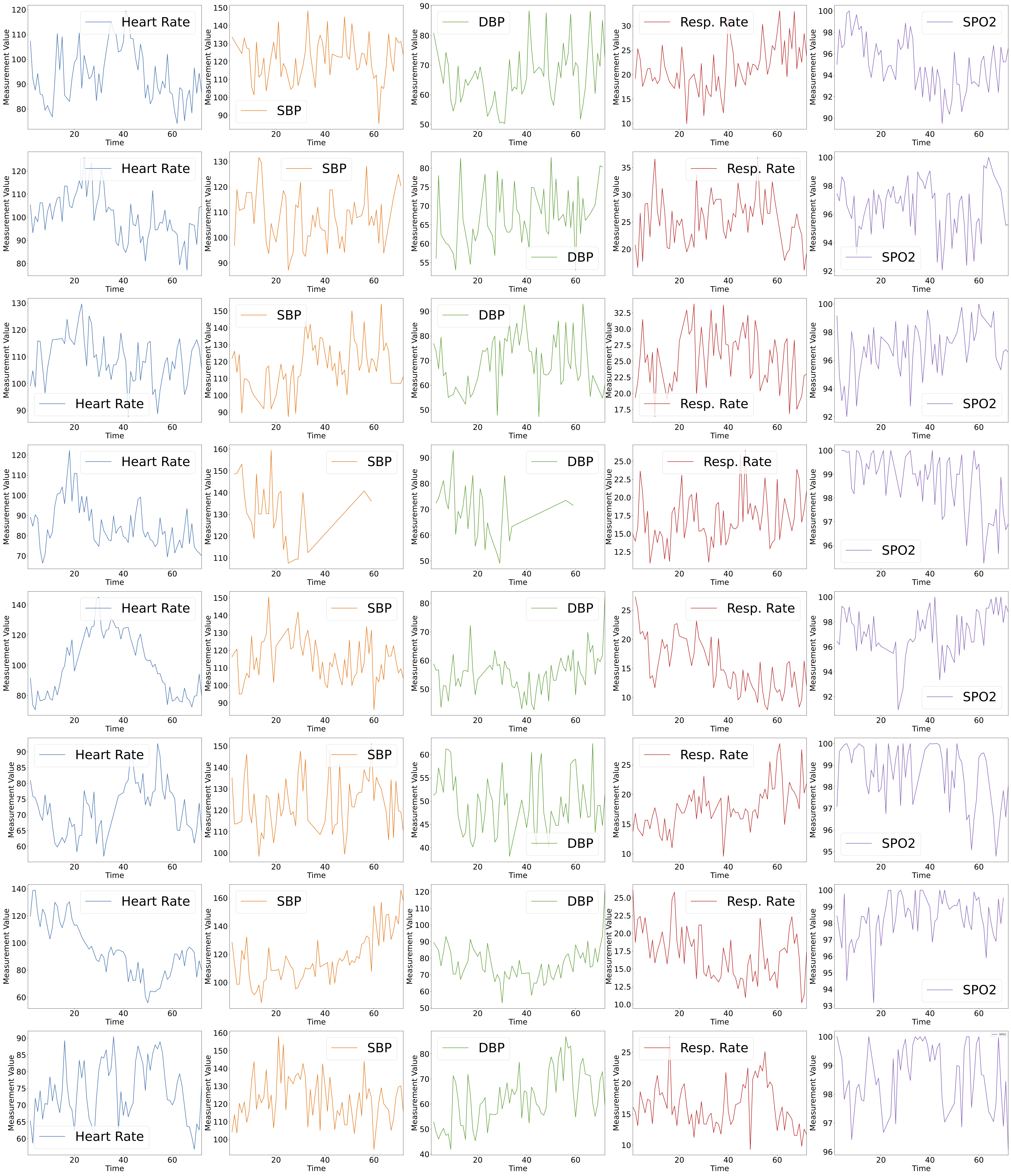}
    \caption*{MIMIC-IV: synthetic time series produced by \ours.}
\end{figure}
\begin{figure}[h!]
    \centering
    \includegraphics[width=\linewidth]{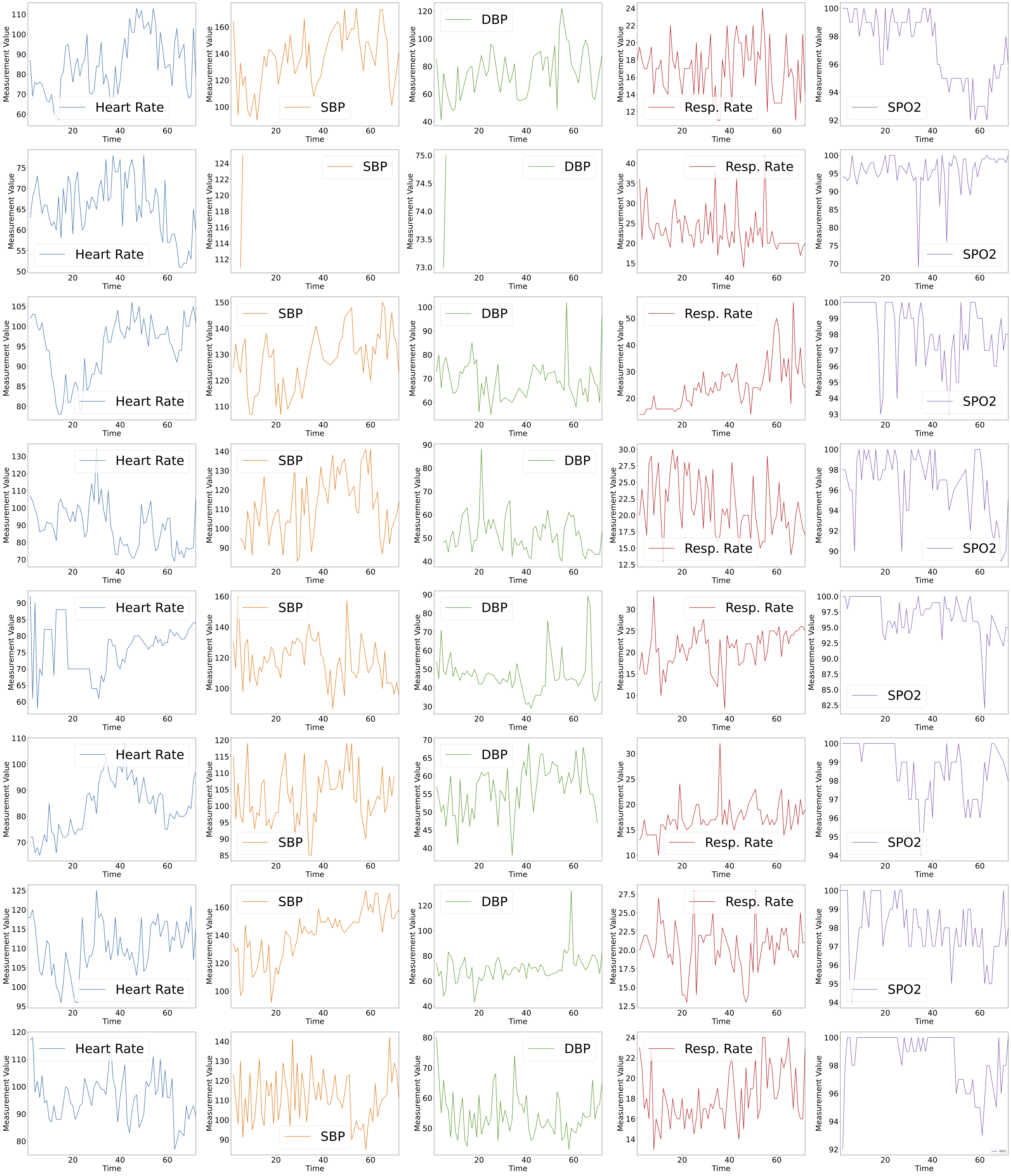}
    \caption*{MIMIC-IV: time series in real testing data.}
\end{figure}

\begin{comment}
\begin{figure}[h!]
    \centering
    \includegraphics[width=\linewidth]{figs/time-series/visualize_hirid_fake.pdf}
    \caption*{HiRID: synthetic time series produced by \ours.}
\end{figure}
\begin{figure}[h!]
    \centering
    \includegraphics[width=\linewidth]{figs/time-series/visualize_hirid_real.pdf}
    \caption*{HiRID: time series in real testing data.}
\end{figure}
\end{comment}

\FloatBarrier
\subsection{t-SNE Visualizations \label{appendix:add:tsne}}
In this section, we present our visualizations for all the baselines in our experiments.
For all the figures, synthetic samples are in \textcolor{blue}{\textbf{blue}}, real samples in train split are in \textcolor{red}{\textbf{red}}, and real samples in test split are in \textcolor{orange}{\textbf{orange}}.
We discuss our procedure for t-SNE visualizations in Supplementary Material \ref{appendix:detail:evaluate:tsne}.

\begin{figure}[hbt!]
  \centering
  \includegraphics[width=\textwidth]{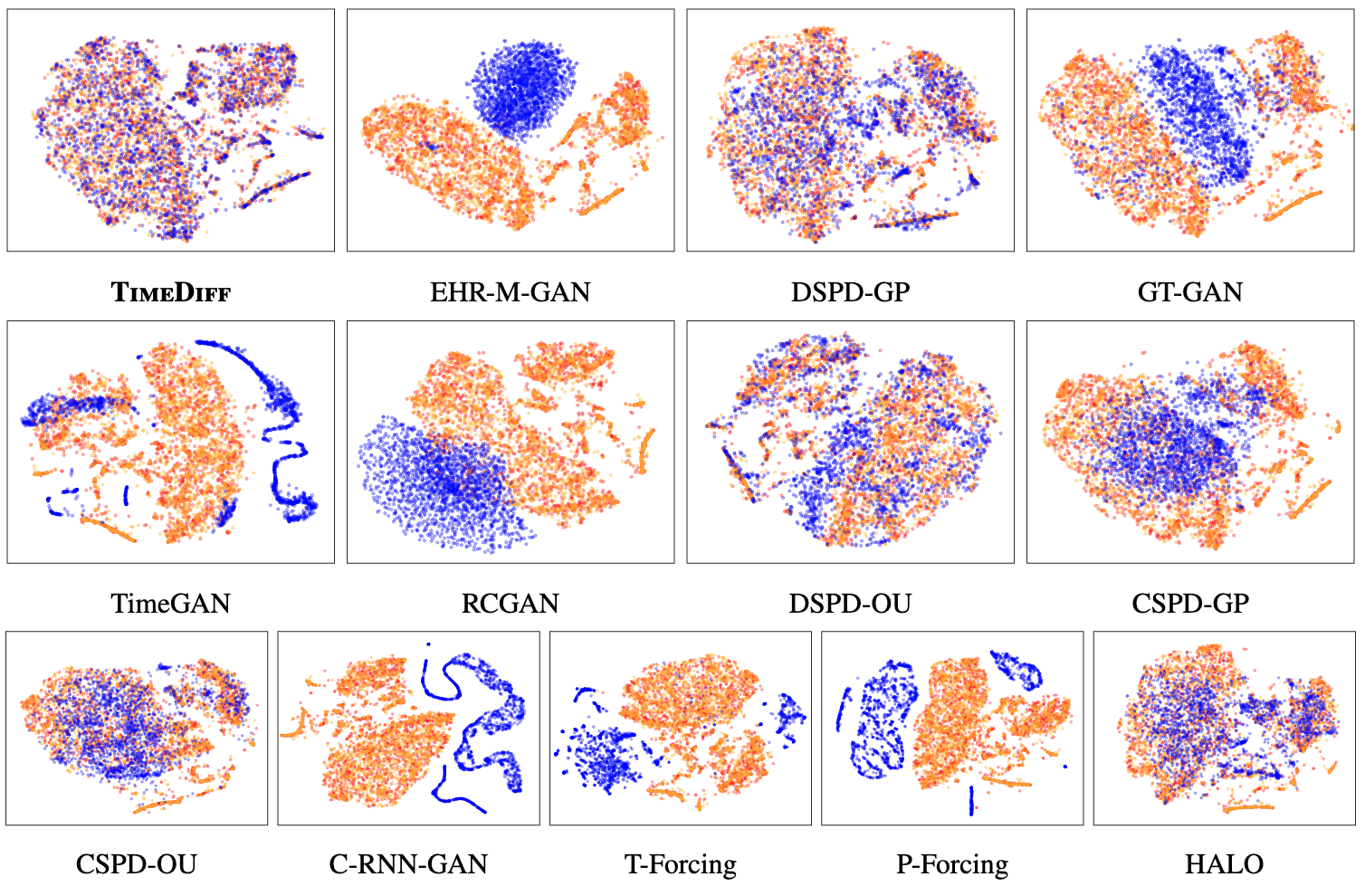}
  \caption{\firstedit{t-SNE dimension reduction visualization for eICU.}}
\end{figure}

\begin{figure}[hbt!]
  \centering
  \includegraphics[width=\textwidth]{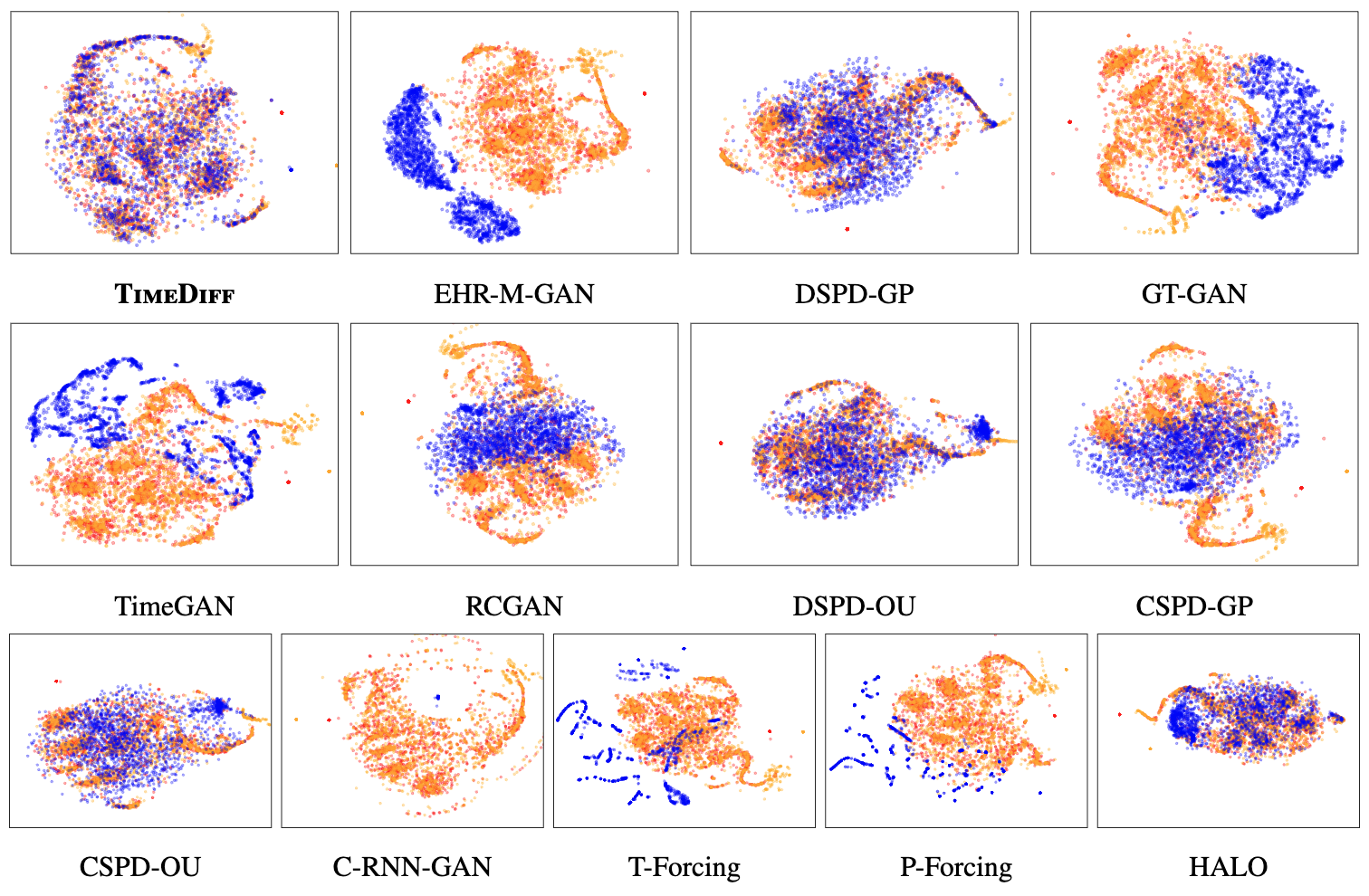}
  \caption{\firstedit{t-SNE dimension reduction visualization for MIMIC-III.}}
\end{figure}

\begin{comment}
\begin{figure}[hbt!]
    \centering
    \AppendixTSNEHiRID
    \caption{t-SNE dimension reduction visualization for HiRID.}
\end{figure}
\end{comment}

\begin{figure}[hbt!]
  \centering
  \includegraphics[width=\textwidth]{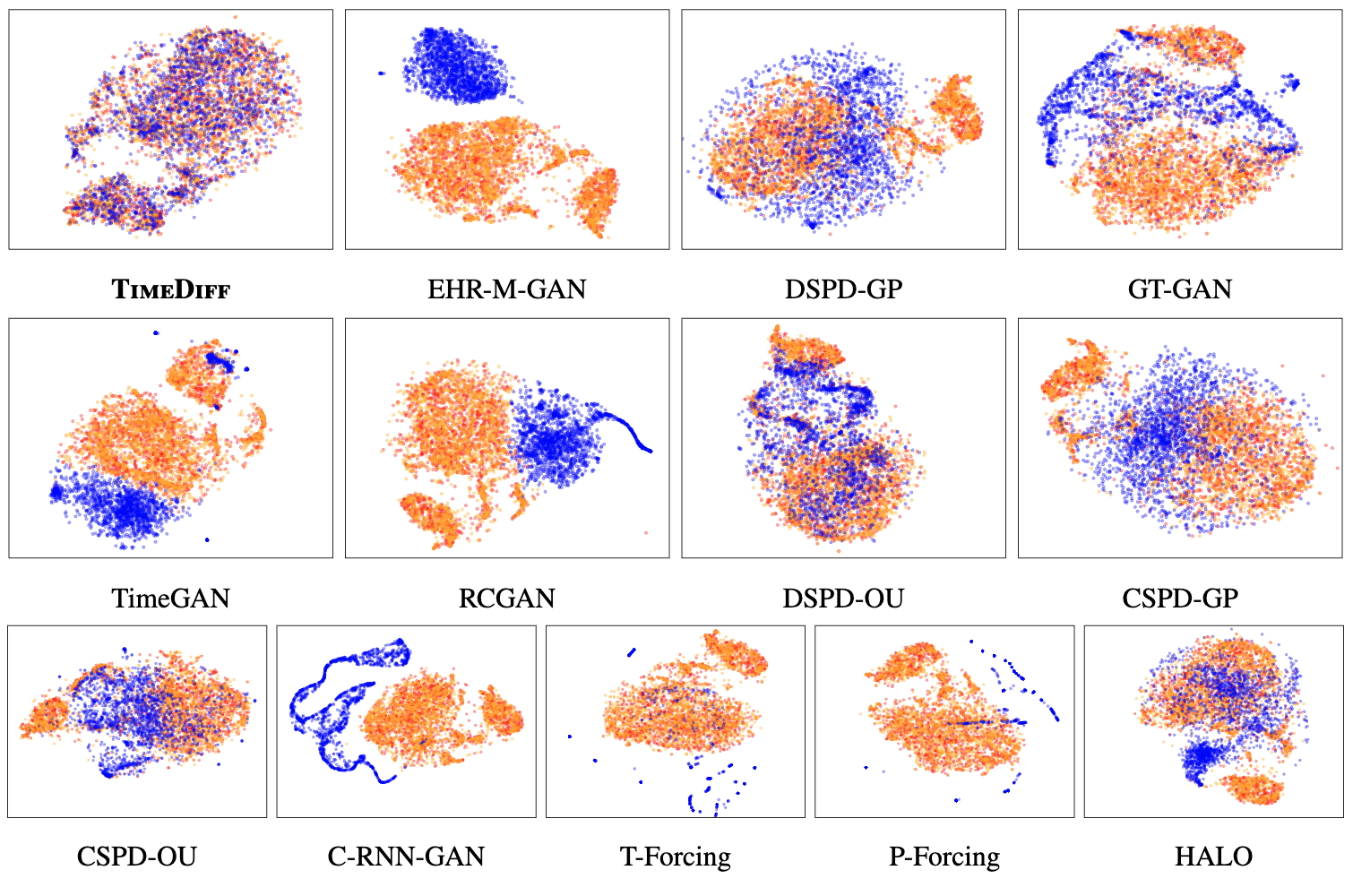}
  \caption{\firstedit{t-SNE dimension reduction visualization for MIMIC-IV.}}
\end{figure}

\begin{figure}[hbt!]
    \centering
    \includegraphics[width=\textwidth]{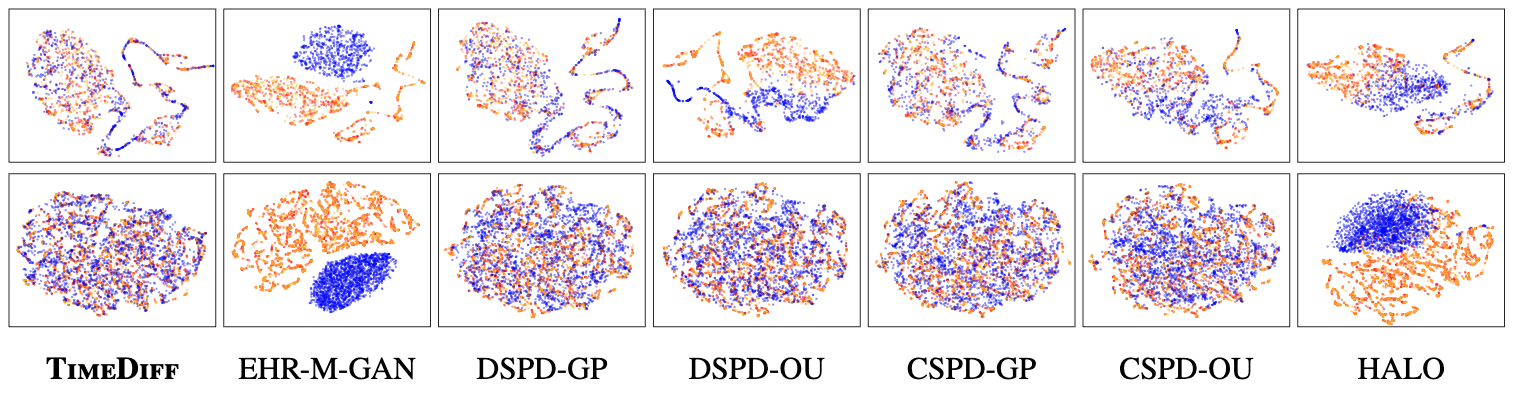}
    \caption{\firstedit{t-SNE dimension reduction visualization for non-EHR datasets. The first row is from Stocks and the second row is from Energy.}}
\end{figure}

\FloatBarrier
\firstedit{\subsection{UMAP Visualizations \label{appendix:add:umap}}
In this section, we present our visualizations for all the baselines in our experiments.
For all the figures, synthetic samples are in \textcolor{blue}{\textbf{blue}}, real samples in train split are in \textcolor{red}{\textbf{red}}, and real samples in test split are in \textcolor{orange}{\textbf{orange}}.
We follow the same procedure as t-SNE for data preprocessing.}

\begin{figure}[hbt!]
  \centering
  \includegraphics[width=\textwidth]{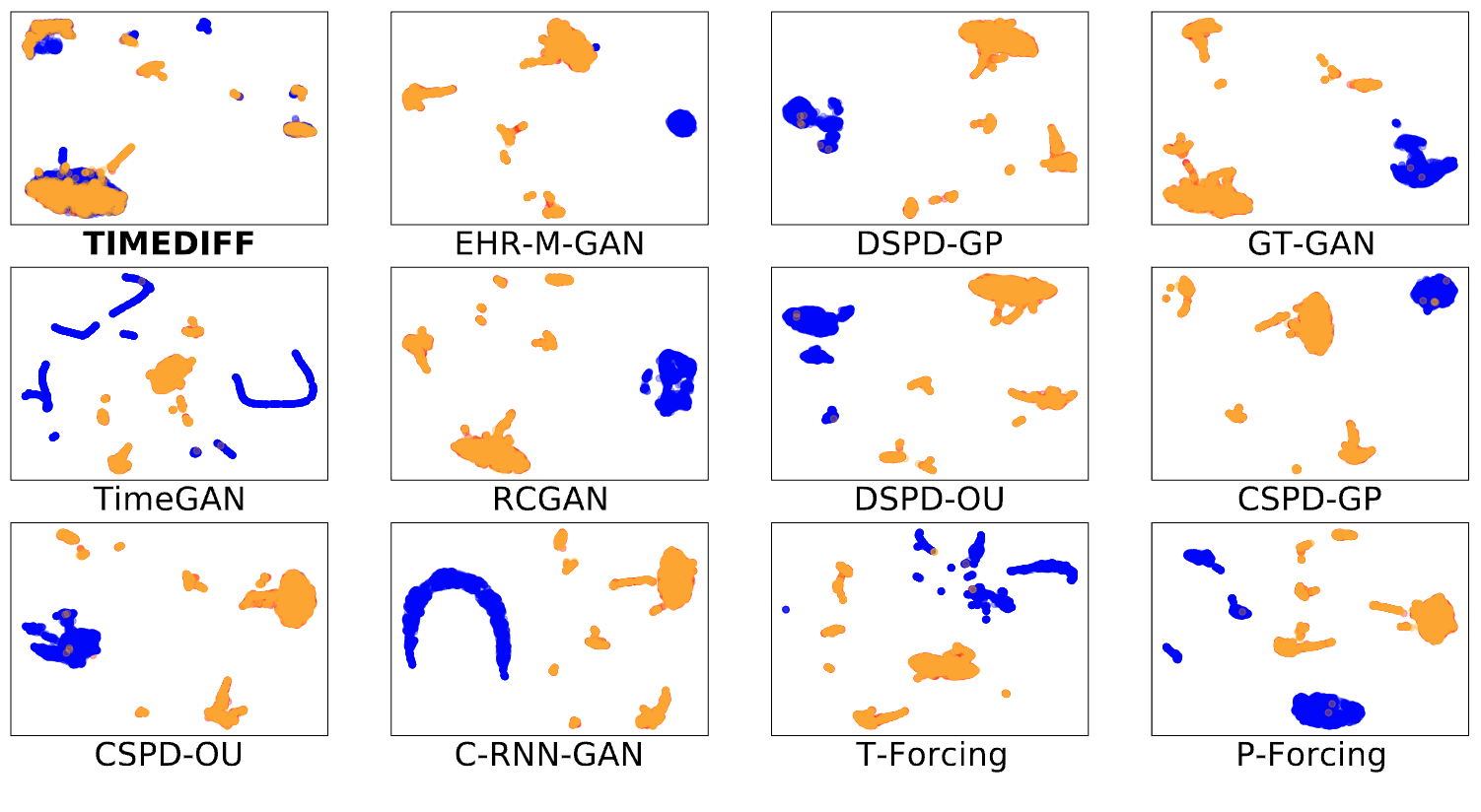}
  \caption{\firstedit{UMAP dimension reduction visualization for eICU.}}
\end{figure}

\begin{figure}[hbt!]
  \centering
  \includegraphics[width=\textwidth]{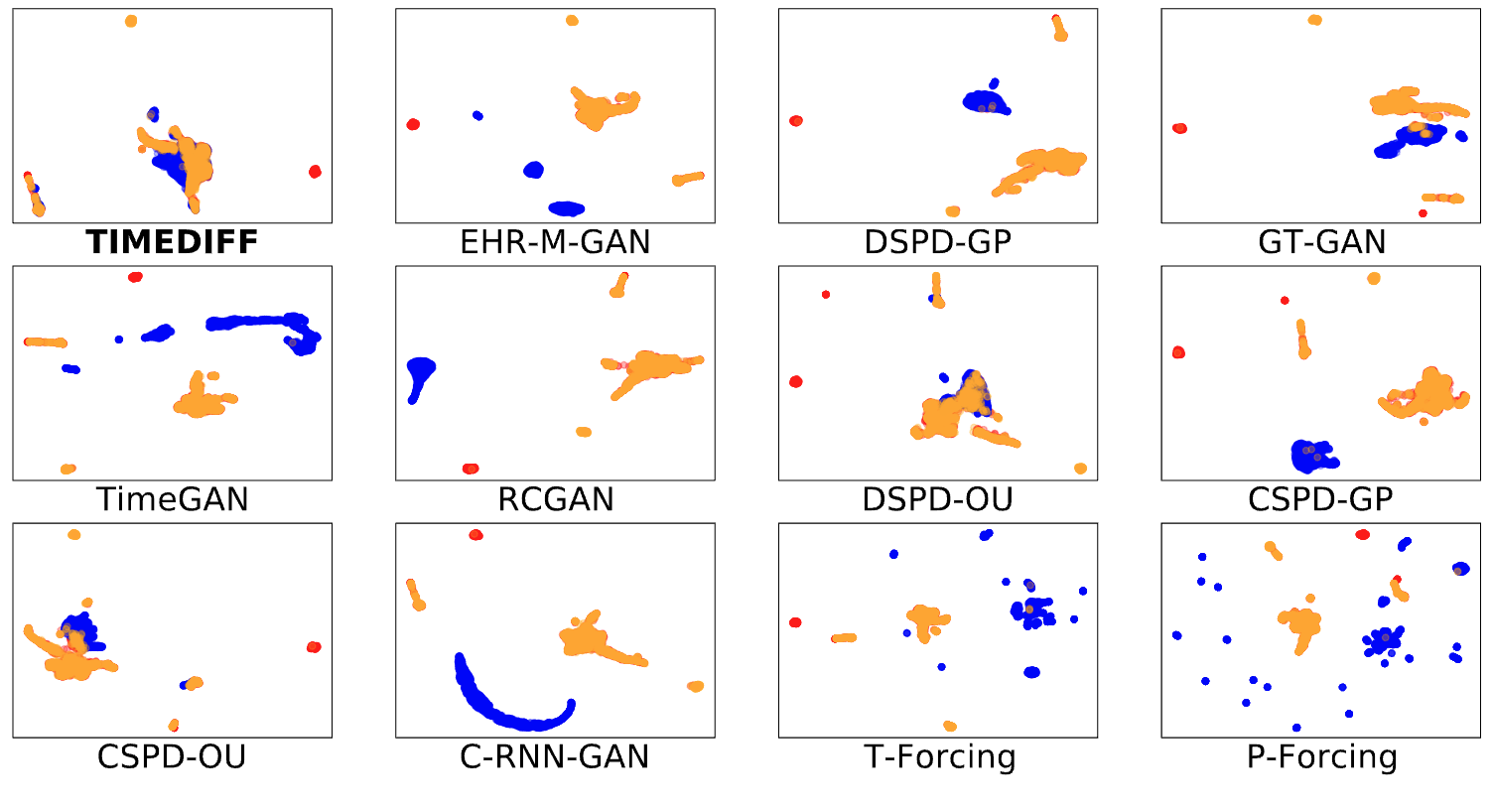}
  \caption{\firstedit{UMAP dimension reduction visualization for MIMIC-III.}}
\end{figure}

\begin{figure}[hbt!]
  \centering
  \includegraphics[width=\textwidth]{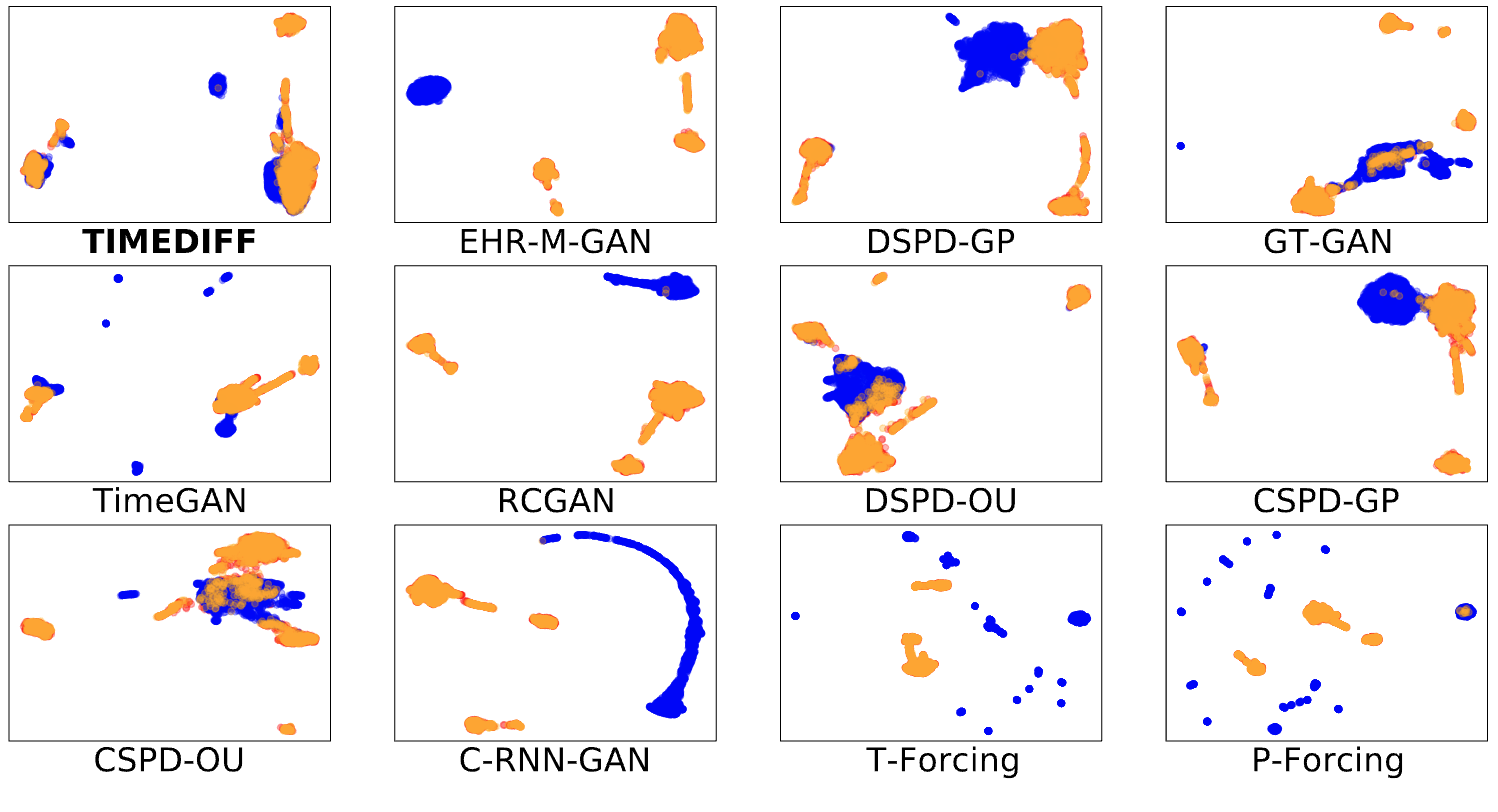}
  \caption{\firstedit{UMAP dimension reduction visualization for MIMIC-IV.}}
\end{figure}

\begin{figure}[htb!]
    \includegraphics[width=\textwidth]{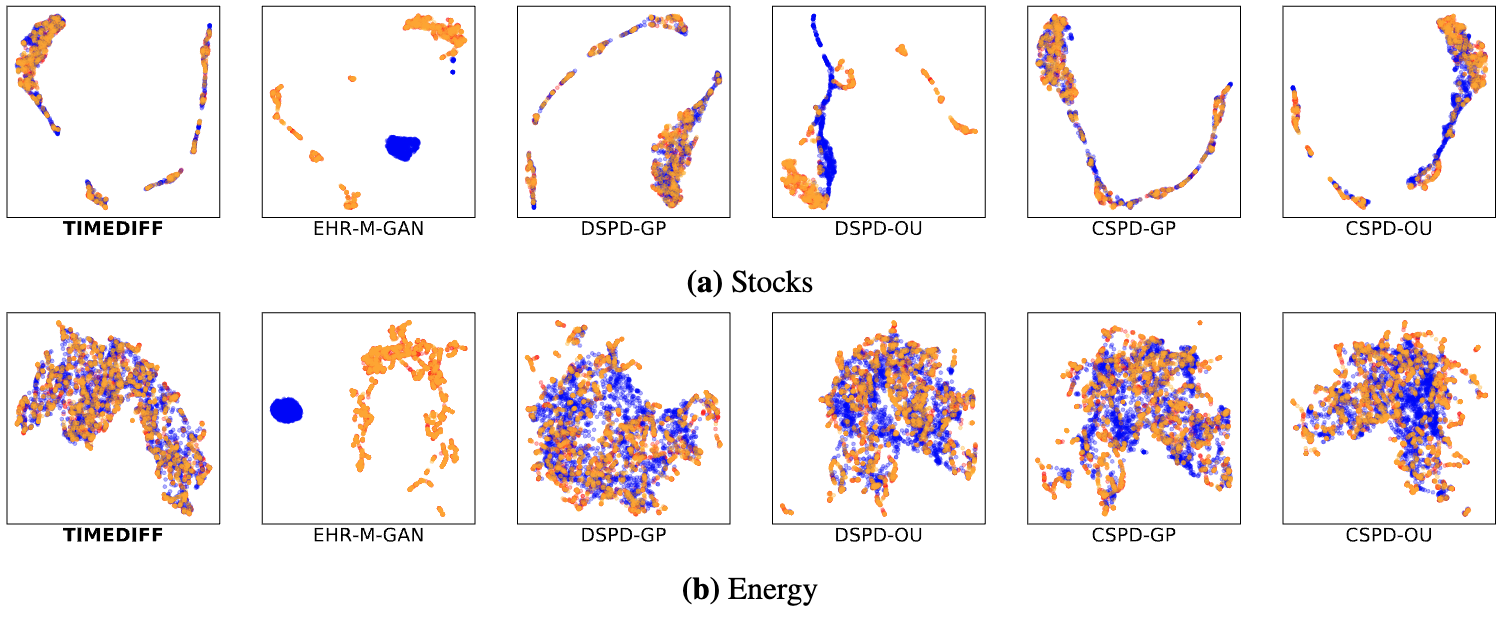}
    \caption{\firstedit{UMAP dimension reduction visualization for non-EHR datasets.}}
\end{figure}

\FloatBarrier
%\clearpage
\subsection{TSTR/TSRTR and Privacy Risk Evaluations \label{appendix:add:tstr}}

\subsubsection{\firstedit{\ours}}
This section provides additional results for TSTR and TSRTR scores across all four EHR datasets we considered in this study.
We train ML models using one of two methods: flattening the feature dimension of raw time series data, or using summary statistics such as initial measurement, minimum, maximum, range, mean, standard deviation, mode, and skewness.
\begin{figure}[hbt!]
    \centering
    \includegraphics[width=\textwidth]{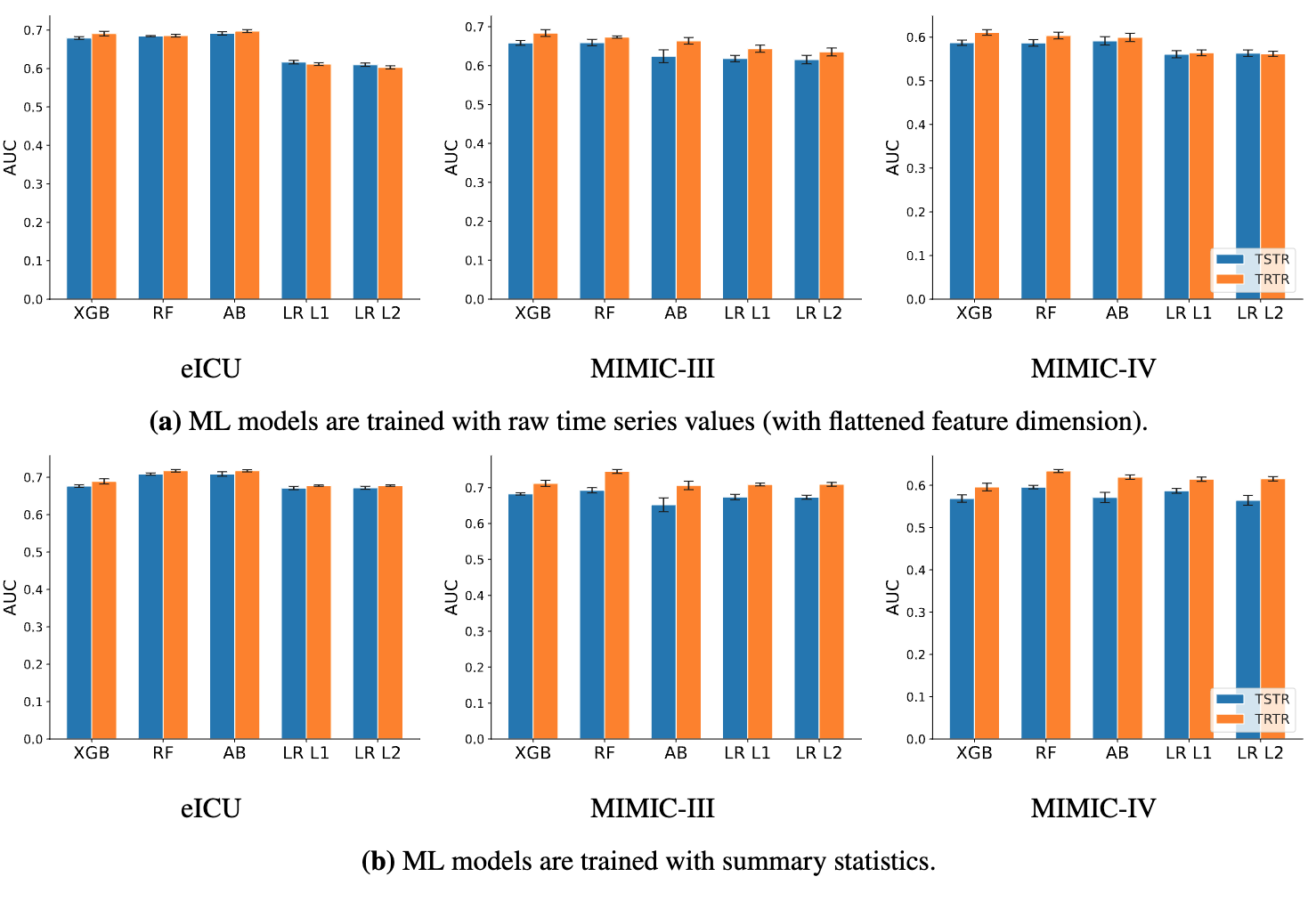}
    \caption{TSTR results.}
\end{figure}
%\vspace{-20pt}
\begin{figure}[hbt!]
    \centering
    \includegraphics[width=\textwidth]{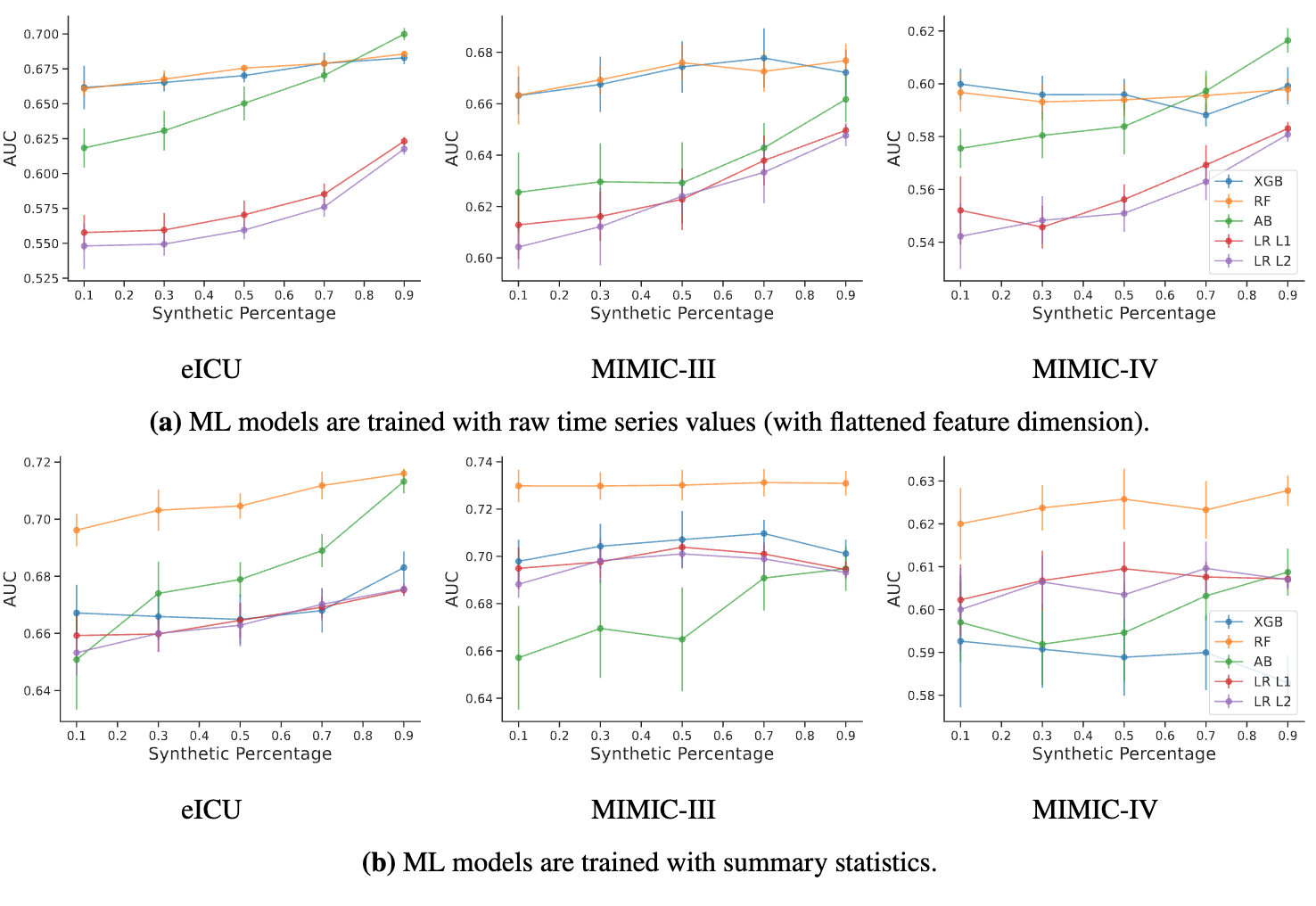}
    \caption{TSRTR results.}
\end{figure}
\begin{table}[hbt!]
    \caption{Full results for privacy risk evaluations.}
    % \vspace{-10pt}
    \centering
    \AppendixPrivacyTableone
    \label{appendix:add:privacy:nnaa_table}
\end{table}
\begin{table}[hbt!]
    \caption{Full results for privacy risk evaluations (continued).}
    % \vspace{-10pt}
    \centering
    \AppendixPrivacyTabletwo
    \label{appendix:add:privacy:nnaa_table_2}
\end{table}

\newpage
\FloatBarrier
\subsubsection{\firstedit{Baselines} \label{tstr_tsrtr:baselines}}
\firstedit{Next, we present the TSTR results for baseline generative models.
We observe that the patient labels produced by some generative models contain values outside of $\{0,1\}$ (an example is $\{0,1,2,3\}$), and some only produce one label across all patients.
In these cases, we cannot calculate TSTR and TSRTR scores, and thus, we leave them blank.
Specifically, we found that DSPD-GP, DSPD-OU, CSPD-GP, and CSPD-OU have these issues, so we do not include results for them in this section.
}
\begin{figure}[h]
    \centering
    \includegraphics[width=\textwidth]{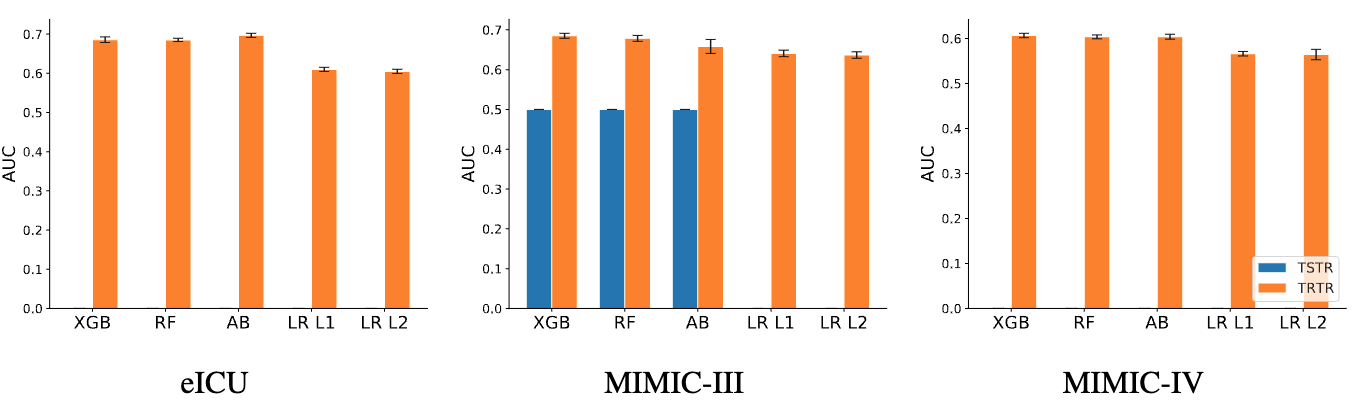}
    \caption{\firstedit{EHR-M-GAN}}
\end{figure}
% \begin{figure}[h]
%     \centering
%     \includegraphics[width=\textwidth]{pics/appendix_fig19.png}
%     \caption{\firstedit{DSPD-GP}}
% \end{figure}
% \begin{figure}[h]
%     \centering
%     \includegraphics[width=\textwidth]{pics/appendix_fig20.png}
%     \caption{\firstedit{DSPD-OU}}
% \end{figure}
% \begin{figure}[h]
%     \centering
%     \includegraphics[width=\textwidth]{pics/appendix_fig21.png}
%     \caption{\firstedit{CSPD-GP}}
% \end{figure}
% \begin{figure}[h]
%     \centering
%     \includegraphics[width=\textwidth]{pics/appendix_fig22.png}
%     \caption{\firstedit{CSPD-OU}}
% \end{figure}
\begin{figure}[h]
    \centering
    \includegraphics[width=\textwidth]{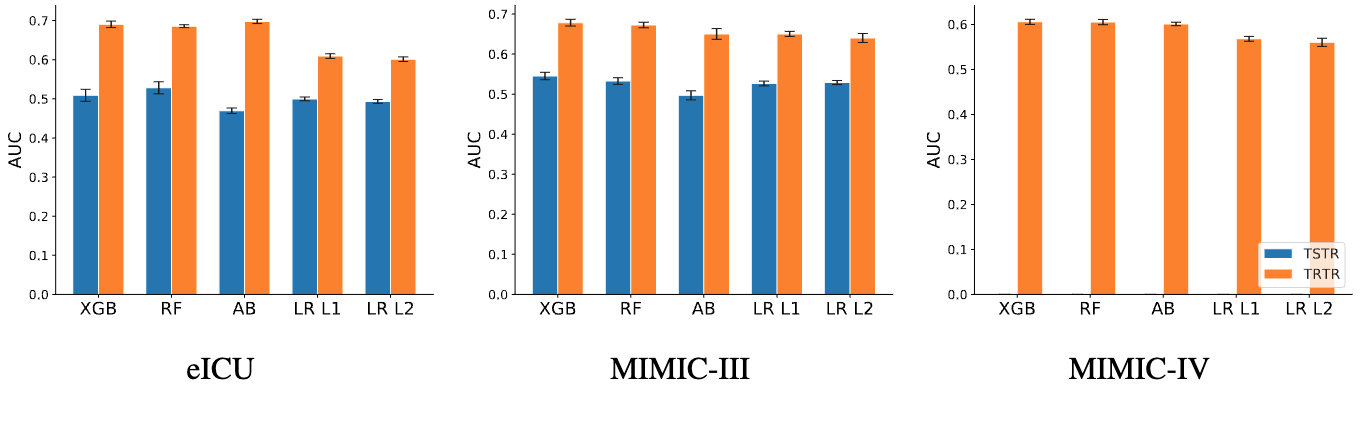}
    \caption{\firstedit{GT-GAN}}
\end{figure}
\begin{figure}[h]
    \centering
    \includegraphics[width=\textwidth]{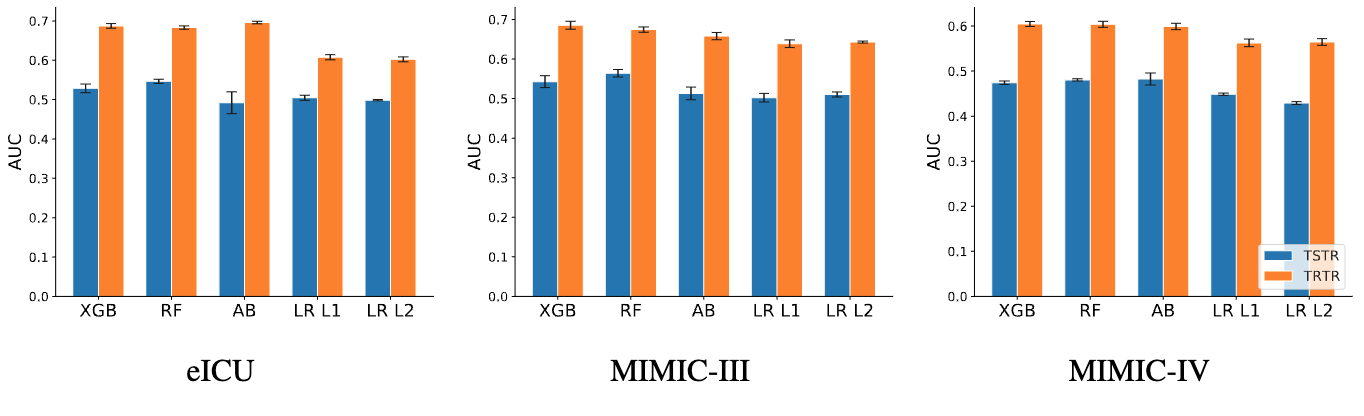}
    \caption{\firstedit{TimeGAN}}
\end{figure}
\begin{figure}[h]
    \centering
    \includegraphics[width=\textwidth]{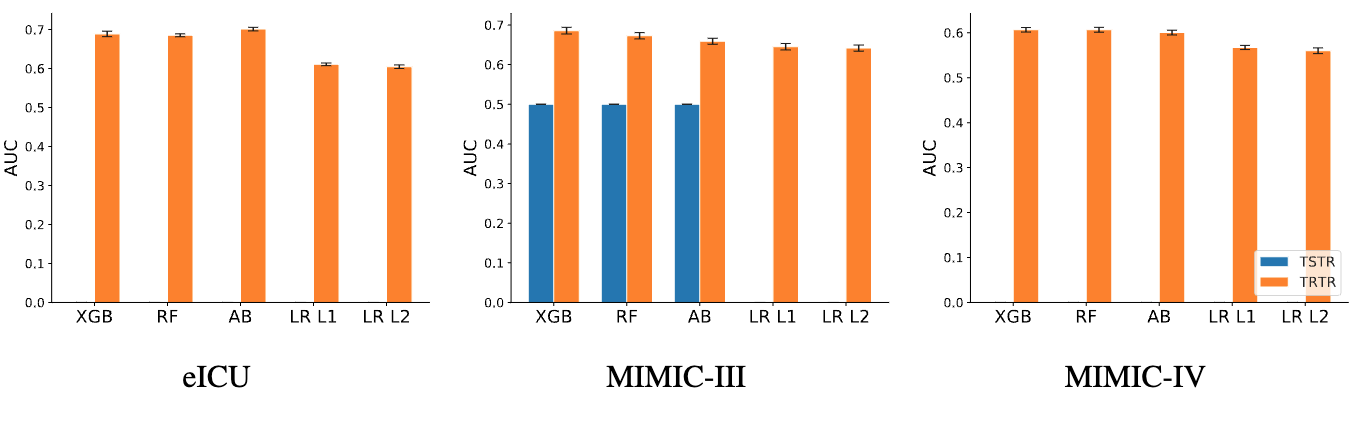}
    \caption{\firstedit{RCGAN}}
\end{figure}
\begin{figure}[h]
    \centering
    \includegraphics[width=\textwidth]{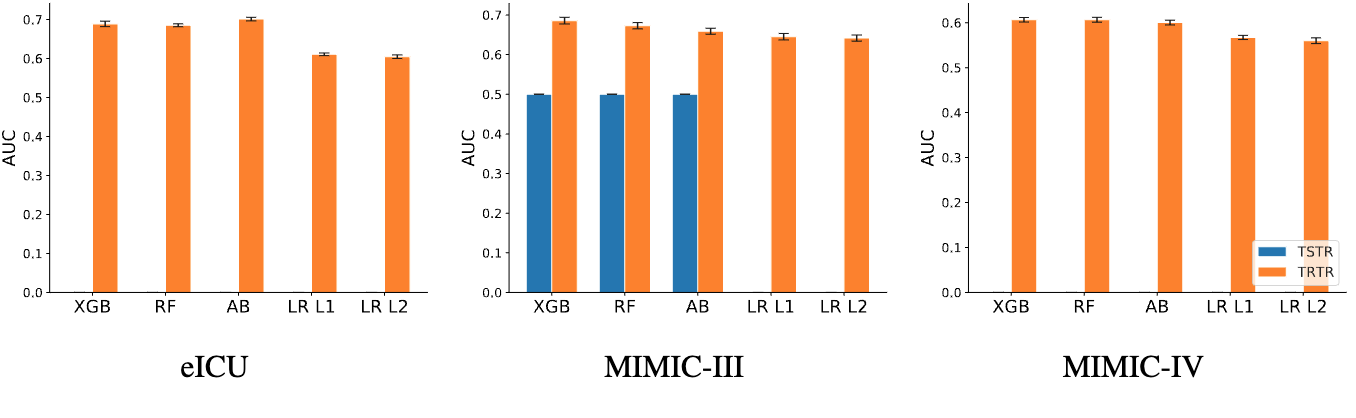}
    \caption{\firstedit{C-RNN-GAN}}
\end{figure}
\begin{figure}[h]
    \centering
    \includegraphics[width=\textwidth]{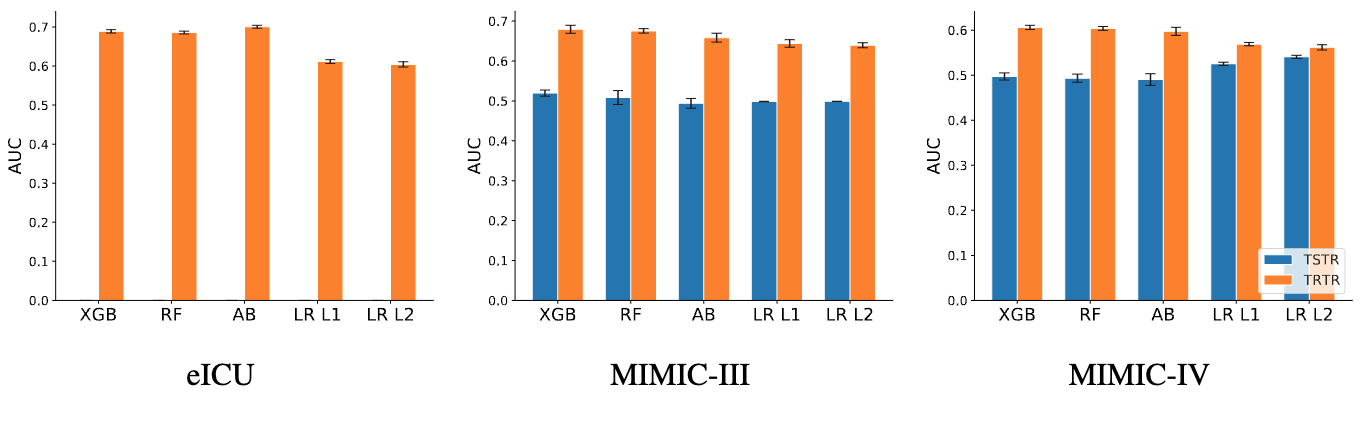}
    \caption{\firstedit{T-Forcing}}
\end{figure}
\begin{figure}[h]
    \centering
    \includegraphics[width=\textwidth]{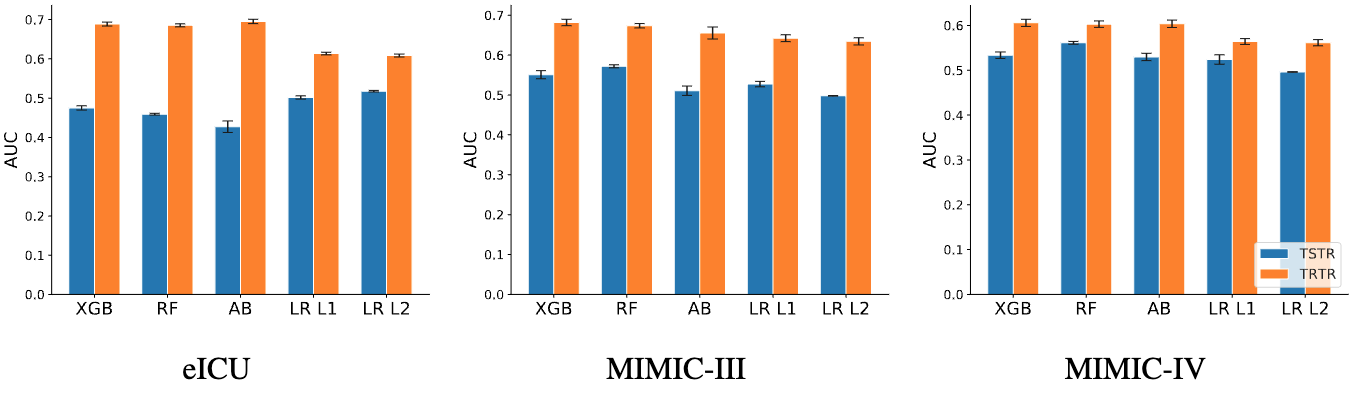}
    \caption{\firstedit{P-Forcing}}
\end{figure}
\begin{figure}[h]
    \centering
    \includegraphics[width=\textwidth]{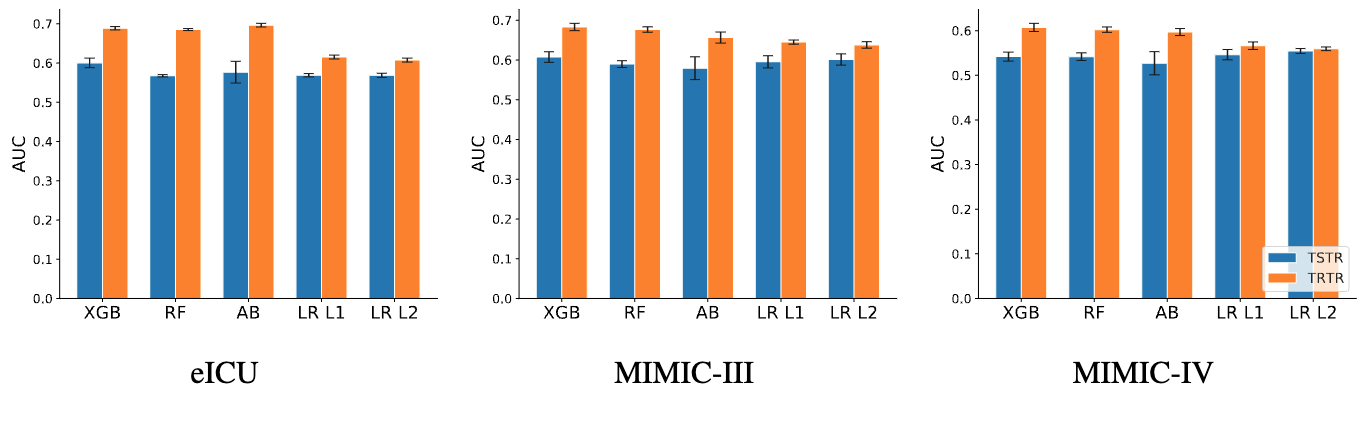}
    \caption{\firstedit{HALO} \label{tstr halo}}
\end{figure}
\begin{figure}[h]
    \centering
    \includegraphics[width=\textwidth]{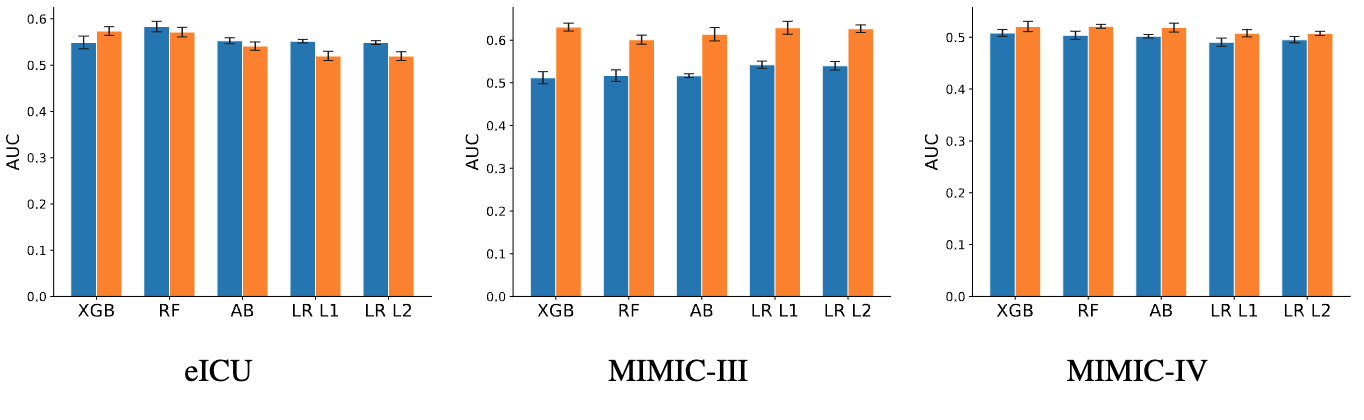}
    \caption{\firstedit{Gaussian Diffusion and Softmax} \label{gauss softmax tstr}}
\end{figure}

\begin{figure}[h]
    \centering
    \includegraphics[width=\textwidth]{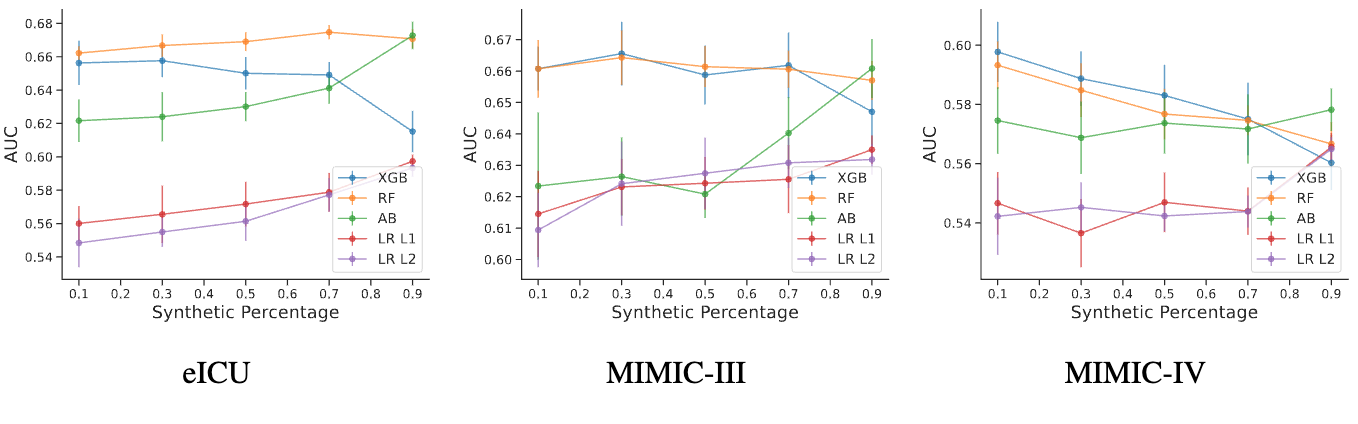}
    \caption{\firstedit{HALO}}
\end{figure}
\begin{figure}[h]
    \centering
    \includegraphics[width=\textwidth]{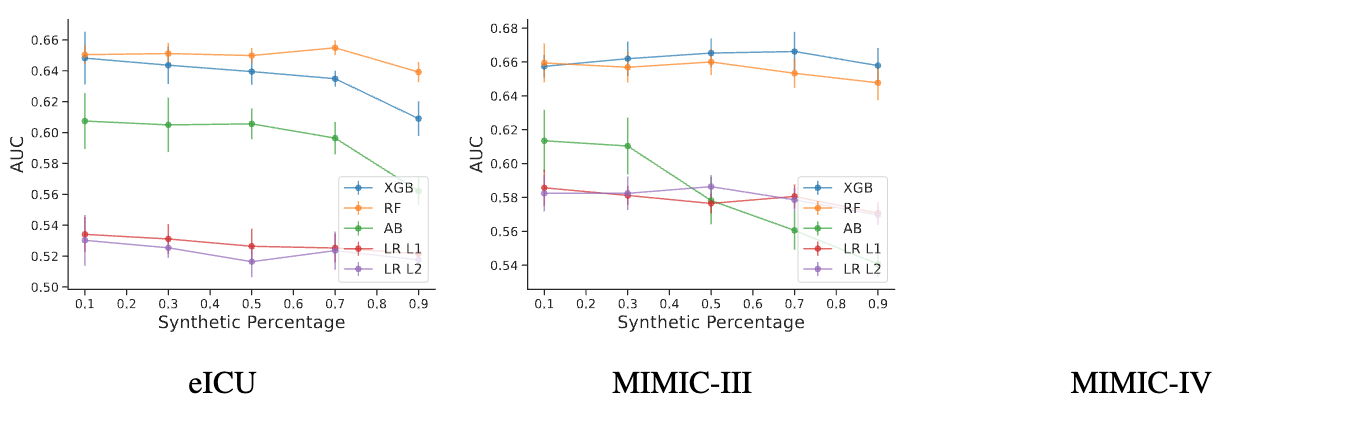}
    \caption{\firstedit{GT-GAN}}
\end{figure}
\begin{figure}[h]
    \centering
    \includegraphics[width=\textwidth]{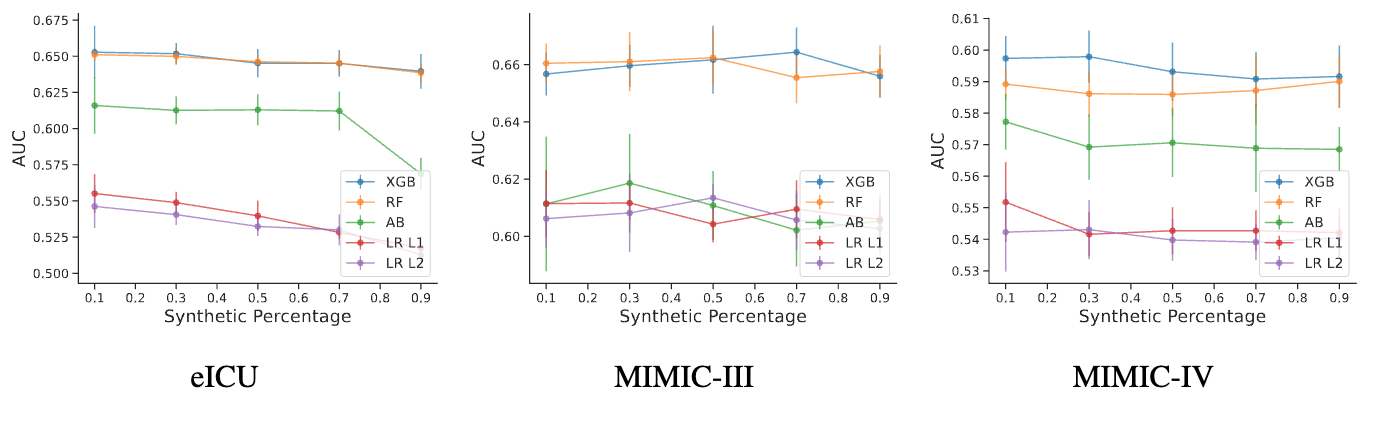}
    \caption{\firstedit{P-Forcing}}
\end{figure}
\begin{figure}[h]
    \centering
    \includegraphics[width=\textwidth]{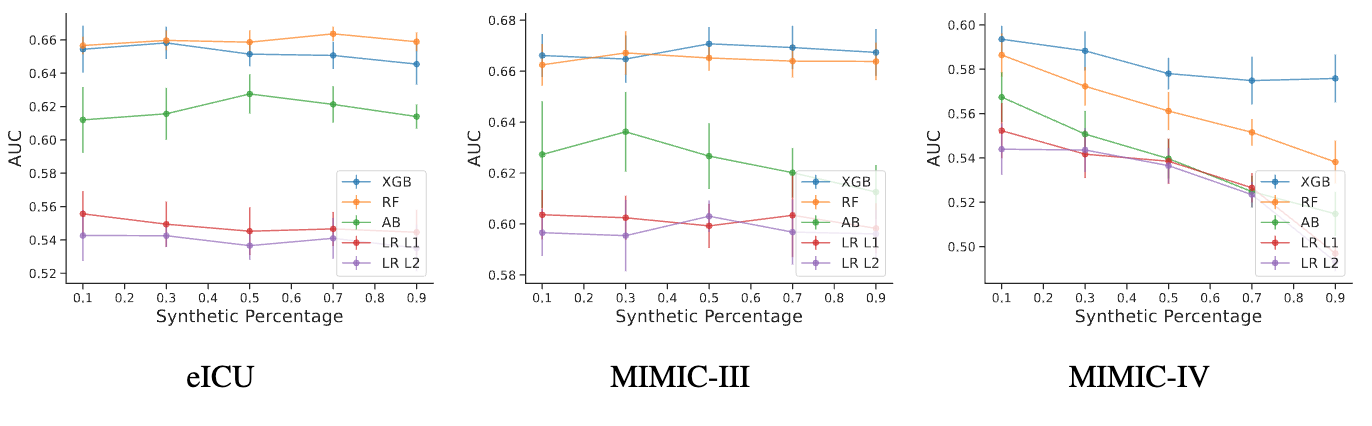}
    \caption{\firstedit{TimeGAN}}
\end{figure}
\begin{figure}[h]
    \centering
    \includegraphics[width=\textwidth]{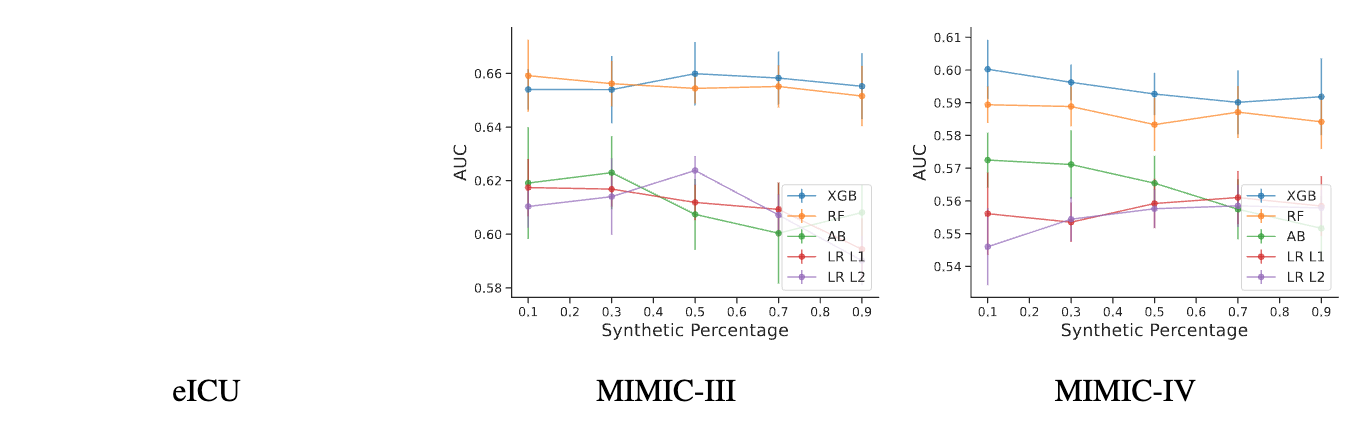}
    \caption{\firstedit{T-Forcing}}
\end{figure}
\begin{figure}[h]
    \centering
    \includegraphics[width=\textwidth]{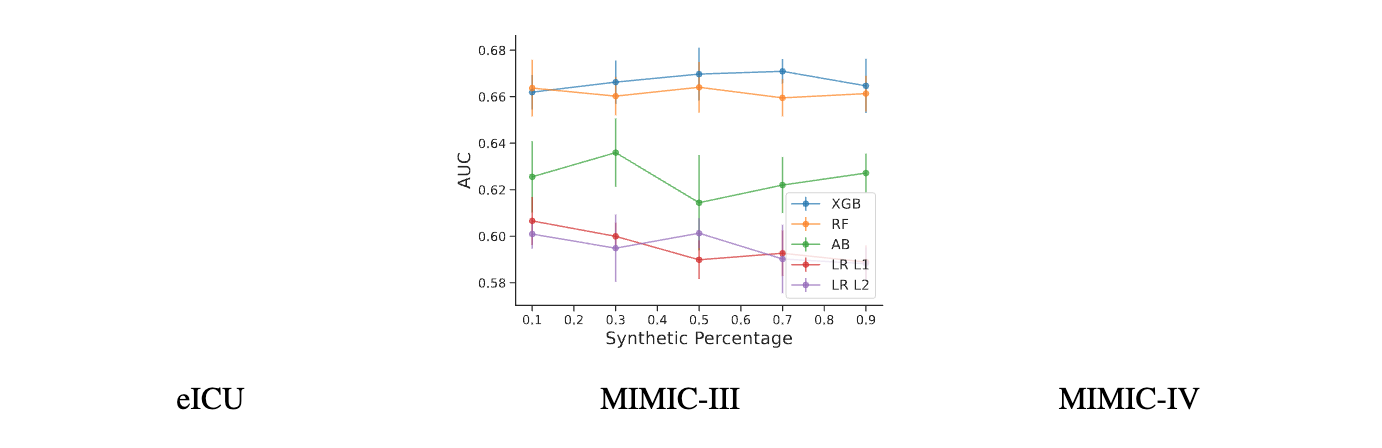}
    \caption{\firstedit{C-RNN-GAN}}
\end{figure}
\begin{figure}[h]
    \centering
    \includegraphics[width=\textwidth]{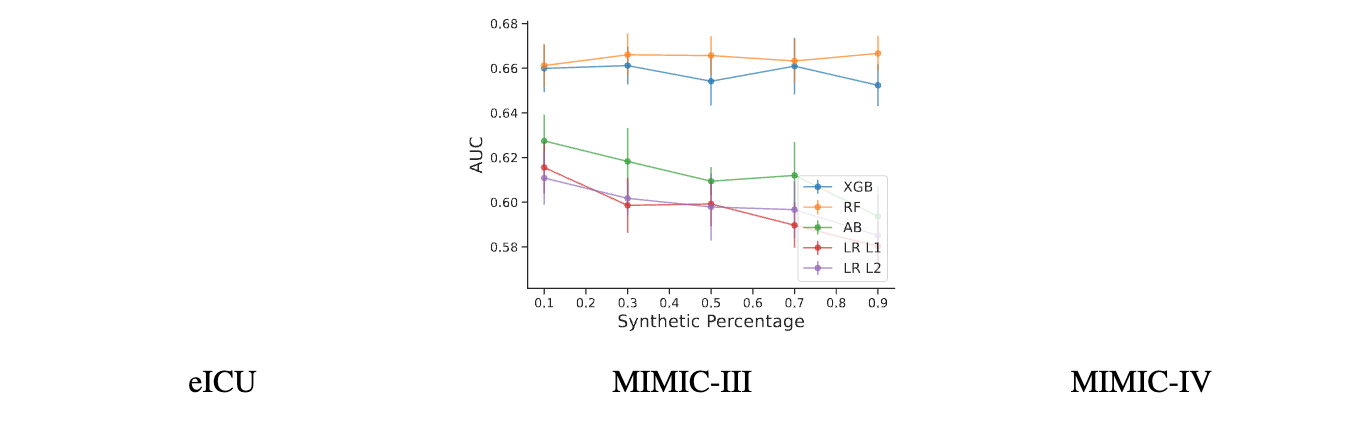}
    \caption{\firstedit{EHR-M-GAN}}
\end{figure}
\begin{figure}[h]
    \centering
    \includegraphics[width=\textwidth]{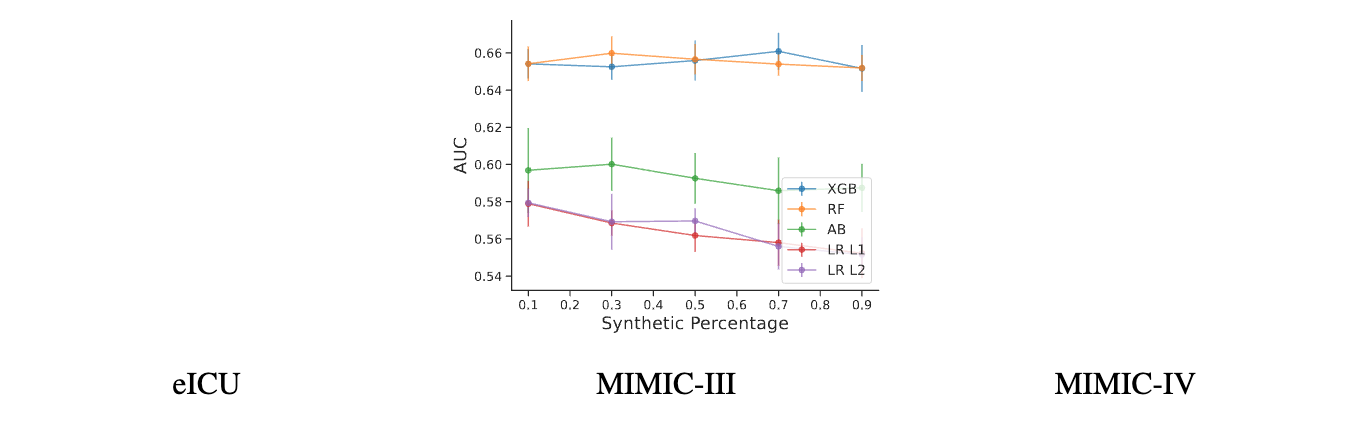}
    \caption{\firstedit{RCGAN}}
\end{figure}
\begin{figure}[h]
    \centering
    \includegraphics[width=\textwidth]{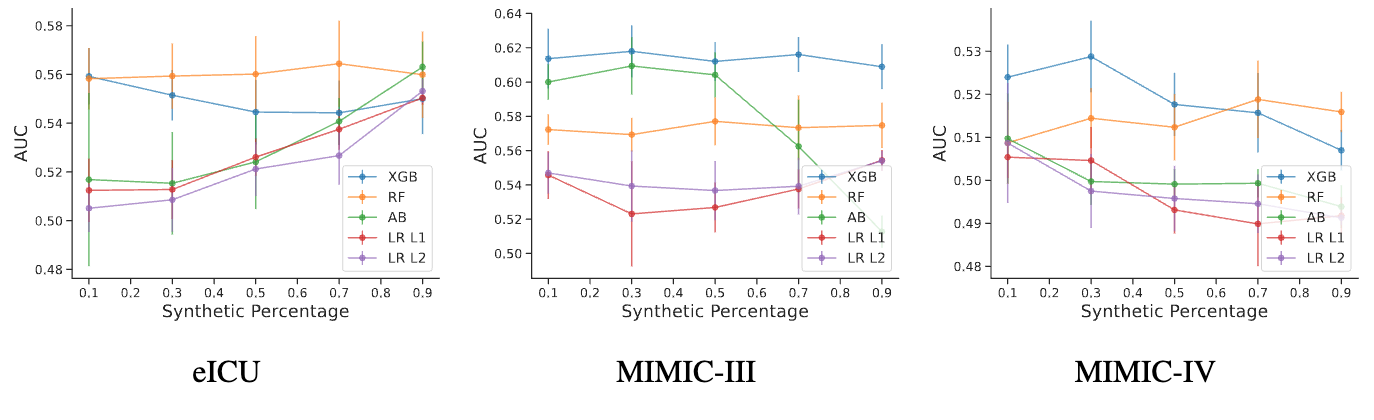}
    \caption{\firstedit{Gaussian Diffusion and Softmax} \label{gauss softmax tsrtr}}
\end{figure}

\FloatBarrier
\newpage
\subsubsection{\firstedit{Recurrent Neural Network Classifiers}\label{appendix:add:tstr:rnn}}
\firstedit{As an additional evaluation, we train RNN classifiers on synthetic time series data generated from \ours and evaluate their performance on real testing data.
We use bidirectional RNNs with a hidden dimension of 64 for this experiment.}
\begin{table}[hbt!]
    \caption{\firstedit{TSTR and TRTR scores for RNN classifiers.}}
    %\vspace{-10pt}
    \centering
    \AppendixTSTRRNN
    \label{appendix:add:tstr:rnn_table}
\end{table}

\subsubsection{\firstedit{TSRTR with Full Dataset}}
\firstedit{We also further calculated the TSRTR of \ours by starting using the entire real training dataset and adding different percentages of synthetic samples to it.}

\begin{figure}[h]
    \centering
    \includegraphics[width=\textwidth]{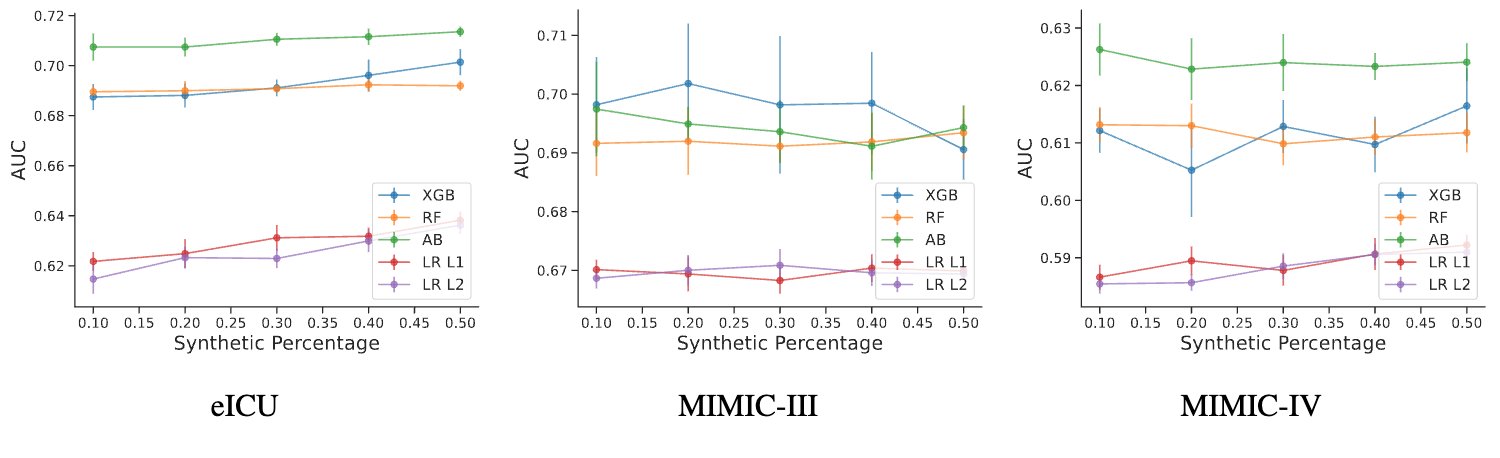}
    \caption{\firstedit{\ours}}
\end{figure}
% \FloatBarrier
% \clearpage
% \subsection{Privacy Scores}
% In this section, we present our additional results for privacy scores.
% Calculations of the $\aatrain$, $\aatest$, NNAA, and MIR scores are discussed in \Cref{appendix:privacy}. We find that the $\aatrain$ and $\aatest$ scores of the synthetic samples generated by \ours are both close to 0.5 for most datasets.
% This indicates that \ours does not overfit the training data and the synthetic samples represent the testing data well.
% %This indicates that the synthetic samples are not overfitting to the training data and represent the testing data well.
% As a baseline, we also use the real data to calculate privacy scores.

% \textbf{Baselines with Real Data:}
% For $\aatest$, it is the accuracy of the adversarial classifier for distinguishing between real testing and real training data.
% This also yields values around 0.5 since the training and testing sets share a similar distribution.
% $\aatrain$ has an even lower score because it represents the accuracy for distinguishing between different sub-samples of the real training data.
% Lastly, the NNAA score is the difference between $\aatest$ and $\aatrain$.
% \begin{table}[h]
%     % \vspace{-5pt}
%     \centering
%     \AppendixPrivacyTable
%     \caption{Privacy score evaluations.}
%     % \vspace{-5pt}
% \end{table}

\FloatBarrier
\subsection{Runtime Comparisons}
In this section, we present additional runtime comparisons across all EHR datasets.
We consider EHR-M-GAN, stochastic process diffusion models, TimeGAN, and GT-GAN.
We use Intel Xeon Gold 6226 Processor and Nvidia GeForce 2080 RTX Ti to train all the models for a fair comparison.

\begin{table}[hbt!]
\caption{Comparison of runtime (hours).}
%\vspace{-10pt}
\centering
\AppendixTimeModelSize
\label{appendix:add:time model size}
%\vspace{-15pt}
\end{table}

\FloatBarrier
\subsection{Effect of \texorpdfstring{$\lambda$}{lambda} \label{appendix:add:lambda}}
In this section, we investigate the effect of hyperparameter $\lambda$ on \ours.
We trained \ours using $\lambda \in \{0.001, 0.01, 0.1, 1, 10\}$ while keeping the other hyperparameters identical as those described in \Cref{appendix:detail:hyperparams:training}.

\begin{table}[hbt!]
\caption{Effect of $\lambda$ on data utility.}
%\vspace{-10pt}
\centering
\AppendixAblationLambda
\label{appendix:add:tab:lambda}
%\vspace{-20pt}
\end{table}

\begin{figure}[hbt!]
    \centering
    \includegraphics[width=\textwidth]{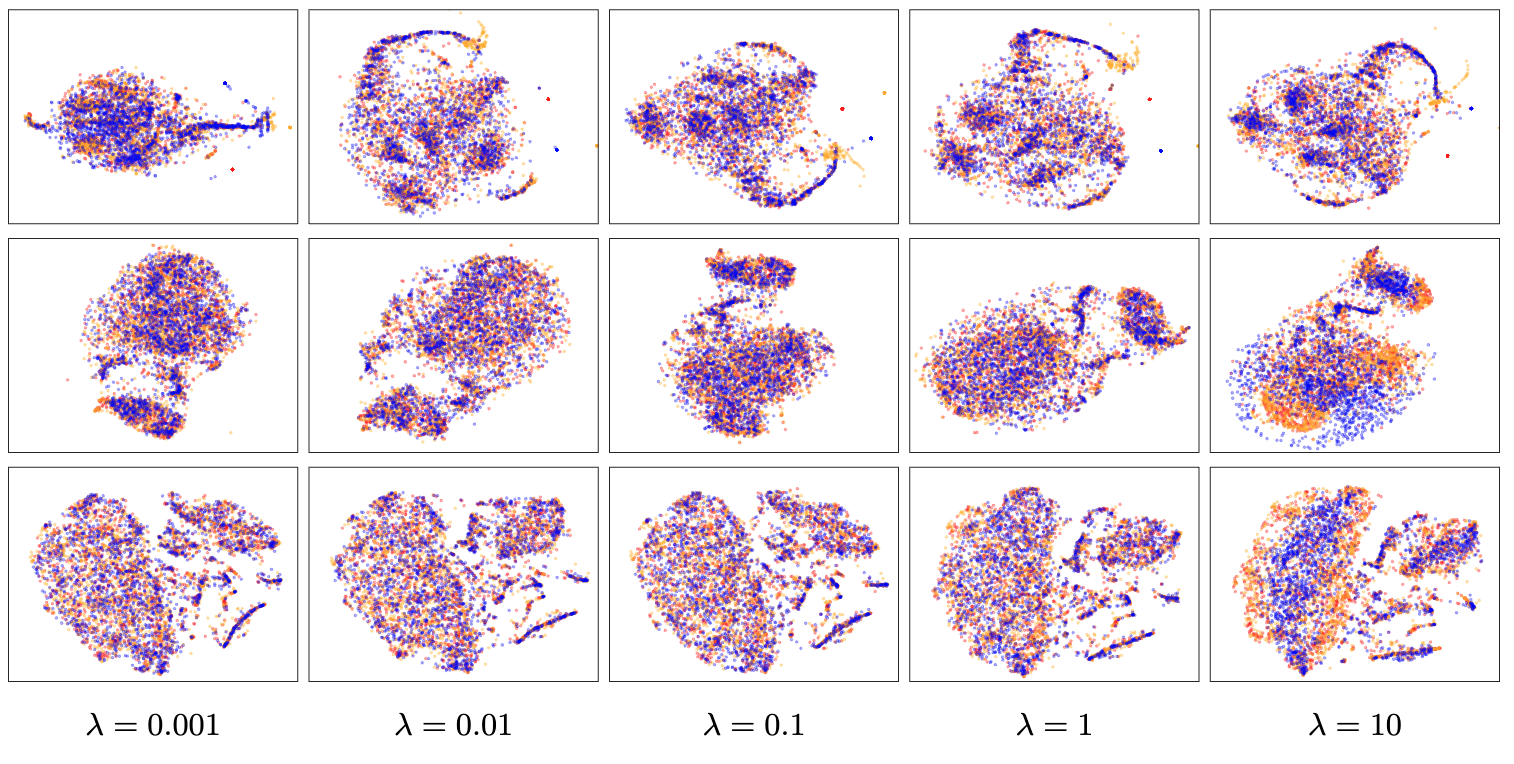}
    \caption{t-SNE visualizations. The first row is from MIMIC-III, the second row is from MIMIC-IV, and the third row is from the eICU dataset.}
\label{appendix:add:fig:tsne}
\end{figure}

\FloatBarrier
\subsection{\firstedit{Additional Experiments on Infrequent Time Series, ICD Codes, Antibiotics, and Age} \label{appendix:infrequent static}}
\firstedit{In this section, we include the results of generating infrequent time series as well as other dynamic and static data. Specifically, we added channels for glucose, ICD codes, administered antibiotics, and the age at patients' initial visits. Experiments were conducted using the MIMIC-IV dataset. Specifics on data extraction and formatting are detailed below.}

\subsubsection{\firstedit{Glucose Channel Format}}
\firstedit{For infrequent time series generation, we used glucose measurements of each patient across a 72-hour interval. Since glucose measurements are extremely sparse and never occur at fixed time intervals, sampling every hour from original time stamps without rounding (our approach with vital signs) would result in nearly no other measurements apart from the initial one. Therefore, we rounded each measurement timestamp to the nearest 20 minutes. We then resampled the data at every hour in the same way as vital signs.
%We acknowledge that rounding weakens both our training and generated data in terms of accurately representing real-world glucose measurements. However, this was the best approach we attempted within a limited schedule. We also did not use an additional discrete masking channel to generate our missing measurements for our final results with glucose time series.
We represented missing measurements with -1 within the glucose channel. After generating samples, we replaced all generated values outside of the range of possible measurement values (predetermined from the training and testing dataset) with -1. We empirically found that preprocessing the data in such a way to add skewness to the distribution works better with our architecture for infrequent or sparse time series. We suspect that the added skewness makes the occurrence of a measurement more prominent for the BRNN. Further investigations into reasons for this behavior would be an interesting direction for future work.}

\subsubsection{\firstedit{ICD Code Channel Format}}
\firstedit{The medGAN repository: \href{https://github.com/mp2893/medgan}{https://github.com/mp2893/medgan} was referenced for preprocessing and extracting patient ICD codes from the MIMIC-IV database. The codes were selected based on their frequency in the entire MIMIC-IV database. The top 72 most frequent ICD codes were selected to maintain a length consistent with the 72-hour time series. Each patient's ICD code channel is represented by a binary vector where 1 denotes the presence of a specific ICD code and 0 denotes the absence of the ICD code.}

\firstedit{\textbf{The 72 most frequent ICD codes are detailed below in order from the most frequent to the least frequent:}
ICD9: 4019
 , ICD9: 2724
 , ICD10: I10
 ,  ICD10: E785
 ,  ICD9: 53081
 ,  ICD9: 25000
 ,  ICD10: Z87891
 ,  ICD9: 42731
 ,  ICD9: 311
 ,  ICD9: 4280
 ,  ICD9: 41401
 ,  ICD10: K219
 ,  ICD9: V1582
 ,  ICD10: F329
 ,  ICD9: 5849
 ,  ICD9: 2449
 ,  ICD10: I2510
 ,  ICD9: 3051
 ,  ICD9: 2859
 ,  ICD9: 40390
 ,  ICD10: F419
 ,  ICD9: V5861
 ,  ICD9: 30000
 ,  ICD10: N179
 ,  ICD9: 5990
 ,  ICD9: 2720
 ,  ICD9: 49390
 ,  ICD9: V5867
 ,  ICD10: Z794
 ,  ICD10: E039
 ,  ICD9: 5859
 ,  ICD10: Z7901
 ,  ICD10: E119
 ,  ICD9: 32723
 ,  ICD10: F17210
 ,  ICD10: Y929
 ,  ICD9: V4582
 ,  ICD9: 412
 ,  ICD9: V5866
 ,  ICD10: G4733
 ,  ICD9: 496
 ,  ICD9: 27800
 ,  ICD10: E669
 ,  ICD10: I4891
 ,  ICD10: D649
 ,  ICD10: J45909
 ,  ICD10: Z7902
 ,  ICD9: V4581
 ,  ICD9: 2761
 ,  ICD9: 41400
 ,  ICD9: 73300
 ,  ICD9: 30500
 ,  ICD10: J449
 ,  ICD9: 33829
 ,  ICD9: V1251
 ,  ICD10: Z66
 ,  ICD9: 486
 ,  ICD10: N390
 ,  ICD9: 2749
 ,  ICD9: 27651
 ,  ICD10: D62
 ,  ICD10: I129
 ,  ICD9: V1254
 ,  ICD9: V4986
 ,  ICD9: V270
 ,  ICD10: E1122
 ,  ICD9: 78650
 ,  ICD10: I252
 ,  ICD9: 2851
 ,  ICD10: N189
 ,  ICD9: 60000
 ,  ICD9: 56400.}
 
\subsubsection{\firstedit{Antibiotic Channel Format}}
\firstedit{Similar to formatting the ICD codes, the top 72 most frequent antibiotic codes were selected to be represented in the channel. Each patient's antibiotic channel is represented by a binary vector where 1 denotes the presence of an administered antibiotic code and 0 denotes the absence of the code.}

\firstedit{\textbf{The 72 most frequent antibiotics are detailed below in order from the most frequent to the least frequent:}
       'Vancomycin', 'CefazoLIN', 'Ciprofloxacin HCl', 'CefePIME',
       'MetRONIDAZOLE (FLagyl)', 'CeftriaXONE', 'Levofloxacin',
       'Piperacillin-Tazobactam', 'Ciprofloxacin IV', 'CefTRIAXone',
       'CeFAZolin', 'Azithromycin', 'MetroNIDAZOLE',
       'Sulfameth/Trimethoprim DS', 'Ampicillin-Sulbactam', 'Clindamycin',
       'Amoxicillin-Clavulanic Acid', 'Cephalexin', 'Meropenem',
       'Doxycycline Hyclate', 'Vancomycin Oral Liquid',
       'Sulfameth/Trimethoprim SS', 'Cefpodoxime Proxetil',
       'Ampicillin Sodium', 'CefTAZidime', 'Mupirocin Ointment 2\%',
       'Azithromycin', 'Linezolid', 'Gentamicin Sulfate', 'Gentamicin',
       'Nitrofurantoin Monohyd (MacroBID)', 'Piperacillin-Tazobactam Na',
       'Nafcillin', 'Amoxicillin', 'Mupirocin Nasal Ointment 2\%',
       'Erythromycin', 'Aztreonam', 'Tobramycin Sulfate',
       'Sulfameth/Trimethoprim Suspension', 'Penicillin G Potassium',
       'LevoFLOXacin', 'Clarithromycin', 'Ampicillin',
       'Sulfamethoxazole-Trimethoprim', 'Rifampin',
       'Nitrofurantoin (Macrodantin)', 'CeftazIDIME', 'DiCLOXacillin',
       'Vancomycin Enema', 'Minocycline', 'Ciprofloxacin',
       'Amoxicillin-Clavulanate Susp.', 'Neomycin Sulfate',
       'Penicillin V Potassium', 'Sulfameth/Trimethoprim',
       'Vancomycin Antibiotic Lock', 'Amikacin', 'Ceftaroline',
       'vancomycin', 'Erythromycin Ethylsuccinate Suspension',
       'Moxifloxacin', 'Tetracycline HCl', 'Tobramycin Inhalation Soln',
       'Tetracycline', 'moxifloxacin', 'AMOXicillin Oral Susp.',
       'Neomycin/Polymyxin B Sulfate', 'Penicillin G Benzathine',
       'Trimethoprim', 'CefTRIAXone Graded Challenge',
       'Cefepime Graded Challenge', 'Meropenem Graded Challenge'.}

\subsubsection{\firstedit{Age Channel Format}}
\firstedit{The MIMIC-IV database contains patients with ages ranging from 18 to 89 inclusive. All ages above 89 are replaced with 91.
%As a result, there are 73 possible age numbers to be represented.
The age channel for each patient is represented by a numerical time series of length 72 containing the same age value. After sampling, we obtained the age value by computing the rounded mean of the age time series.}

\begin{table}[hbt!]
    \caption{\firstedit{Numerical Evaluation Metrics on MIMIC-IV }}
    \centering
    \AppendixMIMICIVRevisionAllMetrics
    \label{appendix:add:tab:mimiciv_revision}
\end{table}

\begin{figure}[h]
    \includegraphics[width=\linewidth]{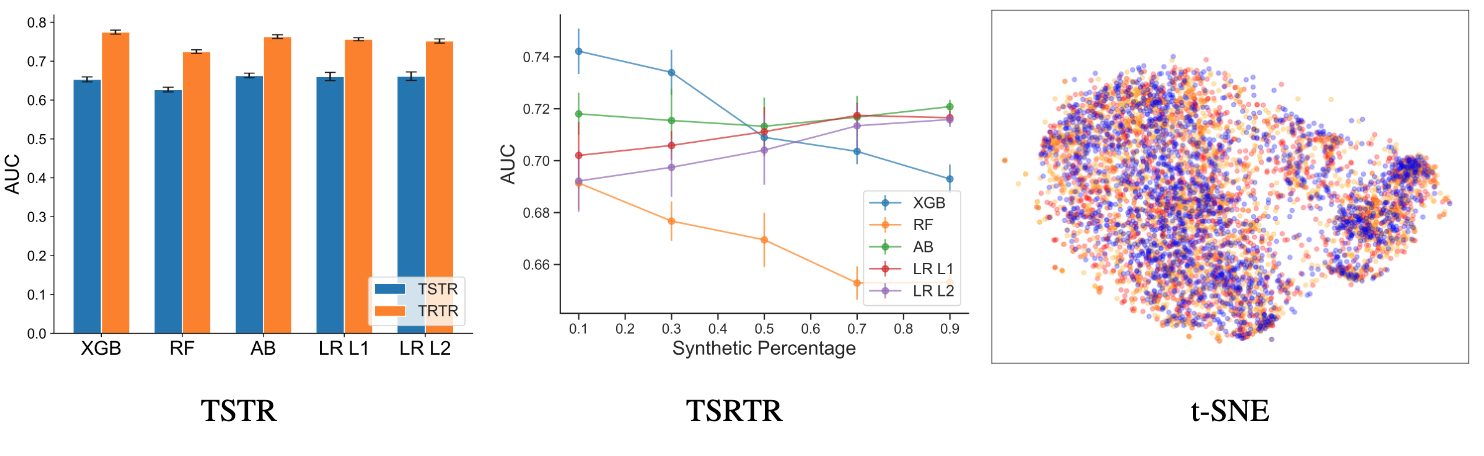}
    \caption{\firstedit{Evaluations of the generated MIMIC-IV time series, ICD codes, antibiotics, age, and mortality.}}
\end{figure}

% \begin{figure}[hbt!]
%     \centering
%     \includegraphics[height=0.33\textheight]{pics/mimiciv_remove_infrequent_masks_tstr_trtr.png}
%     \caption{\firstedit{TSTR and TRTR scores of MIMIC-IV time series, ICD codes, antibiotics, age, and mortality}}
%     \label{appendix:add:fig:mimiciv_revision_tstr}
% \end{figure}

% \begin{figure}[hbt!]
%     \centering
%     \includegraphics[height=0.33\textheight]{pics/mimiciv_remove_infrequent_masks_tsrtr.png}
%     \caption{\firstedit{TSRTR scores of MIMIC-IV time series, ICD codes, antibiotics, age, and mortality}}
%     \label{appendix:add:fig:mimiciv_revision_tsrtr}
% \end{figure}

% \begin{figure}[hbt!]
%     \centering
%     \includegraphics[height=0.33\textheight]{pics/mimiciv_remove_infrequent_masks_t-sne.png}
%     \caption{\firstedit{t-SNE visualization of MIMIC-IV time series, ICD codes, antibiotics, age, and mortality}}
%     \label{appendix:add:fig:mimiciv_revision_tsne}
% \end{figure}

\begin{figure}[hbt!]
    \centering
    \includegraphics[width=\textwidth]{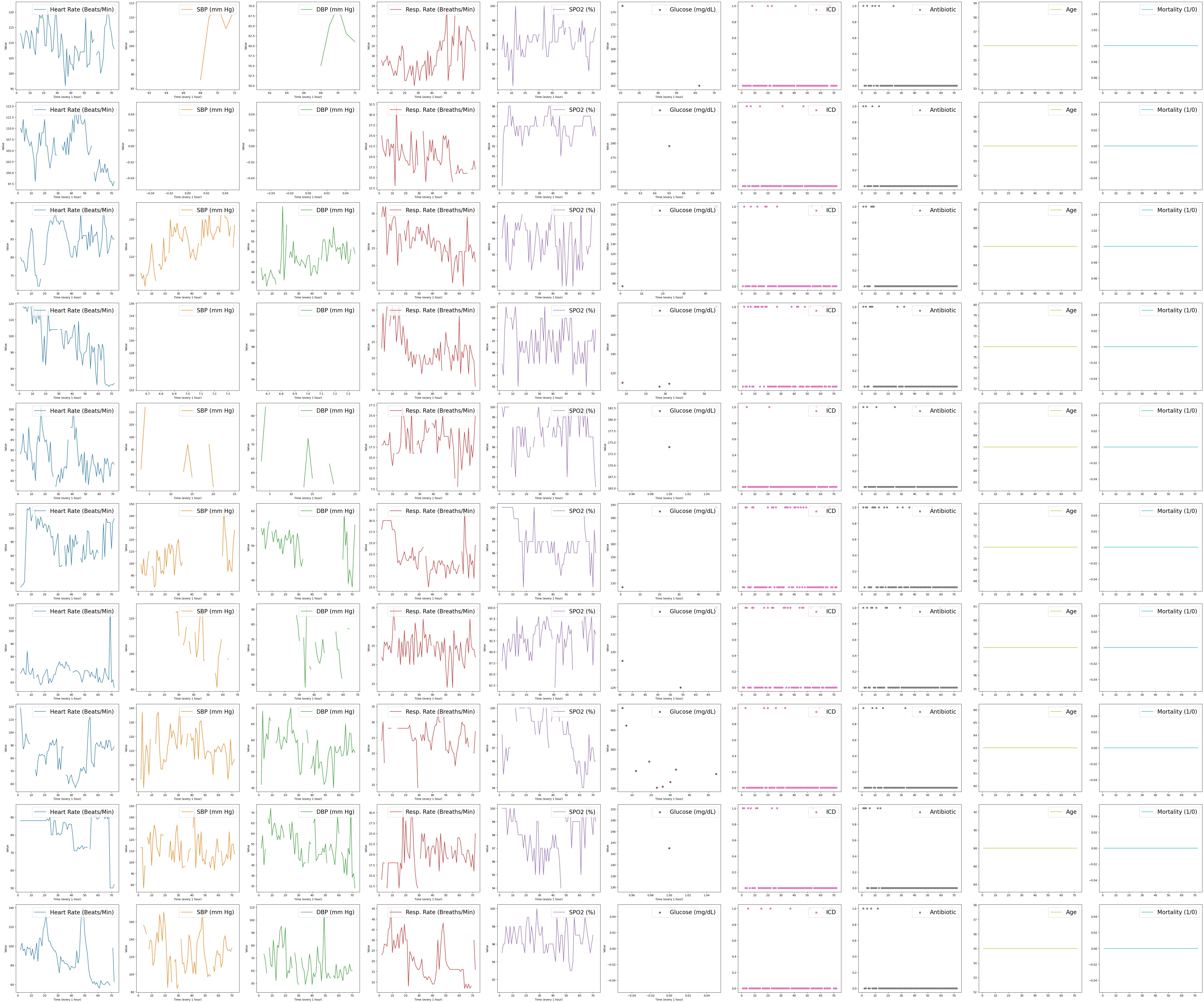}
    \caption{\firstedit{Visualization of real MIMIC-IV time series, ICD codes, antibiotics, age, and mortality data}}
    \label{appendix:add:fig:mimiciv_revision_real}
\end{figure}

\begin{figure}[hbt!]
    \centering
    \includegraphics[width=\textwidth]{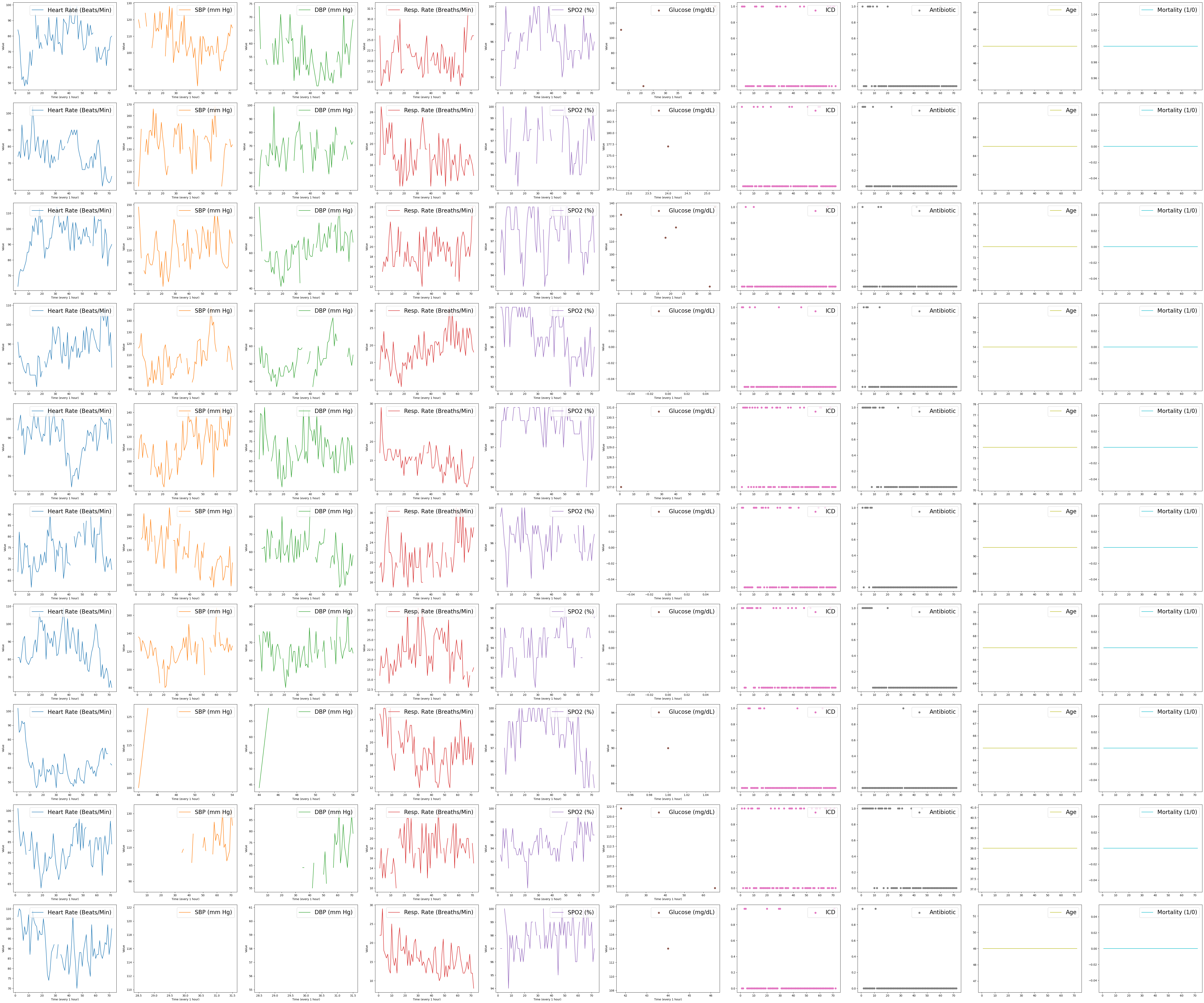}
    \caption{\firstedit{Visualization of synthetic MIMIC-IV time series, ICD codes, antibiotics, age, and mortality data}}
    \label{appendix:add:fig:mimiciv_revision_synth}
\end{figure}
\end{document}